\documentclass[11pt,letterpaper]{article}

\usepackage[letterpaper,margin=1in]{geometry}
\usepackage[T1]{fontenc}
\usepackage[utf8]{inputenc}
\usepackage{mathptmx}      % Times-like serif body (matches reference)
\usepackage{helvet}        % Helvetica for sans (headings)
\usepackage{graphicx}
\usepackage{xcolor}
\usepackage{titlesec}
\usepackage[most]{tcolorbox}
\usepackage{booktabs}
\usepackage{multirow}
\usepackage{tabularx}
\usepackage{enumitem}     % description 环境可选参数（style=unboxed 等）
\usepackage{amsmath,amssymb}
\usepackage[numbers,sort&compress]{natbib}
\usepackage[colorlinks=true]{hyperref}
\usepackage{wrapfig}     % text-wrapped figures in the experiments section
\usepackage{placeins}     % \FloatBarrier: keep floats near their references
\usepackage{subcaption}   % subtable (benchmark tables pasted from two-column template)

\definecolor{lfIndigo}{HTML}{4F46E5}
\definecolor{lfPurple}{HTML}{7C3AED}
\definecolor{lfPink}{HTML}{DB2777}
\definecolor{lfBoxBg}{HTML}{F4F2FD}
\definecolor{lfBoxLine}{HTML}{D9D2F5}

\hypersetup{linkcolor=lfPurple, citecolor=lfPurple, urlcolor=lfPurple}

\titleformat{\section}{\normalfont\sffamily\large\bfseries}{\thesection}{0.6em}{}
\titleformat{\subsection}{\normalfont\sffamily\normalsize\bfseries}{\thesubsection}{0.6em}{}
\titleformat{\subsubsection}{\normalfont\sffamily\normalsize\bfseries}{\thesubsubsection}{0.6em}{}

\begin{document}
\pagestyle{plain}

\thispagestyle{empty}
\noindent\includegraphics[height=1.05cm]{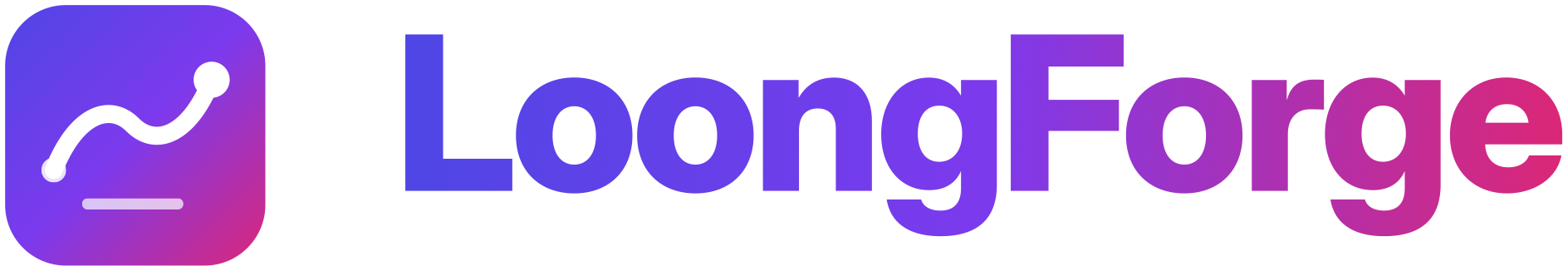}\\[-8pt]
\noindent\textcolor{lfBoxLine}{\rule{\textwidth}{1pt}}

\vspace{1.2em}
\begin{center}
  {\fontsize{22}{26}\selectfont\bfseries LoongForge-Embodied: A Unified Framework \\[2pt]
   for Training Heterogeneous Embodied Models\par}
\end{center}

\vspace{0.6em}
% ---- Author block (edit names/affiliations as needed) ---------------
\begin{center}
  % {\normalsize
  % Author One$^{*}$ \quad Author Two$^{*\dagger}$ \quad Author Three \quad Author Four\\[3pt]
  % Author Five \quad Author Six \quad Author Seven$^{\ddagger}$\par}
  % \vspace{6pt}
  % {\small $^{*}$Equal Contribution \quad $^{\dagger}$Project Lead \quad $^{\ddagger}$Corresponding Author\par}
  % \vspace{4pt}
  {\normalsize
  LoongForge Team\\}
  \vspace{6pt}
  {\small 
  Baidu, Inc.\par}
\end{center}

\vspace{0.8em}
% ---- Shaded abstract box with project-link block --------------------
\begin{tcolorbox}[colback=lfBoxBg, colframe=lfBoxLine, boxrule=0.6pt,
  arc=3pt, left=12pt, right=12pt, top=10pt, bottom=10pt]
  \small
  Training Vision-Language-Action (VLA) models and World-Action Models (WAMs) is computationally expensive, 
  while their heterogeneous architectures and fragmented training stacks make system optimizations difficult 
  to reuse across models. To address these challenges, we present \textbf{\textit{LoongForge-Embodied}}, 
  a native-PyTorch framework that unifies and accelerates the training of heterogeneous embodied models. 
  At its core, LoongForge-Embodied introduces a common abstraction layer that decouples model-specific 
  implementations from shared training infrastructure, enabling cross-model reuse without constraining 
  individual architectures or execution requirements. Building on this unified infrastructure, the framework 
  combines reusable acceleration mechanisms with model-specific and hardware-aware optimization strategies across 
  computation, execution, memory, data I/O, and distributed communication, including FP8-based communication 
  optimizations for DDP and FSDP. 
  Beyond training acceleration, LoongForge-Embodied integrates a unified data infrastructure with a pipeline 
  for synthesizing robot data from egocentric data, and an efficient evaluation framework that supports 
  concurrency and optimized computation in simulator.
  Experiments across representative VLA and WAM models demonstrate 
  1.47×–4.38× end-to-end training throughput improvements over their corresponding open-source baselines, as well as
  4.1×–23.2× faster evaluation under the same protocols.

  \vspace{6pt}
  \noindent\textbf{Website:}~\url{https://baidu-baige.github.io/LoongForge}  \\
  \noindent\textbf{Github:}~\url{https://github.com/baidu-baige/LoongForge} \\
  \noindent\textbf{Checkpoints:}~\url{https://huggingface.co/LoongForge} 
\end{tcolorbox}

\vspace{0.6em}

% =====================================================================
% Body (skeleton — sections follow the report outline)
% =====================================================================
\section{Introduction}
\label{sec:intro}
Recent advances in Vision-Language-Action (VLA) models and World-Action Models (WAMs) have driven embodied 
learning toward larger and more heterogeneous model architectures. VLA models such as OpenVLA \cite{openvla2024}, 
$\pi_{0.5}$ \cite{pi05}, and GR00T \cite{groot_n1} integrate pretrained vision-language backbones with diverse 
action-generation modules, while world-action approaches such as Unified World Models \cite{uwm2025}, 
Cosmos Policy \cite{cosmos_policy2026}, and DreamZero \cite{dreamzero2026} couple action prediction with 
video generation or future-state modeling. Meanwhile, large-scale multi-robot datasets \cite{openx2023,droid2024} 
have substantially expanded the volume and diversity of embodied training data. Together, these developments 
give rise to heterogeneous training workloads with distinct computation graphs, memory footprints, 
communication patterns, and data-processing requirements. Consequently, efficiently scaling embodied 
foundation models requires training infrastructure capable of accommodating such diversity while delivering 
high performance.

Efficient training of heterogeneous VLA and WAM models faces two closely related challenges. First,
diverse model architectures and training objectives give rise to substantially different performance
bottlenecks. Variations in attention structures, computation graphs, activation-memory requirements,
and distributed execution patterns shift the dominant costs among computation, memory, execution
overhead, data I/O, and communication. These bottlenecks also depend on hardware characteristics,
making a fixed optimization strategy insufficient across models and hardware platforms. Second, the
embodied learning ecosystem remains fragmented. Despite recent efforts toward unified robot learning
frameworks \cite{lerobot2024,starvla2026}, many embodied foundation models still rely on independently
developed training stacks, with customized distributed execution, data processing, and evaluation
pipelines. Although broadly reusable system optimization mechanisms exist, including optimized
attention \cite{flashattn2_2023,dong2024flex}, graph compilation \cite{torch2_2023}, CUDA Graphs
\cite{cudagraphs}, and communication optimization \cite{zeropp2023}, their effective integration and
configuration often depend on model architectures and hardware characteristics. Consequently, similar
optimizations must be repeatedly integrated, tuned, and validated across separate training stacks,
limiting cross-model optimization reuse and increasing engineering overhead. This raises a fundamental
systems question: \textbf{can heterogeneous embodied models share a common training infrastructure while still
enabling model-specific, high-performance optimization?}

We address this question with \textbf{LoongForge-Embodied}, a native-PyTorch framework that unifies
and accelerates the training of heterogeneous VLA and WAM models. The key design principle is to
separate model-specific computation from reusable training infrastructure. Although these models
differ substantially in architecture, learning objectives, and execution patterns, they share common
system-level functionalities, including training orchestration, distributed execution, data processing,
optimization, checkpoint management, and evaluation. To capture this commonality without imposing a
fixed computational structure, LoongForge-Embodied introduces a unified abstraction layer built around
configuration-driven model registration, a lightweight model interface, and an extensible training
engine. These abstractions allow heterogeneous models to reuse shared training infrastructure while
preserving their individual computation graphs and training requirements. More importantly, this
separation provides a foundation for cross-model optimization reuse: shared optimization mechanisms
can be integrated once at the framework level, while architecture-specific and hardware-aware
optimization choices remain customizable. In this way, LoongForge-Embodied combines the benefits of
unified training infrastructure with the flexibility required for model-specific, high-performance
optimization.

Building on this unified abstraction, LoongForge-Embodied provides a composable acceleration toolkit
that combines reusable system-level optimization mechanisms with model-specific and hardware-aware
strategies. Guided by the performance bottlenecks of individual workloads, the framework integrates
efficient attention kernels \cite{flashattn2_2023,dong2024flex,flashattn3}, mixed-precision computation,
and fused operators and optimizers to improve computational efficiency. Graph compilation
\cite{torch2_2023} and configurable CUDA Graph capture \cite{cudagraphs} reduce execution overhead,
while activation checkpointing \cite{chen2016training} and parameter-efficient adaptation address
memory constraints. Precomputation, caching, and asynchronous checkpointing further mitigate
data-processing and I/O overhead. For distributed training, LoongForge-Embodied supports both Distributed Data Parallel (DDP) and
Fully Sharded Data Parallel (FSDP) with FP8-based communication optimizations, reducing gradient synchronization traffic in DDP and
parameter AllGather traffic in FSDP, particularly in bandwidth-constrained environments. Beyond
communication compression, configurable synchronization and prefetching strategies improve
computation--communication overlap. Rather than imposing a uniform optimization recipe,
LoongForge-Embodied enables heterogeneous models to reuse shared acceleration mechanisms while
retaining the flexibility to select architecture-specific implementations and hardware-aware
configurations.

Beyond training acceleration, LoongForge-Embodied provides a unified data infrastructure and 
evaluation framework. The data infrastructure of LoongForge-Embodied 
accommodates heterogeneous dataset formats, model-specific processing requirements, and cross-embodiment data 
synthesis from egocentric vision to robot manipulation, while the evaluation framework enables composable model 
and benchmark integration through shared inference implementations. In addition, LoongForge-Embodied accelerates 
evaluation through concurrency, reduced per-episode overhead, and reuse of computation.
We evaluate LoongForge-Embodied across representative VLA and WAM models with diverse architectures 
and training characteristics. Compared with their corresponding open-source implementations, the framework achieves 
1.47$\times$--4.38$\times$ improvements in end-to-end training throughput, demonstrating its effectiveness across 
heterogeneous embodied workloads. 
With the same protocols, the evaluation of LoongForge-Embodied achieves 
4.1\(\times\)–23.2\(\times\) speedup over the baseline.

Our main contributions are summarized as follows:

\begin{itemize}
	\item \textbf{A unified training framework.} We introduce LoongForge-Embodied, a native-PyTorch framework that unifies the training of heterogeneous VLA and WAM architectures through a common abstraction layer, enabling reusable training infrastructure across models.
	\item \textbf{System-level acceleration.} We develop a composable acceleration stack spanning efficient attention, compilation, CUDA Graphs, operator and optimizer fusion, data loading, and FP8-based distributed communication for both FSDP and DDP.
	\item \textbf{Evaluation acceleration.} We accelerate the evaluation loop through concurrent episode execution, reduced per-episode inference and rendering cost, and reuse of policy-independent computation, reducing evaluation wall-clock time by 4.1$\times$--23.2$\times$ under an unchanged benchmark protocol.
	\item \textbf{Broad empirical validation.} We evaluate LoongForge-Embodied across nine representative VLA/WAM models and widely used embodied benchmarks, achieving 1.47$\times$--4.38$\times$ training speedups while maintaining task success rates comparable to official baselines.
\end{itemize}

\section{Background and Motivation}
\label{sec:background}

This section introduces the modeling and training paradigms of VLA and WAM models, examines the challenges arising from their heterogeneous architectures and workloads, and outlines the key design requirements for a unified and efficient embodied training framework.

\subsection{Vision-Language-Action Models}
\label{sec:background_vla}

VLA models learn robot control policies
that map multimodal observations and language instructions to
continuous robot actions. Given a visual observation $o_t$, a
language instruction $l$, and a proprioceptive state $s_t$,
a VLA model predicts an action chunk
$A_t = (a_t,\ldots,a_{t+H-1})$ over a horizon of $H$ steps.
The policy is formulated as
\begin{equation}
    \pi_\theta(A_t \mid c_t),
    \qquad c_t = (o_t,l,s_t),
    \label{eq:vla_policy}
\end{equation}
where $\theta$ denotes the model parameters and $c_t$ represents
the multimodal conditioning information. Representative VLA
models supported by LoongForge-Embodied, including
$\pi_{0.5}$~\cite{pi05}, GR00T~\cite{groot_n1},
X-VLA~\cite{xvla}, and Wall-OSS-0.5~\cite{walloss05},
employ flow-based generative policies to produce continuous
action sequences conditioned on multimodal observations.

A common approach to learning such policies is conditional
flow matching, which trains a velocity field to transform
Gaussian noise into demonstrated action sequences. Given
a ground-truth action chunk $A_t$, Gaussian noise
$\epsilon \sim \mathcal{N}(0,I)$, and a flow time
$\tau \in [0,1]$, an interpolated action is constructed as
$A_t^\tau = (1-\tau)\epsilon + \tau A_t$.
The velocity field $v_\theta$ is trained to predict the
corresponding flow direction by minimizing
\begin{equation}
    \mathcal{L}_{\mathrm{FM}}(\theta)
    =
    \mathbb{E}_{(c_t,A_t)\sim\mathcal{D},
                \tau\sim p(\tau),
                \epsilon\sim\mathcal{N}(0,I)}
    \left[
        \left\|
            v_\theta(A_t^\tau,\tau,c_t)
            - (A_t-\epsilon)
        \right\|_2^2
    \right],
    \label{eq:vla_fm}
\end{equation}
where $\mathcal{D}$ denotes the demonstration dataset and
$p(\tau)$ is the flow-time sampling distribution.
This objective provides a representative formulation of
continuous action learning rather than an identical loss
implementation across all models. In practice, these models
differ in their conditioning mechanisms and training
objectives: $\pi_{0.5}$ incorporates high-level semantic
supervision, X-VLA introduces embodiment-specific soft
prompts, and Wall-OSS-0.5 combines continuous action
learning with discrete action-token and multimodal
prediction objectives. Such differences allow a common
policy-learning formulation to be realized through
heterogeneous model architectures and training procedures.

\subsection{World-Action Models}
\label{sec:background_wam}

WAMs extend action learning by
incorporating predictive modeling of future world dynamics.
While VLA models primarily learn to generate actions
conditioned on multimodal observations, WAMs additionally
learn the relationships between future world states and
robot actions. Let $Z_t^+$ denote a sequence of future
world representations, such as video latents, and let
$c_t$ and $A_t$ follow the definitions in
Section~\ref{sec:background_vla}. A representative
world-action generative formulation is
\begin{equation}
    p_\theta(Z_t^+, A_t \mid c_t),
    \label{eq:wam_joint}
\end{equation}
which models future world dynamics and action sequences
conditioned on the current multimodal context. Depending
on the model architecture, world representations may be
jointly generated with actions, used to condition action
generation, or introduced as auxiliary supervision during
training. Consequently, explicit future-world generation
is not necessarily required during policy execution.

The training of WAMs typically combines world prediction
and action learning, which can be summarized by a
representative multi-objective formulation:
\begin{equation}
    \mathcal{L}_{\mathrm{WAM}}
    =
    \lambda_w \mathcal{L}_{\mathrm{world}}
    +
    \lambda_a \mathcal{L}_{\mathrm{action}},
    \label{eq:wam_loss}
\end{equation}
where $\mathcal{L}_{\mathrm{world}}$ and
$\mathcal{L}_{\mathrm{action}}$ supervise future-world
modeling and action generation, respectively.
Both objectives can be implemented through conditional
flow matching, as introduced in
Section~\ref{sec:background_vla}, although their
generative variables, conditioning information, and
loss weighting may differ. Representative WAMs supported
by LoongForge-Embodied illustrate these variations:
LingBot-VA~\cite{lingbotva2026} and
DreamZero~\cite{dreamzero2026} couple future visual
generation with action prediction, whereas
Fast-WAM~\cite{fastwam2026} employs world modeling
during training without requiring explicit future-video
generation at inference time. Cosmos 3~\cite{cosmos3}
further supports multiple world-action generation tasks,
including forward dynamics and action prediction.
These formulations share the objective of learning
world-action relationships but differ in how world
modeling and action generation are coupled.

\subsection{Training Challenges and System Requirements}

Training large-scale VLA and WAM models is computationally
expensive due to their complex multimodal architectures and
generative training objectives. Within individual models,
vision-language backbones, action experts, and world-modeling
components exhibit different computational structures,
attention patterns, and gradient propagation paths. Across
models, variations in component composition, training
objectives, and world-action coupling further introduce
heterogeneous computation graphs and resource requirements.
Consequently, training performance is constrained by diverse
bottlenecks in computation, execution, memory, communication,
and data I/O, whose relative importance also varies across
hardware configurations. Addressing these bottlenecks
requires targeted training optimizations, yet existing
embodied models typically rely on fragmented, model-specific
training stacks, making optimization techniques difficult
to integrate, reuse, and maintain across architectures.
These challenges call for a unified training infrastructure
that enables efficient and reusable acceleration across
heterogeneous embodied workloads.

\textbf{Desiderata for Embodied Training Systems.}
The architectural heterogeneity and diverse training workloads
of VLA and WAM models motivate the following system design
requirements:

\begin{description}[
    leftmargin=2em,
    labelindent=2em,
    itemsep=3pt]
    \item[\textbf{D1. Unified.}]
    A common training infrastructure for heterogeneous VLA
    and WAM models, eliminating fragmentation across
    model-specific training stacks.

    \item[\textbf{D2. Flexible.}]
    Support for diverse model architectures, training
    objectives, and execution patterns without imposing
    a fixed computational structure.

    \item[\textbf{D3. Reusable.}]
    Cross-model reuse of optimization mechanisms, reducing
    redundant implementation, integration, and tuning efforts.

    \item[\textbf{D4. Model-specific.}]
    Targeted optimizations for specific model architectures,
    computational patterns, and training characteristics.

    \item[\textbf{D5. Hardware-aware.}]
    Hardware-specific optimization strategies tailored
    to different GPU architectures, memory capacities,
    and interconnect bandwidths to enhance training efficiency.

    \item[\textbf{D6. Faithful.}]
    Preservation of consistent training loss behavior and
    downstream task performance under training
    acceleration.

    \item[\textbf{D7. Data-integrated.}]
    Unified and extensible data infrastructure for
    multimodal data representation, processing, loading,
    and generation.

    \item[\textbf{D8. Evaluation-ready.}]
    Standardized evaluation interfaces and protocols
    enabling consistent and reproducible evaluation
    across heterogeneous models and embodied benchmarks.
\end{description}

Guided by these desiderata, LoongForge-Embodied establishes
a unified training infrastructure for heterogeneous VLA
and WAM models. The following
sections present the system architecture and its key
optimization techniques.

\section{LoongForge-Embodied Overview}
\label{sec:overview}

This section presents the overall architecture of LoongForge-Embodied and describes the unified model and training abstractions that enable heterogeneous VLA and WAM models to share common training infrastructure while supporting model-specific execution and optimization.

\subsection{System Architecture}

\begin{figure}[!htbp]
  \centering
  \includegraphics[width=0.98\textwidth]{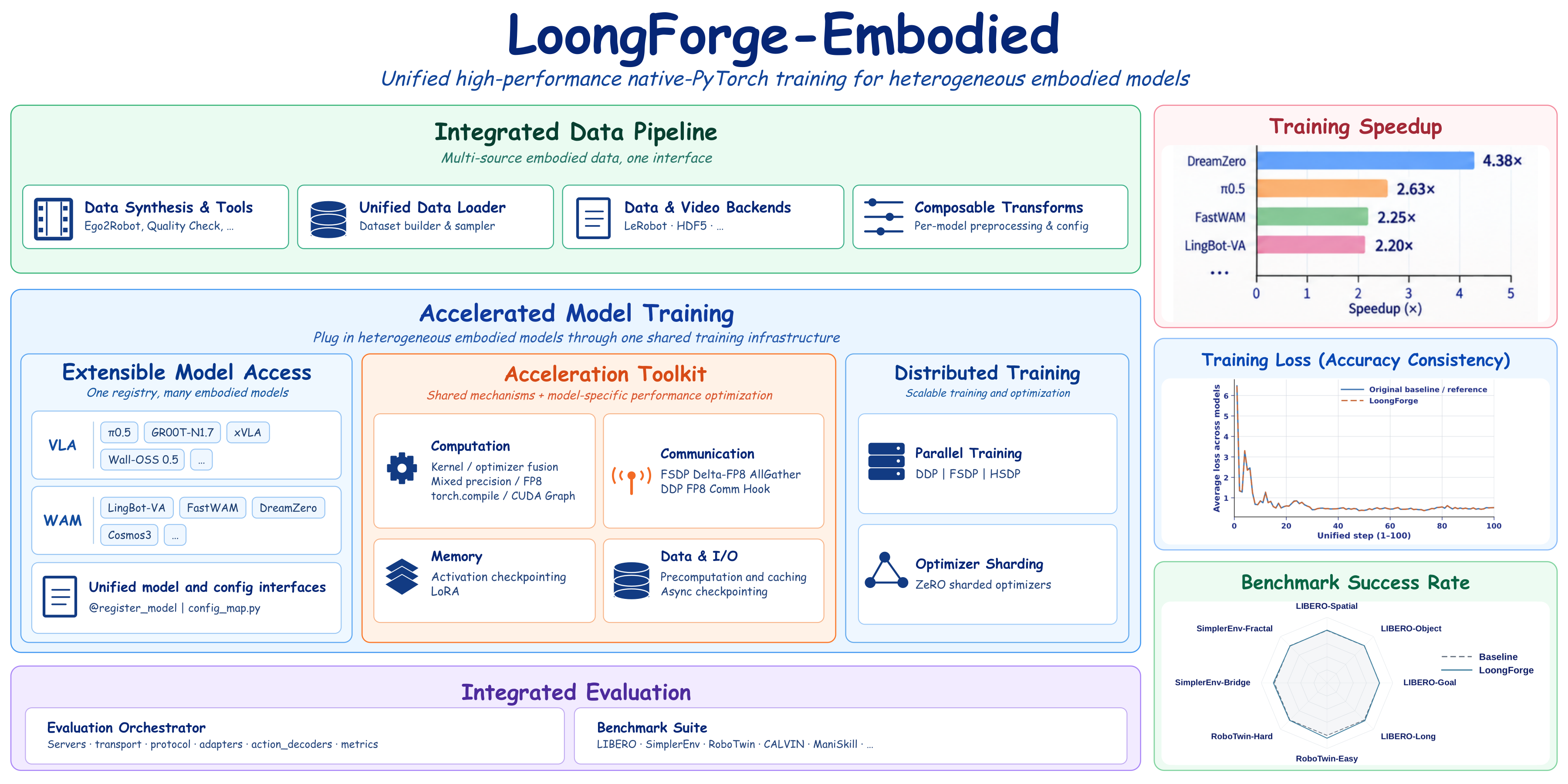}
  \caption{LoongForge-Embodied system architecture. The framework unifies
  heterogeneous VLA and WAM models under a single native-PyTorch training
  stack, combining an integrated data pipeline (Ego2Robot, unified data
  loaders, and composable transforms), an acceleration toolkit of shared
  optimizations with model-specific execution paths, distributed training
  with FP8 communication, and an integrated evaluation orchestrator. The
  right-hand panels preview representative results on training speedup,
  training-loss consistency, and benchmark success rates
  (Section~\ref{sec:exp}).}
  \label{fig:overview}
\end{figure}

Figure~\ref{fig:overview} illustrates the overall architecture
of LoongForge-Embodied. The central design principle is to
decouple model-specific computations from reusable training
infrastructure. Rather than maintaining separate training
stacks for different VLA and WAM models, LoongForge-Embodied
provides a unified system in which heterogeneous models
share common infrastructure while retaining their individual
architectures, training objectives, and execution requirements.

The system consists of several cooperating components.
A configuration-driven model integration layer manages
model registration, construction, and model-specific
interfaces. A unified data infrastructure processes
heterogeneous robotic datasets through common data
representations and model-specific adaptations. The
shared training engine orchestrates training execution,
while an acceleration toolkit provides reusable
optimizations across computation, execution, memory,
and data I/O. Distributed training infrastructure
supports multi-GPU execution and communication
optimization. Finally, a unified evaluation framework
connects model inference with benchmark environments
for consistent downstream performance assessment.

These components are integrated into a common training
workflow. Given a training configuration, the framework
resolves the corresponding model and trainer, constructs
the data pipeline, and initializes the training runtime.
Model-specific optimization requirements are then
integrated with shared acceleration and distributed
execution mechanisms. During training, the framework
manages optimization, checkpointing, and state recovery,
while the evaluation subsystem provides a common interface
for assessing trained models. This architecture separates
model implementation from system-level execution, enabling
both cross-model infrastructure reuse and model-specific
performance optimization.

\subsection{Unified Model and Training Abstractions}

\textbf{Configuration-driven model integration.}
LoongForge-Embodied adopts a configuration-driven
registration mechanism to integrate heterogeneous models
without modifying the shared training infrastructure.
A structured configuration schema associates each model
with its configuration types and corresponding data
configuration. A model registry maps model identifiers
to implementation classes, which are instantiated through
a common construction interface. Model implementations
are automatically discovered and registered, avoiding
manual modifications to a centralized registry when
introducing new models. Similarly, a trainer registry
selects the appropriate trainer implementation according
to the training configuration. Together, these mechanisms
decouple model and trainer selection from their
implementations.

\textbf{Unified model contract.}
All integrated models expose a common interface consisting
of model construction and a forward operation that accepts
a training batch and returns a loss and diagnostic metrics.
This contract allows the training engine to execute
heterogeneous models without requiring knowledge of
their internal architectures or learning objectives.
Model-specific behaviors are expressed through extensible
interfaces, including the selection of FP8 and LoRA
target modules and model-specific distributed wrapping.
For example, individual models can exclude
precision-sensitive components from FP8 conversion or
restrict optimization to selected network modules.
Consequently, the framework standardizes how models
interact with the training infrastructure without
constraining how they implement their computations.

\textbf{Template-based training execution.}
The training engine follows a hierarchical template-method
design that separates training orchestration from
model-specific execution. At the highest level,
a shared training loop controls the iteration lifecycle.
A common training-step abstraction handles operations
such as gradient initialization, optimizer execution,
and gradient clipping, while an extensible
forward-backward interface encapsulates model-dependent
computation. Most models reuse the default training
procedure, whereas specialized trainers can override
selected execution hooks to support architecture-specific
requirements, such as customized CUDA Graph execution.
This design avoids duplicating complete training loops
while preserving flexibility for model-specific
optimizations.

\textbf{Composable execution and state management.}
LoongForge-Embodied further provides a common mechanism
for composing optimization and distributed execution
strategies. Model wrapping follows an explicitly
ordered sequence: FP8 conversion, activation checkpointing,
and distributed wrapping through DDP or FSDP.
This ordering preserves compatibility between module
transformation and activation recomputation, preventing
inconsistencies that may arise from applying these
transformations in an incompatible order. Models may
specialize selected wrapping procedures through
well-defined hooks without modifying the shared
execution framework. The same infrastructure also
provides unified checkpoint management across multiple
serialization formats, supporting asynchronous saving
and recovery of model, optimizer, data-loading, and
random-number-generator states. For distributed
checkpoints, state restoration is coordinated with
model wrapping to support resharding across different
distributed configurations. These mechanisms maintain
a common execution and recovery lifecycle across
heterogeneous training workloads.

\section{Training Acceleration}
\label{sec:accel}
\subsection{Computation Acceleration}
\label{sec:computation_acceleration}

\textbf{Shared mechanisms and model-specific optimizations.}
LoongForge-Embodied separates reusable optimization infrastructure from architecture-dependent implementations. Shared capabilities include optimizer management, activation checkpointing, precision control, LoRA, and data loading, while attention kernels, operator fusion, graph compilation, CUDA-graph capture, and feature caching admit model-specific implementations. The framework unifies their configuration, registration, and compatibility management, allowing heterogeneous models to compose shared mechanisms with specialized optimizations. We organize these techniques by the bottlenecks they address.

\textbf{Computation efficiency.}
Attention is a major computational bottleneck in VLA and WAM models. LoongForge-Embodied provides a unified attention dispatch mechanism across PyTorch, FlashAttention-2/3, and Transformer Engine, with SDPA fallback for unsupported backends or hardware. Hardware-aware selection accommodates the Ampere requirement of FlashAttention-2 and the Hopper requirement of FlashAttention-3~\cite{flashattn2_2023,flashattn3}. For model-specific attention patterns, FlexAttention supports compiled block masks for joint attention and sequence packing, avoiding materialization of dense attention masks~\cite{dong2024flex}.

Additional efficiency comes from fused operators and optimizers. Model-specific fusion of rotary positional embedding (RoPE), normalization, modulation, and loss computation reduces kernel launches and memory traffic~\cite{tillet2019triton}. A shared optimizer registry integrates fused AdamW implementations from PyTorch, Transformer Engine, and Apex~\cite{loshchilov2019decoupled}; multi-dtype zero-redundancy optimizers shard optimizer states, and foreach multi-tensor updates reduce per-tensor overhead.

Precision management supports module-wise BF16/FP16 computation with FP32 master weights, as well as FP8 linear computation through two interchangeable backends: Transformer Engine modules and TorchAO float8 linear layers, selected per model according to the backbone implementation and hardware support~\cite{transformerengine,torchao}. Under the Transformer Engine backend, configurable scaling recipes include blockwise, current, delayed, and MXFP8, with FP8 autocast covering forward and backward computation~\cite{mxfp2024}. This compute-side FP8 optimization is distinct from the FP8 communication mechanism in Section~\ref{sec:fp8}. For convolutional workloads, cuDNN autotuning selects efficient algorithms for stable shapes, with explicit compatibility checks against deterministic execution and coordinated TF32 settings. Fake/meta implementations further preserve \texttt{torch.compile} compatibility for custom cuDNN-backed kernels~\cite{chetlur2014cudnn}.

Beyond kernel selection and fusion, graph compilation and CUDA graphs reduce the cost of generating and submitting kernels. \texttt{torch.compile} enables operator fusion and optimized kernel generation, with configurable compilation scopes spanning the full forward pass, individual layers, or grouped layers with final normalization and action heads. The grouped scope can facilitate overlap with DDP. Compilation mode, fullgraph execution, and dynamic-shape handling are configurable to limit graph breaks and recompilation~\cite{torchcompile2024}. Complementarily, CUDA graphs eliminate repeated CPU-side kernel launches through capture and replay. The framework supports whole-iteration and per-microbatch capture, with warmup, sequence padding, and optional in-graph gradient synchronization. Captured training graphs can also be pipelined with separate frozen-backbone graphs, while subfunctions incompatible with capture are compiled separately and bypass it~\cite{cudagraphs}. Thus, compilation optimizes kernel generation, whereas CUDA graphs optimize kernel submission.

\textbf{Memory efficiency.}
Activation checkpointing reduces memory consumption by recomputing selected activations during backpropagation~\cite{chen2016training}. LoongForge-Embodied exposes module-level inclusion and exclusion patterns, controlled through model configuration and consistently applied during distributed wrapping. For parameter-efficient adaptation, shared LoRA integration supports configurable ranks, target modules, and initialization, reducing trainable parameters, optimizer states, and gradient communication~\cite{hu2021lora}. Together with memory-efficient attention, these mechanisms address both activation and parameter memory bottlenecks.

\textbf{Data preparation and I/O.}
The framework reduces repeated preprocessing through a structured feature-cache abstraction comprising configuration, artifact contracts, and dataset-level build, validation, and loading utilities. Model-specific features, including text embeddings and episode statistics, can be precomputed and reused through an explicitly configured cache directory~\cite{lerobot2024}. The data pipeline additionally supports five video-decoding backends, prefetching, persistent workers, and stateful loading. Asynchronous checkpointing overlaps disk I/O with training, with at most one in-flight save.

These optimizations are independently configurable and preserve the underlying training objectives. Each model activates only the relevant mechanisms and supplies architecture-specific implementations where necessary. Section~\ref{sec:model_specific} and Appendix~\ref{app:model_analysis} examine these model-dependent choices and their performance implications.

\subsection{Distributed Training Strategies}

\textbf{Unified Parallelism Abstraction.}
VLA and WAM models exhibit diverse parameter scales, architectural structures, and memory requirements, making a single distributed training strategy insufficient for efficient execution across models. LoongForge-Embodied addresses this heterogeneity through a unified parallelism abstraction that decouples distributed execution from model-specific implementations. The framework supports DDP and FSDP2 through a common model-wrapping interface, allowing the parallelism strategy to be selected through configuration without modifying the underlying model or training loop. DDP maintains replicated model parameters and synchronizes gradients across workers, whereas FSDP2 partitions model parameters, gradients, and optimizer states to reduce per-device memory consumption. The framework additionally supports Hybrid Sharded Data Parallelism (HSDP), combining parameter sharding within device groups with replication across groups. These strategies share the same distributed initialization, training-loop control, and gradient accumulation infrastructure, enabling heterogeneous embodied models to adopt different parallel configurations while reusing a common training pipeline.

\textbf{Architecture-Aware Sharding and Memory Management.}
Efficient distributed execution requires adapting parameter partitioning and memory management to the structural characteristics of individual models. Rather than imposing a fixed sharding granularity, LoongForge-Embodied provides an architecture-aware FSDP2 wrapping mechanism that supports both explicit and automatic partitioning. Models may specify module boundaries for fine-grained control, while the automatic strategy constructs sharding units based on module structure and parameter counts, respecting architectural constraints that prevent certain modules from being split. For models containing parameters with heterogeneous data types, the framework further separates mixed-precision submodules into compatible sharding groups and organizes parameter prefetching according to module execution order. These mechanisms allow sharding granularity and parameter materialization to be adapted to different model structures. Beyond parameter sharding, the framework supports complementary memory-efficient configurations, including ZeRO-1 optimizer-state partitioning under DDP and hybrid sharding under FSDP2. Together, these mechanisms provide flexible trade-offs among parameter replication, memory consumption, and communication overhead without requiring separate distributed training implementations for each model.

\textbf{Precision-Aware Wrapping and Gradient Synchronization.}
Combining distributed parallelism with mixed precision, activation checkpointing, and gradient accumulation introduces dependencies that must be coordinated to preserve correct execution. LoongForge-Embodied separates original parameter precision, computation precision, and FP8 conversion, allowing these mechanisms to be configured independently. The framework enforces a consistent model-wrapping sequence in which FP8 conversion precedes activation checkpointing, followed by DDP or FSDP2 wrapping. This ordering is particularly important for FP8 training, where applying checkpointing before precision conversion can introduce inconsistencies between tensors saved during the forward pass and those reconstructed during recomputation. The framework also unifies gradient synchronization across parallel backends through an explicit synchronization context. During gradient accumulation, communication is suppressed for intermediate micro-batches using backend-specific mechanisms, while gradient synchronization is enabled for the final micro-batch before the optimizer update. This avoids redundant communication while preserving consistent gradient accumulation semantics across DDP and FSDP2. By coordinating precision conversion, recomputation, and synchronization within a shared execution pipeline, the framework enables these optimization mechanisms to be composed across heterogeneous models.

\textbf{Model-Specific Distributed Execution.}
Although the shared parallelism abstraction covers common distributed training requirements, certain model architectures and execution patterns require specialized strategies to achieve high performance. LoongForge-Embodied therefore exposes model-specific wrapping and execution hooks that allow individual models to customize distributed behavior while retaining the shared training infrastructure. For example, models employing distributed Muon optimization can explicitly manage parameter ownership and FSDP sharding groups, enabling optimizer-specific parameter routing and overlapping parameter prefetching with optimizer computation. Similarly, GR00T N1.7 integrates full-iteration CUDA Graph execution with DDP by coordinating static gradient bucket construction with the backward execution order. These extensions demonstrate that the unified distributed infrastructure does not enforce identical parallel execution across models. Instead, it provides reusable parallelism mechanisms while preserving the flexibility needed for architecture- and execution-specific performance optimizations.

\subsection{Communication Optimization}
\label{sec:fp8}
Distributed training introduces substantial communication overhead, particularly for large-scale VLA and WAM models, and its impact varies with model architectures, parallel configurations, and hardware interconnects. Rather than organizing communication optimization around individual parallel strategies, LoongForge-Embodied targets the collective operations that occur at each stage of a training step: parameter gathering in the forward and backward passes, gradient reduction in the backward pass, and parameter synchronization in the optimizer step. For each class, the framework provides reusable mechanisms that reduce transferred bytes or hide communication latency behind computation, while allowing model- and hardware-specific configurations.

\textbf{Parameter gathering in the forward and backward passes.}
When parameters are sharded, each layer's parameters are gathered through AllGather before they are used in the forward pass and, if released after the forward pass, gathered again for backward computation. LoongForge-Embodied supports delta-FP8 AllGather for this operation. Instead of repeatedly transmitting full-precision parameters, it quantizes and communicates parameter updates relative to previously reconstructed values. Each rank applies the same reconstructed increments to a shared reference, maintaining consistent parameter replicas while correcting residual quantization errors across iterations. The implementation reuses existing AllGather buffers as reference storage when the parameter layout permits, reducing additional memory overhead, and retains full-precision communication for reference initialization and unsupported configurations. To hide the remaining latency, configurable forward and backward prefetching issues each AllGather ahead of the corresponding parameter consumption, allowing communication to overlap with the computation of preceding layers.

\textbf{Gradient reduction in the backward pass.}
Gradients produced in the backward pass are synchronized across data-parallel ranks, conventionally through AllReduce. LoongForge-Embodied provides configurable communication hooks for this reduction, supporting standard AllReduce, reduced-precision communication, and FP8-based compression. The FP8 path decomposes AllReduce into asynchronous AllToAll and AllGather operations: gradients are quantized blockwise and exchanged together with their scaling factors through AllToAll, the received partial gradients are accumulated in FP32, and the reduced result is requantized and redistributed through AllGather. This design reduces communication volume while limiting numerical errors from repeated quantization. The hooks are selected independently of model implementations and fall back to standard AllReduce when message sizes, hardware capabilities, or memory constraints make compression unsuitable. To avoid redundant reductions, cross-rank gradient synchronization during gradient accumulation is deferred until the final micro-step, and an alternative gradient synchronization path is provided for CUDA Graph execution.

\textbf{Parameter synchronization in the optimizer step.}
The optimizer step can also involve parameter communication, for example when optimizer computation is distributed across ranks, as in the distributed Muon configuration described in the preceding subsection. For this stage, the framework supports asynchronous parameter synchronization during optimizer execution, so that parameter exchanges overlap with optimizer computation rather than being serialized after the update.

\textbf{Hardware-aware adaptation.}
The benefit of these mechanisms is not uniform: communication compression does not necessarily translate into lower training latency, as it depends on message sizes, communication scheduling, and interconnect characteristics. LoongForge-Embodied therefore exposes communication precision, compression thresholds, and synchronization strategies as configuration for each class of collective, allowing different models and hardware platforms to select suitable optimizations. Unsupported or inefficient compression paths fall back to standard collectives, preserving training compatibility while avoiding unnecessary communication overhead.

\subsection{Model-Specific Optimization Strategies}
\label{sec:model_specific}

Although all models share the reusable acceleration mechanisms described in
the preceding subsections, profiling shows that the dominant training
bottleneck differs substantially across architectures. We group these
bottlenecks into four categories: \emph{fragmented execution}, in which
short kernels and host synchronization leave the GPU intermittently idle;
\emph{memory--compilation coupling}, in which activation memory, batch size,
compilation granularity, and communication overlap constrain one another;
\emph{data delivery}, in which CPU-side decoding and preprocessing stall the
GPU; and \emph{distributed communication}, in which gradient or parameter
collectives are exposed on the critical path, particularly over
bandwidth-constrained interconnects. Table~\ref{tab:model_opt_summary}
summarizes, for each evaluated model, the dominant bottleneck and the
principal model-specific optimizations. Per-model descriptions of the
architectures, bottlenecks, and optimization designs are provided in
Appendix~\ref{app:model_analysis}, and the corresponding experiments are
reported in Sections~\ref{sec:opt_analysis} and~\ref{sec:dist_comm}.

These model-specific optimizations follow a common design principle: rather
than maintaining a separately engineered training stack for each model,
LoongForge-Embodied derives each model's optimization recipe by composing
the reusable mechanisms described above, guided by profiling. For each
model, we first profile the original training implementation to locate the
dominant bottleneck on the critical path, then select and configure the
mechanisms that address it---such as the compilation granularity,
kernel-fusion targets, distributed strategy, communication precision, and
data-path configuration---and finally verify numerical fidelity against
the reference implementation (Section~\ref{sec:training_fidelity}).
Model-specific code is confined to what the reusable mechanisms cannot
express, such as restructuring data-dependent control flow into static
graphs or adapting parameter resharding to the gradient-accumulation
schedule. Consequently, the same mechanism may be applied differently, or
not at all, depending on where profiling locates the bottleneck.

\newlength{\modelcolwidth}
\begin{table}[!ht]
\centering
\caption{\textbf{Dominant training bottlenecks and principal model-specific
optimizations.} Bottlenecks are identified by profiling each model's
original training implementation. The last column lists where the
corresponding experiments are reported.}
\label{tab:model_opt_summary}
\footnotesize
\setlength{\tabcolsep}{4pt}
\renewcommand{\arraystretch}{1.15}
\settowidth{\modelcolwidth}{(Wan2.2-TI2V-5B)}
\begin{tabularx}{\textwidth}{@{}>{\raggedright\arraybackslash}p{\modelcolwidth} l >{\raggedright\arraybackslash}X >{\raggedright\arraybackslash}X l@{}}
\toprule
\textbf{Model} & \textbf{Type} & \textbf{Dominant bottleneck} & \textbf{Key optimizations} & \textbf{Evaluated in} \\
\midrule
$\pi_{0.5}$ & VLA & Activation memory, compilation granularity, gradient synchronization & FlexAttention, selective checkpointing, multi-group compilation, DDP bucket tuning & \S\ref{sec:opt_analysis}, App.~\ref{app:pi05} \\
GR00T N1.7 & VLA & Fragmented execution and host synchronization (78.2\% of kernels $<20\,\mu$s) & Metadata caching, operator fusion, CUDA Graph, cross-iteration pipelining & \S\ref{sec:opt_analysis}, App.~\ref{app:gr00t} \\
X-VLA & VLA & Host synchronization, fine-grained operators, data loading & Static-graph restructuring, \texttt{torch.compile}, data-loader tuning & \S\ref{sec:opt_analysis}, App.~\ref{app:xvla} \\
Wall-OSS-0.5 & VLA & Unfused element-wise operators and kernel launches, exposed FSDP parameter communication & Bitwise-exact SwiGLU/RMSNorm kernels, execution-order prefetching, FSDP resharding & \S\ref{sec:opt_analysis}, App.~\ref{app:walloss} \\
FastWAM & WAM & Bandwidth-limited gradient communication & ZeRO-1 $\rightarrow$ DDP, FP8 AllToAll--AllGather, asynchronous overlap, fused kernels & \S\ref{sec:dist_comm}, App.~\ref{app:fastwam} \\
LingBot-VA & WAM & Redundant FSDP parameter gathers, recomputed short kernels, rank load imbalance from variable-length samples & Accumulation-aware FSDP2 resharding, selective compilation, TE/Triton kernels, cost-balanced sampling & \S\ref{sec:dist_comm}, App.~\ref{app:lingbotva} \\
DreamZero (Wan2.2-TI2V-5B) & WAM & FSDP parameter communication, video data processing & No-reshard FSDP, Delta-FP8 AllGather, whole-block compilation, precomputed-feature cache, first-frame-only processing & \S\S\ref{sec:opt_analysis}--\ref{sec:dist_comm}, App.~\ref{app:dreamzero} \\
Cosmos 3 & WM & CPU preprocessing, memory footprint, FSDP communication & GPU preprocessing, \texttt{torch.compile}, selective FSDP sharding, FP8 & \S\S\ref{sec:opt_analysis}--\ref{sec:dist_comm}, App.~\ref{app:cosmos3} \\
\bottomrule
\end{tabularx}
\end{table}

\section{Unified and Extensible Data Infrastructure}
\label{sec:data_infrastructure}

The heterogeneity of embodied models extends beyond their architectures
to the data representations and processing procedures required for
training. Different VLA and WAM models rely on diverse dataset formats,
temporal observation structures, action representations, sampling
policies, and preprocessing operations. In model-specific training
stacks, these requirements are often embedded directly into individual
data-loading implementations, resulting in duplicated functionality
and making data infrastructure difficult to reuse across models.
A unified training framework must therefore accommodate heterogeneous
data requirements without imposing a fixed data-processing procedure
or altering the training inputs expected by individual models.

LoongForge-Embodied addresses these challenges through a unified and
extensible data infrastructure that separates storage formats from
model-specific data processing. Heterogeneous datasets are exposed
through a common sample contract, while model-specific behaviors are
introduced through registries and orthogonal extension interfaces.
On top of these abstractions, the framework integrates efficient data
delivery, validated caching, and stateful resumption to support
high-throughput and reproducible training. It further incorporates
Ego2Robot~\cite{ego2robot2026} to synthesize robot-compatible trajectories from human
interaction data, with the generated data directly integrated into
the same training pipeline.

\subsection{Unified Data Abstraction and Contract}
\label{sec:unified_data}

\begin{figure}[!ht]
  \centering
  \includegraphics[width=0.88\textwidth]{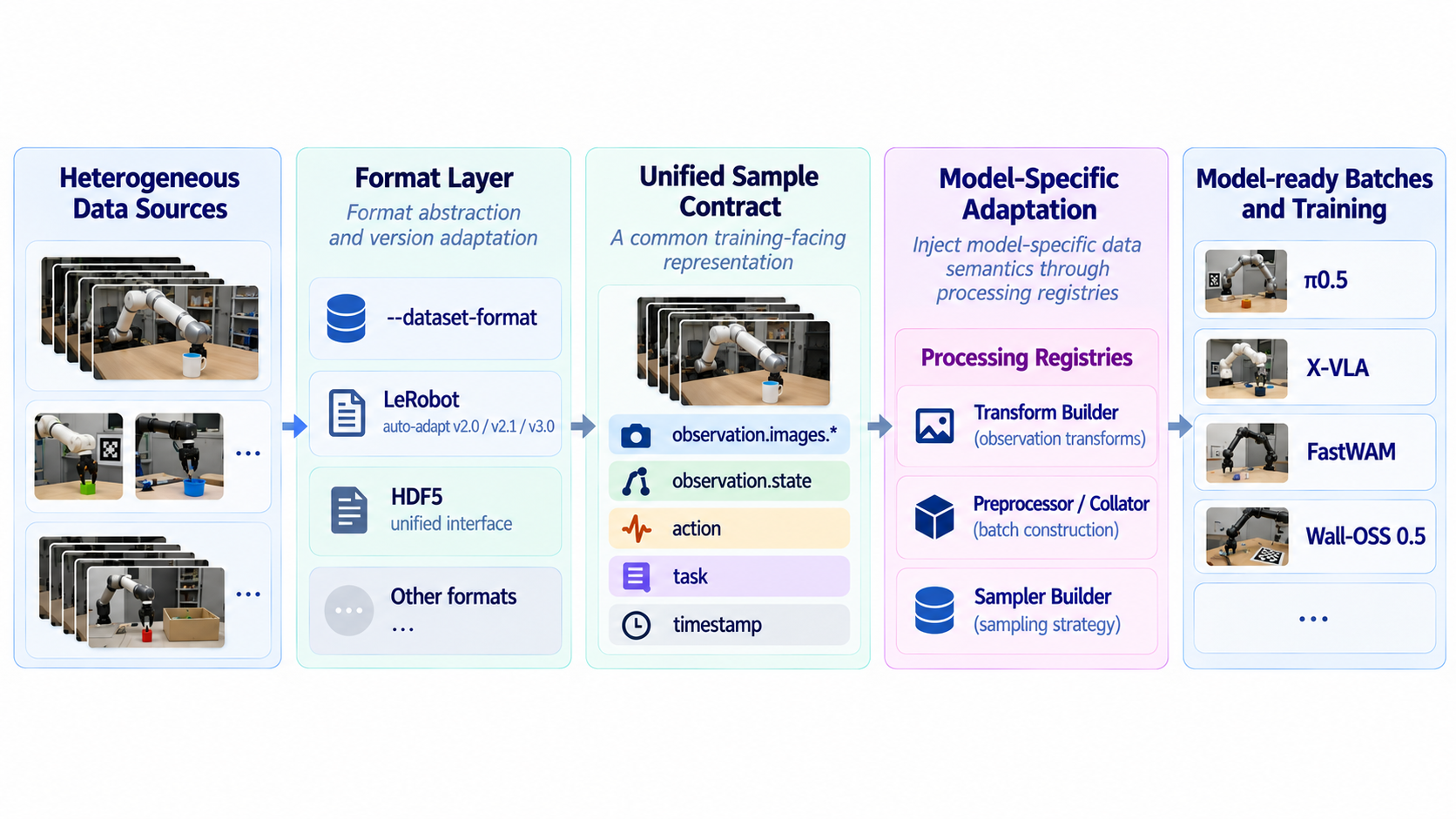}
  \caption{Unified data architecture of LoongForge-Embodied. Heterogeneous
  embodied datasets are integrated through a format abstraction layer that
  exposes a common sample contract. Model-specific data processing is
  supported through extensible registries for transformations, preprocessing,
  collation, and sampling, enabling diverse VLA and WAM models to share the
  same data infrastructure.}
  \label{fig:data-architecture}
\end{figure}

\textbf{Format-independent data abstraction.}
Embodied training datasets differ in their storage organization,
metadata conventions, and mechanisms for retrieving multimodal
observations and actions. Supporting these differences through
separate model-specific loaders would couple data storage to model
implementations, limiting the reuse of shared training infrastructure.
LoongForge-Embodied instead introduces a format-independent dataset
abstraction that separates the physical representation of data
from the logical samples consumed during training
(Figure~\ref{fig:data-architecture}).

The framework currently supports LeRobot and HDF5 datasets through
a common data-loading interface, with automatic adaptation across
LeRobot versions 2.0, 2.1, and 3.0. The underlying storage format
is selected independently of the model architecture. Each dataset
implementation is responsible for interpreting its storage-specific
organization and constructing samples according to the shared
representation, while downstream training components operate
without requiring knowledge of the original dataset format.
This separation allows different storage backends to share the
same training infrastructure.

\textbf{Unified training-facing contract.}
To decouple data storage from model execution, LoongForge-Embodied
defines a common sample structure consisting of visual observations
(\texttt{observation.images.*}), robot states
(\texttt{observation.state}), actions (\texttt{action}),
task information (\texttt{task}), and temporal metadata
(\texttt{timestamp}). The contract standardizes how these
information categories are exposed without requiring all models
to use identical temporal windows, action horizons, or observation
configurations.

\subsection{Model-Specific Data Adaptation}
\label{sec:model_specific_data}

A unified sample representation does not imply uniform data
requirements. VLA and WAM models may differ substantially in their
temporal context, sample indexing, observation transformations,
and batching strategies. Directly embedding these differences into
the shared dataset implementation would introduce model-specific
branches and undermine extensibility. LoongForge-Embodied therefore
provides two complementary mechanisms for injecting model-specific
data semantics while preserving the common infrastructure.

\textbf{Registry-based processing extensions.}
The framework maintains three independently extensible registries,
keyed by model type, for transformation builders, preprocessing
and collation components, and sampling strategies. These registries
separate the construction of model-specific data-processing
components from their invocation by the shared training pipeline.
New implementations are discovered automatically through module
discovery, avoiding modifications to the central registration
logic when additional models are integrated.

\textbf{Orthogonal dataset-level adaptation.}
Beyond sample transformations, models may require different
temporal retrieval and indexing semantics. To support these
requirements, the LeRobot dataset abstraction exposes three
independent extension interfaces:
\texttt{delta\_timestamps\_fn},
\texttt{length\_fn}, and
\texttt{index\_map\_fn}.
They respectively control the temporal observations retrieved
for a sample, the logical length of the dataset, and the mapping
from logical sample indices to underlying data indices.
The base dataset implementation contains no model-specific
branching logic, and models override only the interfaces
corresponding to their particular requirements.

For example, FastWAM customizes only temporal retrieval to
construct multi-frame geometric observations, while retaining
the default dataset length and indexing behavior. In contrast,
Wall-OSS modifies the dataset length and index mapping to
implement its train/test partitioning without changing the
temporal retrieval mechanism. These two cases illustrate how
different model requirements can be accommodated independently:
temporal context can be modified without changing dataset
indexing, and the logical sample space can be modified without
changing observation retrieval.

\subsection{Efficient and Reproducible Data Delivery}
\label{sec:data_delivery}

Beyond functional compatibility, a shared data system must
provide sufficient throughput for GPU-intensive training
while preserving the data semantics of the original model
implementations. LoongForge-Embodied addresses these requirements
through coordinated data-loading optimizations, cache validation,
stateful recovery, and reference-equivalence testing.

\textbf{High-throughput data loading.}
The framework combines block-oriented sample dispatch,
prefetching, and host-memory staging to reduce data-loading
overhead and overlap data preparation with model execution.
For video-based observations, a unified decoding interface
supports five video backends, allowing backend-specific
decoding implementations to be selected without changing
the surrounding dataset and training logic. Pinned-memory
handling additionally includes fallback behavior to preserve
compatibility when the preferred memory-staging path is
unavailable.

Repeated preprocessing is reduced through the shared
feature-cache mechanism introduced in
Section~\ref{sec:computation_acceleration}. Caching is a
framework-level capability rather than a model-specific
implementation: it is enabled through configuration without
changes to model code, and any integrated dataset can
precompute and reuse representations through it. The savings
are largest for video-heavy workloads such as DreamZero,
where cache-aware retrieval avoids repeatedly decoding
frames whose required representations have already been
generated. Cache reuse
is guarded by compatibility checks covering the dataset
manifest, preprocessing configurations, and associated
hashes. In particular, changes to operations such as
cropping, resizing, or color jitter are detected rather
than silently reusing representations generated under
different preprocessing settings. This ensures that
performance-oriented caching does not inadvertently
change the intended training inputs.

\textbf{Stateful and deterministic resumption.}
Checkpointing a training run requires more than restoring
model parameters and optimizer states. The data pipeline
must also recover its sample progression and stochastic
processing state; otherwise, restarting a run may change
the subsequent training data even when the model state
has been restored correctly.

LoongForge-Embodied extends checkpoint recovery to the
data subsystem through coordinated restoration of sampler
state, iterator progress, and dataset-level augmentation
random-number-generator states. The sampler maintains a
minimal persistent state and reconstructs sampling behavior
deterministically, while the stateful iterator records
sample-yielding progress. Together with the restored
augmentation state, these mechanisms support continuation
of the data stream from its saved execution state,
rather than restarting sampling from the beginning
of an epoch.

\textbf{Reference-equivalence validation.}
A further challenge of unifying previously independent
training stacks is preventing silent changes to the
samples observed by a model. Such changes can arise
from differences in temporal indexing, preprocessing,
data types, or action representations, even when the
resulting tensors appear structurally compatible.

To address this issue, LoongForge-Embodied treats
existing model-specific data implementations as
executable references and validates the unified
pipeline against their outputs. For X-VLA, the
unified data pipeline has been verified to produce
byte-wise identical training samples to the reference
implementation. This reference-equivalence approach
provides a concrete correctness criterion for model
integration and helps isolate the effects of training
system optimizations from unintended changes in
data preprocessing.

\subsection{Cross-Embodiment Data Synthesis with Ego2Robot}
\label{sec:ego2robot}

The preceding abstractions enable heterogeneous robot
datasets to share a common training pipeline. However,
the availability of robot-specific demonstrations
remains a major constraint for embodied learning.
Human egocentric interaction data provide an additional
source of manipulation experience, but differences in
kinematics, coordinate conventions, and embodiment
prevent such trajectories from being directly used
for robot training.

LoongForge-Embodied re-implements and integrates
Ego2Robot~\cite{ego2robot2026}, a
cross-embodiment data synthesis pipeline that
converts human interaction trajectories into
robot-compatible demonstrations. The pipeline
organizes synthesis into four main stages:
interaction reconstruction, human-to-robot action
alignment, robot retargeting, and trajectory
validation and serialization.

During action alignment, human interaction
representations are transformed into robot-compatible
end-effector targets using geometric alignment
procedures, including virtual-finger alignment
and tool-center-point pose transformations.
The retargeting stage then searches for feasible
robot configurations and selects compatible
candidates through joint evaluation of the
corresponding solutions. A registry-based
embodiment interface supports multiple robot
morphologies, allowing robot-specific retargeting
implementations to be incorporated without
restructuring the shared synthesis pipeline.

The generated trajectories undergo structural
validation before being serialized into the native
LeRobot v3.0 format. Importantly, these outputs
can be consumed directly by the unified data-loading
infrastructure described in
Section~\ref{sec:unified_data}, without an additional
format-conversion stage. Ego2Robot thus extends
the framework beyond the unified consumption of
existing datasets to the generation and integration
of new cross-embodiment training data, forming
a closed data-processing workflow within
LoongForge-Embodied. Experimental results are
reported in Section~\ref{sec:exp_ego2robot}.

\subsection{Quality Assessment of Synthesized Trajectories}
\label{sec:ego2robot_eval}

Synthesized cross-embodiment data has no ground truth to
compare against. Unlike a teleoperated demonstration, a
converted trajectory has no recorded robot execution that
defines the correct answer, and the structural validation of
the preceding stage establishes only that the output is well
formed. Downstream task success measures the data indirectly
and only after a full training run has completed, which is
both too slow and too coarse to guide the pipeline itself. 
A direct measure of synthesis quality is needed, computed on the
pipeline output itself and reported per episode.

LoongForge-Embodied therefore introduces an assessment
standard for synthesized trajectories, organized along three
axes (Table~\ref{tab:ego2robot_rubric}). Kinematic
feasibility asks whether the retargeted joint trajectory is
executable, through four lower-is-better metrics computed
programmatically from the robot state and action streams,
with no model in the loop. Visual quality asks whether the
rendered episode is a faithful picture of the interaction,
covering event correctness, object integrity, residual human
appearance, and spatial consistency. Motion naturalness asks
whether the resulting robot motion reads as purposeful over
time, covering coherence, continuity, rhythm, path
directness, pause behavior, and two-arm coordination. The
three are kept separate because they fail separately: a
trajectory can respect every joint limit and still render as
a video in which a human hand performs the manipulation, and
a visually clean episode can carry joint motion no robot
would execute.

The ten visual and motion metrics are graded on a
three-level ordinal scale on which level~2 is best, by a
multimodal model that inspects sampled evidence frames
together with the corresponding source frames for visual
quality, and continuous motion samples with short raw-frame
windows for motion naturalness. For reporting, the levels
are mapped linearly onto a normalized score in which levels
0, 1, and 2 contribute 0\%, 50\%, and 100\%, so that the
score of a metric is $(0.5\,n_1 + n_2)/(n_0 + n_1 + n_2)$
over the episodes that received a level; episodes the grader
marks unknown or not applicable are excluded from both
numerator and denominator, and the number of scored episodes
is reported alongside the score. The four kinematic metrics
stay in their original units and are macro-averaged over
episodes. They are never folded into the ordinal scale,
because a violation rate and a perceptual grade are not
commensurable quantities.

\begin{table}[!ht]
\centering
\caption{\textbf{Assessment standard for synthesized
trajectories.} The four kinematic metrics are computed
programmatically from the robot state and action streams;
the ten visual and motion metrics are graded on a
three-level ordinal scale by a multimodal model from sampled
evidence. Arrows give the direction of improvement.}
\label{tab:ego2robot_rubric}
\small
\setlength{\tabcolsep}{4pt}
\begin{tabularx}{\textwidth}{@{}l l c >{\raggedright\arraybackslash}X@{}}
\toprule
\textbf{Metric} & \textbf{Unit} & \textbf{Dir.}
 & \textbf{What it measures} \\
\midrule
\multicolumn{4}{@{}l}{\textit{Kinematic feasibility}} \\
Velocity violation rate & \% & $\downarrow$
 & Share of valid joint-velocity samples whose magnitude
 exceeds the reference limit of the target embodiment \\
Acceleration violation rate & \% & $\downarrow$
 & Share of valid joint-acceleration samples exceeding the
 reference limit, with accelerations differenced from the
 velocity samples \\
Trajectory jerk P95 & unit/s$^3$ & $\downarrow$
 & 95th percentile of the unfiltered absolute third
 difference of the joint trajectory, aggregated across
 episodes \\
IK rotation error & $^\circ$ & $\downarrow$
 & Angular deviation between the achieved and the commanded
 end-effector orientation \\
\midrule
\multicolumn{4}{@{}l}{\textit{Visual quality}
 (levels 0--2, 2 best)} \\
Key-event alignment & 0--2 & $\uparrow$
 & Whether the key manipulation events occur with the right
 type, order, and objects, and whether object motion is
 supported by a corresponding robot grasp, transport, or
 release \\
Object fidelity & 0--2 & $\uparrow$
 & Whether salient objects keep their identity, count, and
 appearance against the time-matched source frame, without
 erasure, duplication, or persistent deformation \\
Human-residue severity & 0--2 & $\uparrow$
 & Whether identifiable human limbs remain visible in the
 output frames, and whether such residue occludes the
 manipulation \\
Visual-spatial plausibility & 0--2 & $\uparrow$
 & Whether support, occlusion, depth, and geometric
 relationships hold, without floating objects, surface
 penetration, or impossible nesting \\
\midrule
\multicolumn{4}{@{}l}{\textit{Motion naturalness}
 (levels 0--2, 2 best)} \\
Overall naturalness & 0--2 & $\uparrow$
 & Whether the motion as a whole is coherent, task-relevant,
 and interpretable as pursuing the goal \\
Pose continuity & 0--2 & $\uparrow$
 & Whether gripper and arm poses stay continuous, without
 sudden flips, teleports, or repeated jumps that disturb the
 manipulation \\
Velocity rhythm & 0--2 & $\uparrow$
 & Whether displacement, acceleration, deceleration, and
 pauses match the task phase on the true frame timeline \\
Path plausibility & 0--2 & $\uparrow$
 & Whether the gripper path toward objects and targets is
 direct and effective, without aimless drift, detours, or
 repeated retraction \\
Pause and recovery & 0--2 & $\uparrow$
 & Whether stops, phase transitions, and resumptions are
 smooth, without freeze-then-jump or abrupt restarts \\
Bimanual coordination & 0--2 & $\uparrow$
 & Whether the two arms support, grasp, transport, and
 release in a sensible order when the task requires
 cooperation \\
\bottomrule
\end{tabularx}
\end{table}

\section{Evaluation Framework}
\label{sec:evaluation_framework}

Evaluating heterogeneous VLA and WAM models presents
challenges beyond those encountered during training.
Different benchmarks expose distinct observation formats,
action spaces, control protocols, and runtime dependencies,
while individual models require different input representations
and action execution strategies. Maintaining separate
evaluation implementations for each model--benchmark
combination introduces redundant integration efforts and
risks inconsistencies between training and evaluation.
To address these challenges, LoongForge-Embodied extends
its unified model abstraction to evaluation through a
training-aligned architecture and composable adaptation
mechanisms. This design enables heterogeneous models to
share a common evaluation infrastructure while preserving
model-specific execution behaviors and benchmark-specific
evaluation protocols.

\subsection{Training-Aligned Evaluation Architecture}
\label{sec:eval_architecture}

A central design principle of LoongForge-Embodied is to
reuse the same model inference implementation across
training and evaluation. Rather than maintaining separate
evaluation-specific model implementations, the framework
exposes a unified \texttt{predict\_action} interface that
accepts visual observations, language instructions,
robot states, and dataset statistics. The evaluation
framework invokes this interface through a generic policy
wrapper, allowing model-specific inference logic,
including input processing, state normalization, and
action prediction, to remain within the shared model
implementation. In particular, normalization and
denormalization associated with a model's training
configuration are handled by the model itself rather
than reimplemented in individual benchmark adapters.
This alignment reduces duplicated inference logic
and minimizes the risk of inconsistencies introduced
by independently maintained training and evaluation
implementations.

To accommodate heterogeneous runtime dependencies,
LoongForge-Embodied adopts a decoupled client--server
architecture that separates benchmark execution from
model inference. The benchmark client manages environment
initialization, observations, action execution, and episode
termination, while an independent inference server handles
model loading and prediction. The two processes communicate
through a lightweight WebSocket-based RPC protocol using
\texttt{msgpack-numpy} for structured numerical data
exchange, with protocol-version validation during connection
establishment. This separation allows benchmark environments
and model inference services to operate with independent
software dependencies, avoiding conflicts between simulation
libraries and model-specific deep learning stacks.

\begin{figure}[!ht]
  \centering
  \includegraphics[width=0.7\textwidth]{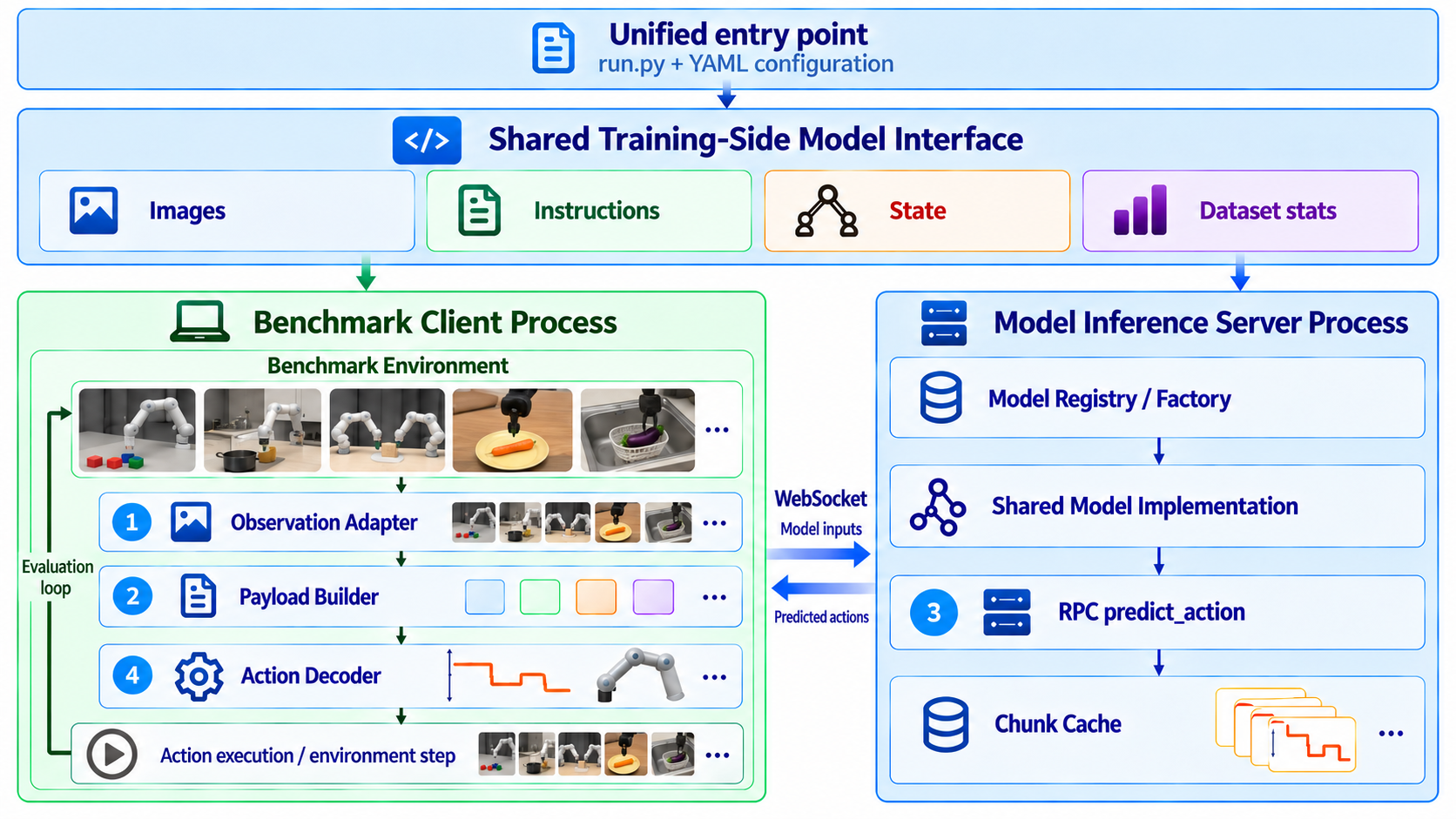}
  \caption{Training-aligned evaluation architecture. A unified entry point
  launches a benchmark client process and a model inference server process,
  which exchange only model inputs and predicted actions over WebSocket RPC.
  Numbered markers denote the four pipeline stages: observation adaptation (1),
  payload construction (2), and action decoding (4) run in the client, whereas
  action prediction through the shared training-side \texttt{predict\_action}
  interface (3) runs in the server, which also hosts the model registry/factory
  and the action-chunk cache. The client loop returns from each environment
  step to observation adaptation until the episode terminates.}
  \label{fig:evaluation_architecture}
\end{figure}

Building on this architecture, the framework organizes
evaluation into a standardized four-stage pipeline.
First, a benchmark-specific observation adapter converts
raw environment observations into a canonical representation.
Second, a model-specific payload builder transforms the
canonical observations into the inputs expected by the
corresponding model. Third, the inference server executes
the shared \texttt{predict\_action} interface through RPC
and returns the predicted actions. Finally, an action
decoder converts the model outputs into the action
representation required by the benchmark environment.
A unified entry point
and configuration interface coordinate these components
across the benchmarks and model implementations
currently integrated into the evaluation framework---five
benchmarks (LIBERO, CALVIN, SimplerEnv, RoboTwin, and
ManiSkill) and seven model implementations---without
requiring dedicated evaluation pipelines for individual
model--benchmark combinations
(Fig.~\ref{fig:evaluation_architecture}). These counts
describe the integration scope of the evaluation framework
and are independent of the models and benchmarks used in
the experiments of Section~\ref{sec:exp}.

\subsection{Composable Model and Benchmark Adaptation}
\label{sec:eval_adaptation}

To support heterogeneous models and environments within
the shared evaluation pipeline, LoongForge-Embodied
introduces registry-based, composable adaptation mechanisms.
Three complementary registries manage model construction,
input payload generation, and action decoding, respectively.
Model factories instantiate the corresponding inference
implementations, while payload builders translate canonical
observations into model-specific inputs according to
declared model capabilities. Action decoding is further
decoupled from individual model--benchmark pairs through
a compositional mapping between model action encodings
and environment action spaces. This separation
allows model and benchmark integrations to evolve
independently and avoids maintaining an explicit adapter
for every supported combination.

Beyond input and output representations, embodied models also differ in
action execution semantics; enforcing a single strategy would either add
redundant inference or alter a model's intended closed-loop behavior.
LoongForge-Embodied therefore supports two execution modes, selected by a
per-model cache switch declared in its payload builder:
\begin{itemize}
  \item \textbf{Open-loop, cached execution.} The client still issues one
  request per environment step, but the server runs model inference only
  when a new action chunk is required and otherwise returns the next action
  from the cached chunk. The model thus does not observe intermediate
  environment states while a chunk is being executed, and redundant
  inference for chunk-based policies is avoided.
  \item \textbf{Closed-loop, uncached execution.} The cache is bypassed, so
  the model is invoked at every environment step with the latest
  observation. This preserves the native behavior of models that condition
  on every observation and maintain their own action queues, such as
  DreamZero and LingBot-VA.
\end{itemize}
Additional model- and benchmark-specific parameters, such as embodiment
identifiers, replanning horizons, and action-step limits, are applied
within the same execution framework, so evaluation orchestration remains
unified without imposing a uniform control policy.

A unified evaluation interface must also preserve the protocols defined by
individual benchmarks so that reported results remain comparable with
official implementations. Benchmarks, however, differ in how much of their
evaluation loop they expose. For benchmarks that provide a programmatic
environment API (LIBERO, CALVIN, SimplerEnv, and ManiSkill), a
\emph{native runner} owns the loop: it creates, resets, and steps the
environment, collects observations, and applies actions itself. Other
benchmarks, such as RoboTwin, ship an official evaluator that owns
environment construction and rollout and exposes only a policy-plugin
callback; observations are delivered to that callback, but no external API
is available for driving the environment. For these benchmarks, a
\emph{bridge} hosts the shared observation adapter, payload builder, policy
client, and action decoder inside the official evaluator as a policy
plugin, so that the official runner drives the environment while model
inference still passes through the same shared pipeline. The two modes
reuse the same adaptation and inference components and differ only in
which side owns the environment loop. To maintain evaluation fidelity, we
validate the protocol-critical components---observation preprocessing,
prompt construction, coordinate and action transformations, initialization
behavior, and episode control rules---against their official counterparts,
and compare the resulting success rates with official reference runs
(Table~\ref{tab:benchmark_sr}). Such checks are particularly important
because seemingly minor implementation differences can substantially affect
robotic task success rates, and they reduce the risk that unified adaptation
introduces unintended changes to benchmark semantics. Completeness and
cross-run reproducibility checks on recorded episodes further guard the
reported results.

\subsection{Evaluation Acceleration}
\label{sec:eval_acceleration}

Evaluation is an essential part of the embodied model training
lifecycle. Reliable performance estimates require repeated trials
across many tasks, so a complete evaluation can consume a substantial
share of the lifecycle. LoongForge-Embodied accelerates evaluation
while preserving the benchmark protocol: replanning intervals, step
limits, planned episode counts per task, and expert-screening criteria
remain unchanged. Only the execution of the evaluation loop is
optimized.

Evaluation time has three main sources: the cost of executing episodes,
including the degree of concurrency; fixed startup costs, such as
starting processes and loading model weights; and work that depends on
the benchmark, task, and seed, but not on the policy being evaluated.
LoongForge-Embodied addresses each source separately.

\begin{itemize}
  \item \textbf{Parallel orchestration.}
  A shared coordinator schedules episodes across
  $S = G \times N$ slots, where $G$ is the number of inference servers
  and $N$ is the number of simulator workers attached to each server.
  Episodes are dispatched in chunks to amortize process startup and
  scene initialization. Units are queued longest-expected-duration
    first; for long-horizon tasks, every chunk is placed ahead of all
    short-task work and split into smaller chunks, so the makespan is
    bounded by the long-task tail. Slots are refilled as soon as a chunk
    completes.
  The capacity of each pool
  depends on the simulator's host resource: CPU-sim pools are limited
  by available CPU cores, while GPU-sim pools share device memory with
  the inference server. For GPU-sim, the coordinator checks the memory
  budget before a run starts. It also limits CPU threads per simulator
  process to prevent oversubscription when many simulators share a
  host. Inference servers remain persistent across chunks, so model
  weights are loaded once per run.

  \item \textbf{Lower per-episode cost.}
  The client holds a predicted action chunk and replans every $K$
  steps, reducing inference requests from one per environment step to
  one per chunk. It also skips camera rendering within a chunk, since
  the model does not consume intermediate observations while that
  chunk executes. On the server, requests from different slots can be
  merged into a batched forward pass. This is enabled only for models
  whose inference cost is a bottleneck: for example, batching brings
  little benefit when the mean batch size remains close to one, but
  can improve throughput when long model requests leave the device
  idle between requests. Disabling per-step traces and video logging
  further avoids serialization and video-encoding work during
  evaluation.

  \item \textbf{Reusable policy-independent work.}
  Some work depends only on the benchmark, task configuration, and
  seed, and can therefore be computed without an inference server and
  reused across policies. For RoboTwin, offline expert screening
  records accepted seeds and task attributes needed to construct the
  scene. For LIBERO, scene snapshots capture the random-number
generator state before each environment reset. Under per-task
seeding the scene depends on how many resets have already occurred
in the process, so snapshots are what allow an episode to run in any
worker; they therefore enable episode-level sharding rather than
merely saving scene-rebuild time. These artifacts are reused
  across later runs with the same task configuration. Their one-time
  generation cost is excluded from the evaluation times reported in
  Table~\ref{tab:eval_speedup}; because the reference path repeats
  expert screening on each run, the RoboTwin precomputation cost is
  recovered from the second run onward.
\end{itemize}

The simulator count $N$ and the server-side batch size share the same
GPU memory budget, so they must be tuned together. Increasing $N$ can
reduce the memory available for batching, and a larger simulator pool
does not necessarily improve throughput once the pool is saturated.
Raising $N$ while shrinking the merge batch size can therefore trigger
  out-of-memory scheduling units and lengthen a run rather than shorten
  it; Section~\ref{sec:eval_throughput} reports a measured instance. The
  degree of concurrency and batching is calibrated for each deployment,
  since throughput saturation depends
on host core count, device memory, model size, and rendering settings.

The protocol is held constant across the reference and accelerated
paths, but individual episode outcomes are not required to match.
Because episodes run concurrently, an unstable scene may be replaced
and the evaluated seed set may consequently change. We therefore
assess the correctness of accelerated evaluation by comparing
aggregate success rates under the same protocol, rather than by
requiring episode-by-episode reproduction (see
Section~\ref{sec:eval_throughput}).

\section{Related Work}
\label{sec:related}

\textbf{Embodied foundation models: VLAs and WAMs.}
Embodied foundation models have developed along two complementary routes. VLA models, reviewed in \cite{vla_tnnls2026}, couple a vision-language backbone with an action decoder and learn from demonstration data, whereas WAMs \cite{wam_next2026, wam_survey2026} learn a generative model of the world from which actions are derived. Within the VLA family, the field has moved from transformer policies over demonstrated trajectories \cite{rt12022, octo2024} and web-scale vision-language transfer \cite{rt22023} toward open VLM-based policies \cite{openvla2024}; what mainly distinguishes the current generation is not scale but training-compute design. ${\pi}_0$ and its successors \cite{pi0, pi05} replace a discrete action head with a flow-matching action expert, whose multi-step denoising adds a distinct computation path to every training iteration; GR00T N1 \cite{groot_n1} decouples a frozen vision-language model from a separate action-generation module, changing which components receive gradients; and RDT2 \cite{rdt2_2026} scales a 7B VLM with flow matching for cross-embodiment training. WAMs are even more heterogeneous in their generative machinery: GR-2 \cite{gr2_2024} conditions action generation on predicted future video, unified world models \cite{uwm2025} couple video and action diffusion within a single denoising process, Cosmos Policy \cite{cosmos_policy2026} transfers a video foundation model to control and planning, and recent systems \cite{dreamzero2026, lingbotva2026, fastwam2026} explore diverse world-action couplings, from joint generation of world and actions to world prediction as a training auxiliary, each with a different generative computation graph and memory profile. These architectural choices translate directly into heterogeneous training workloads: attention patterns, computation graphs, memory footprints, and training objectives (behavior cloning versus multi-step denoising) vary across models, so no single optimization recipe applies uniformly and the training infrastructure must accommodate this diversity.

\textbf{Embodied training frameworks.}
Beyond model architectures, the community has built infrastructure that unifies model development, data processing, training, and evaluation. General-purpose frameworks standardize training pipelines: LeRobot \cite{lerobot2024} integrates data collection, storage, training, and inference for end-to-end robot learning, building on the modular imitation-learning foundations of robomimic \cite{robomimic2021}; RLinf-VLA \cite{rlinfvla2026} unifies VLA training with reinforcement-learning algorithms and simulators in a single framework; and VLA Foundry \cite{vlafoundry2026} and FluxVLA Engine \cite{fluxvla2026} extend this unification to the full LLM-to-VLA training stack and the data-to-deployment workflow, respectively. Modular frameworks expose composable components: StarVLA \cite{starvla2026} provides pluggable VLM backbones and action heads in a Lego-like design, StarVLA-$\alpha$ \cite{starvla_alpha2026} reduces design complexity for reproducible training and comparison, AlphaBrain \cite{alphabrain2026} extends StarVLA with world-model backbones and reinforcement-learning fine-tuning, and OpenWAM \cite{openwam2026} modularizes backbone, representation, and video-action coupling for WAM pretraining. Alongside these, model-family-specific pipelines such as openpi \cite{pi05} and Isaac-GR00T \cite{groot_n1} provide deeply integrated training, fine-tuning, and deployment for their respective model series. Data and evaluation ecosystems complement these frameworks: Open X-Embodiment \cite{openx2023} and DROID \cite{droid2024} aggregate large-scale heterogeneous demonstrations, and LIBERO \cite{libero2023}, SimplerEnv \cite{simplerenv2024}, and RoboTwin \cite{robotwin2024} provide standardized simulation benchmarks. In sum, existing frameworks have achieved substantial functional unification of model interfaces, data pipelines, and evaluation flows, and some offer specialized training efficiency within specific model families or training paradigms. However, they primarily emphasize functional modularity or efficiency within such specific scopes, leaving systematic reuse of architecture-aware optimizations across heterogeneous VLA and WAM workloads insufficiently explored; this is precisely the abstraction level at which our system operates.

\textbf{Training acceleration and distributed systems.}
The deep learning systems community has produced a rich, mature body of optimization techniques covering compute, memory, and communication, organized in system-level surveys \cite{efftrain2026, disttrain_survey2023}. At the distributed level, Megatron-LM \cite{megatronlm2019} introduced tensor and model parallelism, ZeRO \cite{zero2020} shards optimizer states, gradients, and parameters, PyTorch FSDP \cite{fsdp_paper2023} provides PyTorch-native sharded data parallel training at scale, and TorchTitan \cite{torchtitan2025} unifies these PyTorch-native techniques in a single modular system. At the kernel level, FlashAttention \cite{flashattn2022, flashattn2_2023, flashattn3} and its latest successor \cite{flashattn4_2026} make attention IO-aware and parallel, FlexAttention \cite{dong2024flex} compiles programmable attention patterns, and Triton \cite{tillet2019triton}, \texttt{torch.compile} \cite{torch2_2023}, and CUDA Graphs \cite{cudagraphs} address operator fusion, graph compilation, and launch overhead. Memory and precision are addressed by gradient checkpointing \cite{chen2016training}, mixed precision \cite{mixedprecision2018}, and FP8 formats \cite{fp8formats2022, fp8lm2023}, while communication optimization includes gradient quantization and compression \cite{qsgd2017}, error-feedback mechanisms \cite{errorfeedback2019}, ZeRO++'s quantized collective communication \cite{zeropp2023}, and planning and runtime adaptation for collective operations \cite{collcomm2026}. While existing systems provide powerful optimization primitives and composable training mechanisms, adapting and coordinating these optimizations across heterogeneous VLA and WAM architectures remains a model-dependent engineering effort. TorchTitan \cite{torchtitan2025} is the closest system-level work to ours in spirit: it unifies PyTorch-native training techniques into a modular system whose components compose, but it is designed around the homogeneous structure of decoder-only language models, and its optimization interface does not couple to architecture-specific characteristics of multimodal embodied models. Our contribution is therefore not a new optimization primitive but an abstraction layer that adapts, composes, and reuses existing optimizations across VLA and WAM workloads, making architecture-aware acceleration a reusable capability rather than per-model engineering.

\section{Experiments}
\label{sec:exp}
\subsection{Setup}
\label{sec:setup}

\paragraph{Models and reference implementations.}
We evaluate representative embodied foundation models spanning the model
families discussed in Section~\ref{sec:model_specific}: the VLAs
$\pi_{0.5}$~\cite{pi05}, GR00T N1.6 and N1.7~\cite{groot_n1},
X-VLA~\cite{xvla}, and Wall-OSS-0.5~\cite{walloss05}; the WAMs FastWAM~\cite{fastwam2026},
LingBot-VA~\cite{lingbotva2026}, and DreamZero~\cite{dreamzero2026}; and the
omnimodal world model Cosmos 3~\cite{cosmos3}. End-to-end speedups are reported
for all nine models. Each model is compared
against its corresponding publicly released training implementation, namely
the LeRobot PyTorch implementation~\cite{lerobot2024} for $\pi_{0.5}$, the
official Isaac-GR00T implementation for GR00T, and the official or community
releases for the remaining models. Each comparison uses the same model
architecture and training data. When a public release does not ship a
configuration that matches the evaluated setting, we adapt it minimally,
keep its training code path, and note this in the corresponding appendix
(for DreamZero, Appendix~\ref{app:dreamzero}).

\paragraph{Benchmarks.}
We evaluate downstream task performance on three simulation benchmarks.
LIBERO~\cite{libero2023} comprises four task suites, Spatial, Object, Goal,
and Long, which respectively probe spatial reasoning, object generalization,
goal-conditioned behavior, and long-horizon manipulation.
RoboTwin~\cite{robotwin2024} evaluates bimanual manipulation under Easy and
Hard settings. SimplerEnv~\cite{simplerenv2024} evaluates policies in
simulated counterparts of the real-robot Bridge and Fractal evaluation setups.
For each benchmark, we follow the official task definitions and evaluation
protocols, including episode counts and action-execution settings, and report
success rates in percent. Each model is evaluated on the benchmarks for which
its community release provides the training data and hyperparameter
configurations required for reproduction.

\subsection{End-to-End Training Speedup}
\label{sec:e2e_speedup}

\begin{figure*}[!ht]
    \centering
    \includegraphics[width=0.9\textwidth]{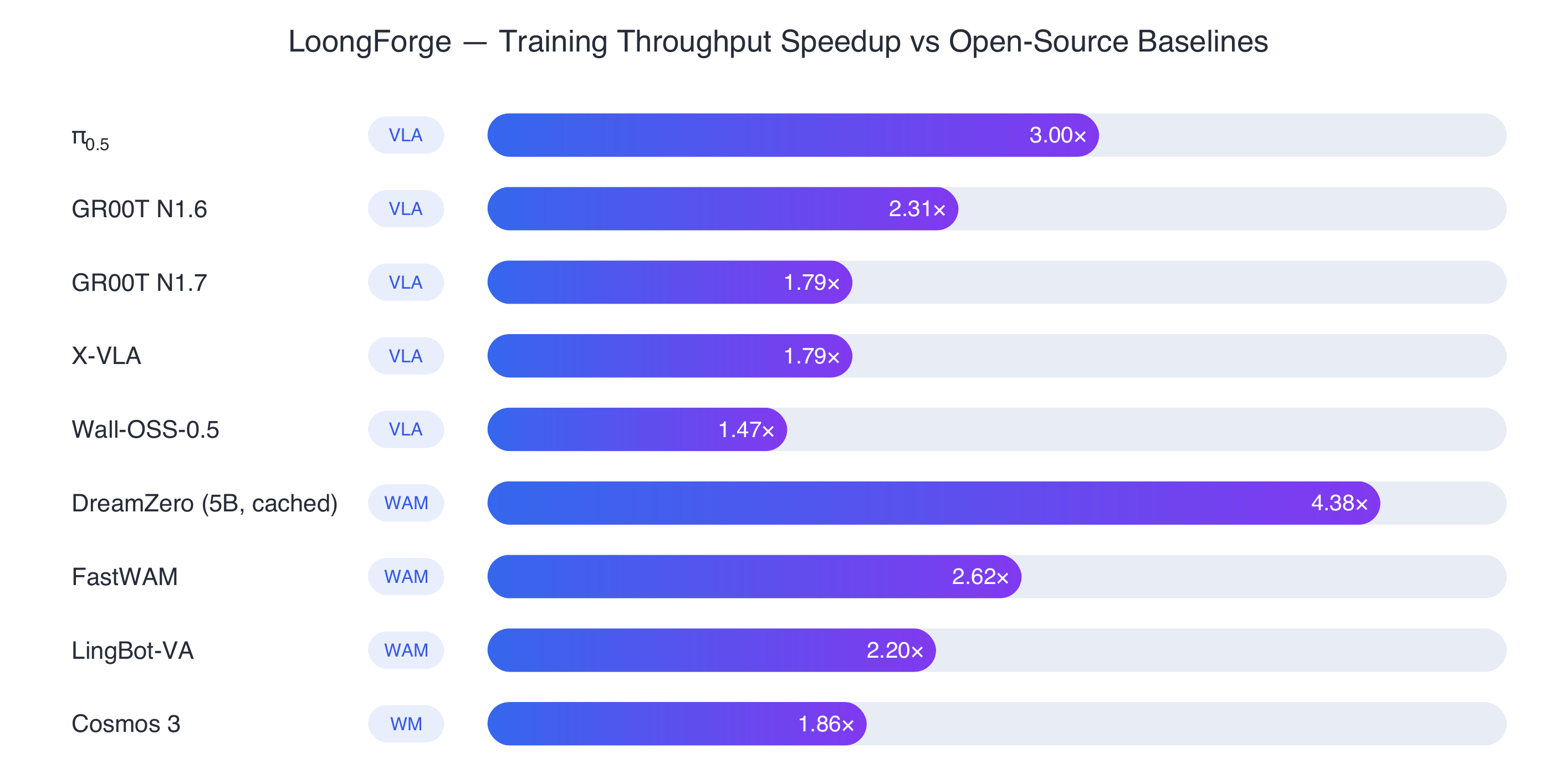}
    \caption{End-to-end training throughput speedup
    of LoongForge-Embodied over the corresponding
    open-source implementations across nine
    embodied foundation models. Higher is better.
    DreamZero is measured with the Wan2.2-TI2V-5B backbone and
    precomputed features on eight GPUs, excluding
    feature-preparation time.}
    \label{fig:training_speedup}
\end{figure*}

We evaluate the end-to-end training efficiency of
LoongForge-Embodied across nine representative embodied
foundation models, covering diverse VLA and WAM
architectures. For each model, we compare the training
throughput of LoongForge-Embodied against its corresponding
open-source implementation and report the resulting
speedup, as shown in Figure~\ref{fig:training_speedup}.

LoongForge-Embodied achieves substantial training
acceleration across all evaluated models, with speedups
ranging from 1.47$\times$ to 4.38$\times$.
Specifically, it achieves 3.00$\times$ on $\pi_{0.5}$,
2.31$\times$ on GR00T N1.6, 1.79$\times$ on GR00T N1.7,
1.79$\times$ on X-VLA, and 1.47$\times$ on Wall-OSS-0.5.
For DreamZero (Wan2.2-TI2V-5B), FastWAM, and LingBot-VA, the framework
delivers speedups of 4.38$\times$, 2.62$\times$,
and 2.20$\times$, respectively. For the omnimodal
world model Cosmos 3, the speedup is 1.86$\times$.
Among these models, DreamZero exhibits the largest
improvement, reaching a 4.38$\times$ increase in
end-to-end training throughput.

These results demonstrate that LoongForge-Embodied
consistently improves training efficiency across
heterogeneous embodied models, despite substantial
differences in their architectures and computational
characteristics. More importantly, the improvements
are achieved at the end-to-end training level,
demonstrating that the combination of reusable
system-level optimizations and model-specific
optimization strategies translates into substantial
practical training acceleration.

\subsection{Training Fidelity and Benchmark Performance}
\label{sec:training_fidelity}

\begin{figure}[!t]
    \centering
    \includegraphics[width=0.98\textwidth]{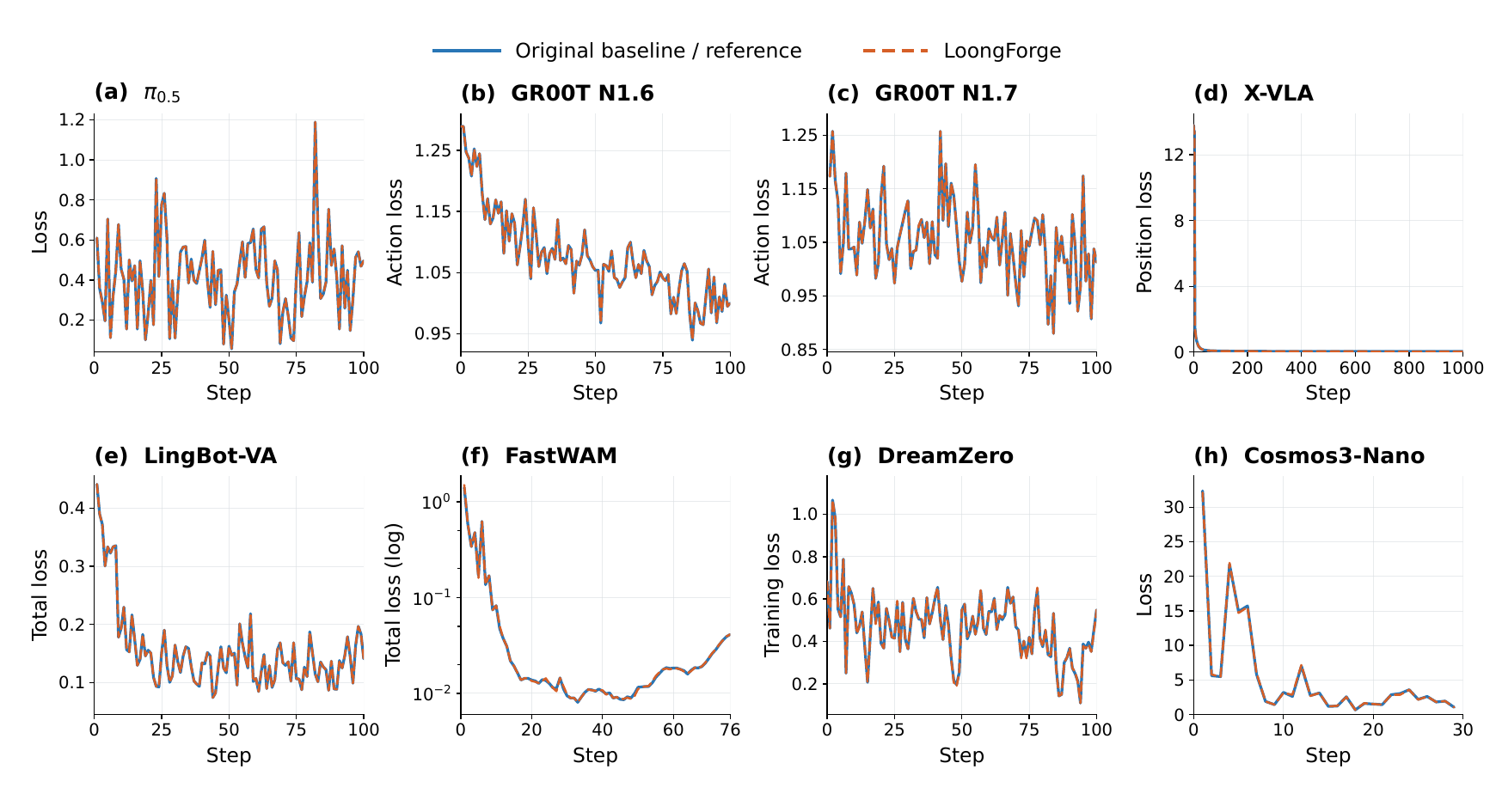}
    \caption{Training-loss alignment between the original reference implementations
    and LoongForge-Embodied across eight embodied models. The solid blue
    and dashed orange curves denote the reference and LoongForge-Embodied,
    respectively. Each model retains its original loss metric and plotted
    training-step interval; FastWAM uses a logarithmic loss axis.}
    \label{fig:loss_alignment_eight_models}
\end{figure}

\textbf{Training fidelity.}
We first investigate whether the proposed training acceleration preserves the
optimization behavior of the original model implementations. For each paired
experiment, the official baseline and LoongForge-Embodied use the same model
initialization, training data, global batch size, optimizer configuration, and
learning-rate schedule. We compare training losses at matched optimizer steps,
rather than wall-clock times, to distinguish optimization fidelity from
throughput improvements. As shown in Fig.~\ref{fig:loss_alignment_eight_models},
the training loss curves of LoongForge-Embodied closely overlap those of the
corresponding baselines across the eight models shown there. The accelerated
implementations reproduce both the rapid initial loss reduction observed in
several models and the step-wise fluctuations of their reference
implementations, with closely matched loss values at the end of the respective
evaluation windows. These results indicate that LoongForge-Embodied preserves
the observed optimization dynamics across heterogeneous model architectures
without introducing noticeable systematic deviations in training loss. This
comparison is visual and limited to the displayed training windows; it does
not by itself establish full-run convergence or numerically identical
results. Quantitative short-window fidelity checks for individual models are
reported in Section~\ref{sec:dist_comm} and Appendix~\ref{app:model_analysis}.

\textbf{Benchmark performance.}
To assess whether training acceleration preserves downstream task performance, we independently train and evaluate representative models on LIBERO, RoboTwin, and SimplerEnv, following the respective benchmarks' task definitions and evaluation protocols, including episode counts and action-execution settings. Table~\ref{tab:benchmark_sr} summarizes the resulting success rates across different models and benchmarks. All baseline results are obtained with the corresponding official implementations and their evaluation protocols. For $\pi_{0.5}$, we follow the LeRobot implementation~\cite{lerobot2024} and its evaluation protocol, whose RoboTwin reference result covers 32 tasks; on this aligned task set LoongForge-Embodied reaches 89.71\%, whereas the RoboTwin entry in the table reports its average over the 50 tasks more commonly adopted by the community. Across the evaluated settings, LoongForge-Embodied achieves success rates comparable to or exceeding the corresponding open-source baselines, while retaining substantial training speedups. These results demonstrate that our framework can significantly improve training efficiency without compromising downstream task performance.

\begin{table*}[!t]
\centering
\caption{Benchmark success rates (\%) of LoongForge-Embodied
and the corresponding open-source baselines across LIBERO,
RoboTwin, and SimplerEnv. Bold values indicate improvements
over the corresponding baseline.}
\label{tab:benchmark_sr}
\small
\renewcommand{\arraystretch}{1.10}

% (a) pi_0.5
\begin{subtable}[t]{0.48\textwidth}
\centering
\begin{tabular*}{\linewidth}{
@{\extracolsep{\fill}}llcc@{}}
\toprule
\textbf{Benchmark} & \textbf{Suite} &
\textbf{Baseline} & \textbf{LoongForge} \\
\midrule
\multirow{5}{*}{LIBERO}
& Spatial & 97.0 & \textbf{99.6} \\
& Object  & 99.0 & \textbf{99.6} \\
& Goal    & 98.0 & \textbf{98.8} \\
& Long    & 96.0 & \textbf{97.0} \\
& Avg.    & 97.50 & \textbf{98.75} \\
\midrule
RoboTwin & Easy & 67.91 & \textbf{77.56} \\
\bottomrule
\end{tabular*}
\caption{$\pi_{0.5}$}
\label{tab:pi05_sr}
\end{subtable}
\hfill
% (b) GR00T N1.7
\begin{subtable}[t]{0.48\textwidth}
\centering
\begin{tabular*}{\linewidth}{
@{\extracolsep{\fill}}llcc@{}}
\toprule
\textbf{Benchmark} & \textbf{Suite} &
\textbf{Baseline} & \textbf{LoongForge} \\
\midrule
\multirow{5}{*}{LIBERO}
& Spatial & 97.65 & \textbf{98.0} \\
& Object  & 98.45 & \textbf{98.5} \\
& Goal    & 97.50 & 96.5 \\
& Long    & 94.35 & \textbf{95.5} \\
& Avg.    & 96.99 & \textbf{97.13} \\
\midrule
\multirow{2}{*}{SimplerEnv}
& Bridge  & 62.3 & \textbf{62.6} \\
& Fractal & 72.5 & 72.1 \\
\bottomrule
\end{tabular*}
\caption{GR00T N1.7}
\label{tab:groot_sr}
\end{subtable}

\vspace{0.8em}

% (c) FastWAM
\begin{subtable}[t]{0.48\textwidth}
\centering
\begin{tabular*}{\linewidth}{
@{\extracolsep{\fill}}llcc@{}}
\toprule
\textbf{Benchmark} & \textbf{Suite} &
\textbf{Baseline} & \textbf{LoongForge} \\
\midrule
\multirow{5}{*}{LIBERO}
& Spatial & 98.2 & 97.0 \\
& Object  & 99.2 & \textbf{99.6} \\
& Goal    & 97.8 & 97.2 \\
& Long    & 95.8 & 94.2 \\
& Avg.    & 97.75 & 97.00 \\
\bottomrule
\end{tabular*}
\caption{FastWAM}
\label{tab:fastwam_sr}
\end{subtable}
\hfill
% (d) LingBot-VA
\begin{subtable}[t]{0.48\textwidth}
\centering
\begin{tabular*}{\linewidth}{
@{\extracolsep{\fill}}llcc@{}}
\toprule
\textbf{Benchmark} & \textbf{Suite} &
\textbf{Baseline} & \textbf{LoongForge} \\
\midrule
LIBERO & Long & 78.8 & \textbf{84.6} \\
\midrule
\multirow{2}{*}{RoboTwin}
& Easy & 86.1 & \textbf{87.6} \\
& Hard & 85.1 & \textbf{85.7} \\
\bottomrule
\end{tabular*}
\caption{LingBot-VA}
\label{tab:lingbot_sr}
\end{subtable}

\end{table*}

\subsection{Evaluation Acceleration}
\label{sec:eval_throughput}

We measure the evaluation acceleration of
Section~\ref{sec:eval_acceleration} on six benchmark runs, using an
eight-GPU configuration throughout. The reference path runs eight
independent client--simulator pairs, one per GPU, without the shared
coordinator of Section~\ref{sec:eval_acceleration}; the accelerated
path drives the same eight GPUs through a larger slot pool. Both paths
use the same checkpoint, the same benchmark protocol, and the same
renderer. A full RoboTwin run of 5,000 episodes drops from 41.03 to
1.77 hours with $\pi_{0.5}$ and from 28.6 to 1.93 hours with FastWAM
(clean configuration), and a full LIBERO run of 2,000 episodes drops
from 173.7 to 16.3 minutes with FastWAM under software rendering,
giving end-to-end speedups of 23.2$\times$, 14.8$\times$, and
10.6$\times$. Across all six runs the end-to-end speedup ranges from
4.1$\times$ to 23.2$\times$ (Table~\ref{tab:eval_speedup}). Under the
steady-state convention defined below, which excludes model loading on
both paths, the range is 4.8$\times$ to 24.6$\times$. The runs exercise
the same mechanisms under different bottlenecks, simulator pools, and
inference costs, and are not averaged.

\begin{table}[!ht]
\centering
\caption{Evaluation wall-clock time of the reference and accelerated
execution paths under two reporting conventions. End-to-end spans
coordinator start through the last recorded episode; steady state
excludes model loading and health checks on both paths, with the
reference-path deduction taken at the magnitude measured on the
accelerated path. One-time offline precomputation is excluded. The
renderer column gives the LIBERO rendering backend, software (OSMesa)
or hardware-accelerated (EGL), which is held identical across the two
paths of a row. The six runs exercise the same mechanisms under
different bottlenecks and are not averaged.}
\label{tab:eval_speedup}
\footnotesize
\setlength{\tabcolsep}{3pt}
\begin{tabular}{@{}lllrrrrrrr@{}}
\toprule
 & & & & \multicolumn{2}{c}{\textbf{Reference}}
 & \multicolumn{2}{c}{\textbf{Accelerated}}
 & \multicolumn{2}{c}{\textbf{Speedup}} \\
\cmidrule(lr){5-6}\cmidrule(lr){7-8}\cmidrule(lr){9-10}
\textbf{Pool} & \textbf{Benchmark / model} & \textbf{Renderer}
 & \textbf{Episodes}
 & E2E & Steady & E2E & Steady & E2E & Steady \\
\midrule
Host CPU & LIBERO / FastWAM & OSMesa & 2{,}000 & 173.7\,min & 170.5\,min
 & 16.3\,min & 13.2\,min & 10.6$\times$ & 12.9$\times$ \\
GPU & LIBERO / FastWAM & EGL & 2{,}000 & 73.9\,min & 70.8\,min
 & 17.8\,min & 14.7\,min & 4.1$\times$ & 4.8$\times$ \\
GPU & LIBERO / $\pi_{0.5}$ & EGL & 2{,}000 & 80.2\,min & 76.3\,min
 & 16.3\,min & 12.6\,min & 4.9$\times$ & 6.1$\times$ \\
GPU & RoboTwin / $\pi_{0.5}$ & --- & 5{,}000 & 41.03\,h & 41.0\,h
 & 1.77\,h & 1.668\,h & 23.2$\times$ & 24.6$\times$ \\
GPU & RoboTwin (clean) / FastWAM & --- & 5{,}000 & 28.6\,h & 28.6\,h
 & 1.93\,h & 1.84\,h & 14.8$\times$ & 15.5$\times$ \\
GPU & RoboTwin (random) / FastWAM & --- & 5{,}000 & 32.0\,h & 32.0\,h
 & 2.36\,h & 2.29\,h & 13.5$\times$ & 14.0$\times$ \\
\bottomrule
\end{tabular}
\end{table}

\textbf{Reporting conventions.}
End-to-end time spans coordinator start, which includes loading the
model into each inference server and the associated health checks,
through the last episode being written. Steady-state time excludes
model loading and health checks on both paths and compares rollout
only. Model loading is measured on the accelerated path at 187 to
222\,s for the LIBERO runs and approximately 0.05\,h for the RoboTwin
runs; on the reference path it was not timed separately and is deducted
at the same magnitude, which corresponds to 1.8 to 5\% of the LIBERO
reference times and below 0.2\% of the RoboTwin reference times. The
share of fixed startup
cost shrinks as runs grow, so short runs show lower end-to-end speedups
than full ones: model loading of roughly 190\,s is about 19\% of the
16.3\,min LIBERO run but under 3\% of the 1.9\,h RoboTwin run.
Counting worker startup as well, the fixed cost incurred before the
first episode is 0.09 to 0.10\,h, or 5 to 6\% of the RoboTwin runs.
The two RoboTwin reference times for FastWAM are derived from one
10,000-episode reference run by taking the halves of its episodes that
correspond to the two evaluated task configurations, at a measured
94\% slot occupancy.
The reported times exclude the one-time offline precomputation
discussed at the end of this subsection.

\textbf{Task performance.}
Table~\ref{tab:eval_fidelity} reports the aggregate success rate of the
same checkpoint on the two execution paths. The protocol is unchanged
across paths: replanning interval, step limit, per-task episode count,
and expert-screening criterion are those of the reference path, and
what differs is how episodes are distributed and how much redundant
work each performs. The comparison is on aggregate success rate rather
than per episode, because simulation on GPUs is not bit-reproducible
and the evaluated seed set can shift, as discussed in
Section~\ref{sec:eval_acceleration}.

\begin{table}[!ht]
\centering
\caption{Aggregate success rate (\%) of the same checkpoint on the
reference and accelerated execution paths, for the runs of
Table~\ref{tab:eval_speedup}.}
\label{tab:eval_fidelity}
\small
\setlength{\tabcolsep}{8pt}
\begin{tabular}{@{}lcccc@{}}
\toprule
\textbf{Benchmark / model} & \textbf{Renderer} & \textbf{Reference}
 & \textbf{Accelerated} & \textbf{$\Delta$} \\
\midrule
LIBERO / FastWAM & OSMesa & 97.00 & 96.80 & $-0.20$ \\
LIBERO / FastWAM & EGL & 96.75 & 97.40 & $+0.65$ \\
LIBERO / $\pi_{0.5}$ & EGL & 98.60 & 98.40 & $-0.20$ \\
RoboTwin / $\pi_{0.5}$ & --- & 86.26 & 86.62 & $+0.36$ \\
RoboTwin (clean) / FastWAM & --- & 91.88 & 92.10 & $+0.22$ \\
RoboTwin (random) / FastWAM & --- & 91.78 & 92.12 & $+0.34$ \\
\bottomrule
\end{tabular}
\end{table}

\textbf{Where the remaining cost lies.}
Both pool types saturate well below the nominal capacity of the
hardware. On the host-CPU pool, steady-state throughput rises from 70.7
to 81.9 episodes per minute as $N$ increases from 16 to 80, with the
plateau beginning near $N{=}48$. On the GPU pool with $\pi_{0.5}$,
throughput is 13.47, 14.05, and 12.61 episodes per minute at $N{=}8$,
$10$, and $12$, falling at the largest setting under memory pressure
and contention. These sweeps select a configuration and their absolute
rates are not comparable with the run throughput of
Table~\ref{tab:eval_speedup}. That saturation occurs this early
indicates that the remaining bound is the serial cost of an individual
episode rather than insufficient parallelism. When the model is the
bottleneck, adding simulator instances does not help and merging
requests does: for $\pi_{0.5}$ the mean server-side batch size stays
close to one and merging yields no gain, whereas the video diffusion
transformer of FastWAM takes 810\,ms per request and leaves the device
idle between requests, so merging raises inference throughput by
roughly 3$\times$ at a measured mean batch size of 2.5 to 3.0. The two
knobs draw on the same device-memory budget: on the same sweep
workload, raising $N$ from 8 to 10 while halving the merge batch size
produced 22 out-of-memory scheduling units and increased the time from
43 to 58\,min.

\textbf{Offline precomputation.}
The times in Table~\ref{tab:eval_speedup} exclude offline
precomputation, whose cost is one-time and independent of the policy
being evaluated. Screening 50 RoboTwin tasks at 100 episodes each
across four shards takes approximately 1.0\,h and produces 1.2\,MB of
accepted-seed and task-attribute records; generating 50 scene snapshots
for each of 40 LIBERO tasks takes 245\,s. The artifacts are reused
across policies, and the same RoboTwin screening artifacts serve both
the $\pi_{0.5}$ and the clean FastWAM runs in
Table~\ref{tab:eval_speedup}. On the reference path the expert
screening is repeated in every run, so the one-time cost is recovered
from the second run onward. Further sweeps and timing arithmetic are
given in Appendix~\ref{app:eval_accel}.

\subsection{Cross-Embodiment Data Synthesis}
\label{sec:exp_ego2robot}

Beyond training efficiency, we evaluate whether our implementation of the Ego2Robot pipeline~\cite{ego2robot2026} (Section~\ref{sec:ego2robot}) produces robot data that can be consumed by the unified training stack and benefit downstream policies.

\begin{wraptable}{R}{0.42\textwidth}
\vspace{1.6em}
\centering
\caption{\textbf{RoboTwin success rate of $\pi_{0.5}$ with Ego2Robot data.}
Success rates are aggregated over 50 RoboTwin tasks.}
\label{tab:ego2robot_robotwin}
\small
\setlength{\tabcolsep}{5pt}
\begin{tabular}{@{}lc@{}}
\toprule
\textbf{Version} & \textbf{Success rate (\%)} \\
\midrule
Baseline (with clean data) & 75.76 \\
LoongForge (with Ego data) & \textbf{77.56} \\
\bottomrule
\end{tabular}
\vspace{-0.2em}
\end{wraptable}

\textbf{Downstream policy performance.}
We train $\pi_{0.5}$ on robot data synthesized by Ego2Robot from human egocentric videos together with RoboTwin data, and compare it with a baseline trained on RoboTwin data alone under the same evaluation protocol.

As shown in Table~\ref{tab:ego2robot_robotwin}, incorporating the synthesized data increases the aggregate success rate over 50 RoboTwin tasks from 75.76\% to 77.56\% (+1.80 points). Although modest, this improvement indicates that Ego2Robot outputs can be integrated directly into the training pipeline and provide useful additional supervision for downstream manipulation.

\begin{figure}[!t]
\centering
\captionsetup[subfigure]{font=footnotesize,skip=1pt}
\begin{subfigure}[t]{0.245\textwidth}\centering
\includegraphics[width=\linewidth]{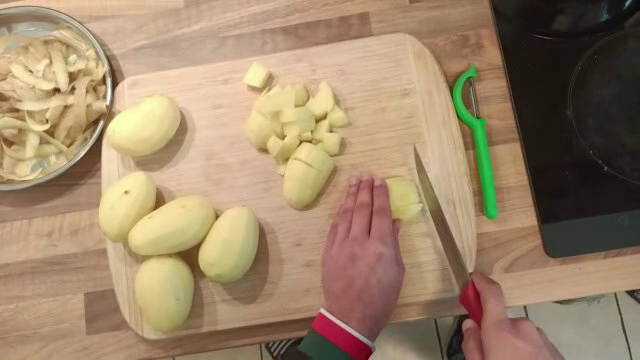}
\caption{Human}
\end{subfigure}\hfill
\begin{subfigure}[t]{0.245\textwidth}\centering
\includegraphics[width=\linewidth]{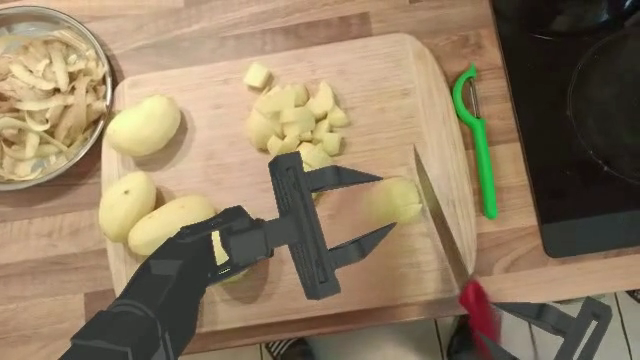}
\caption{ALOHA 2}
\end{subfigure}\hfill
\begin{subfigure}[t]{0.245\textwidth}\centering
\includegraphics[width=\linewidth]{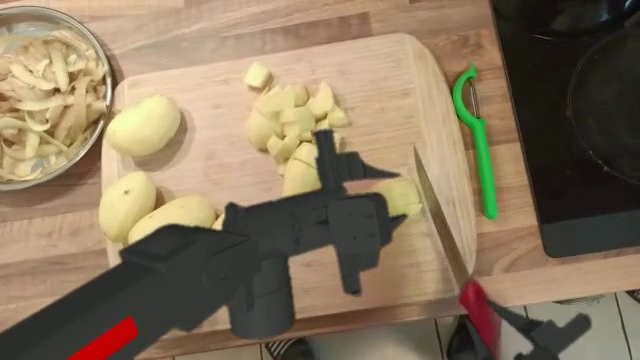}
\caption{ARX L5}
\end{subfigure}\hfill
\begin{subfigure}[t]{0.245\textwidth}\centering
\includegraphics[width=\linewidth]{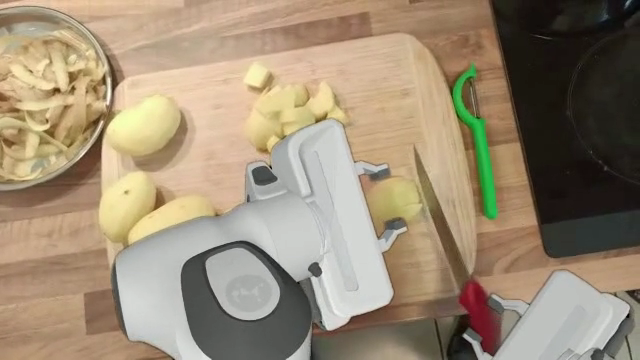}
\caption{Franka Research 3 (FR3)}
\end{subfigure}\\[3pt]
\begin{subfigure}[t]{0.245\textwidth}\centering
\includegraphics[width=\linewidth]{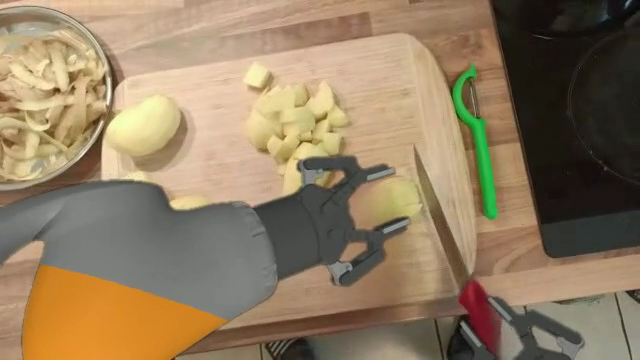}
\caption{KUKA LBR iiwa 14}
\end{subfigure}\hfill
\begin{subfigure}[t]{0.245\textwidth}\centering
\includegraphics[width=\linewidth]{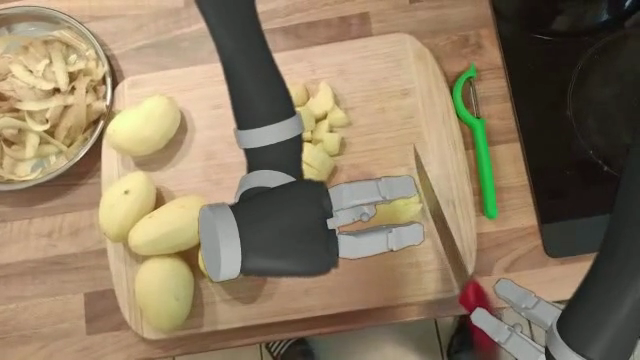}
\caption{Kinova Jaco}
\end{subfigure}\hfill
\begin{subfigure}[t]{0.245\textwidth}\centering
\includegraphics[width=\linewidth]{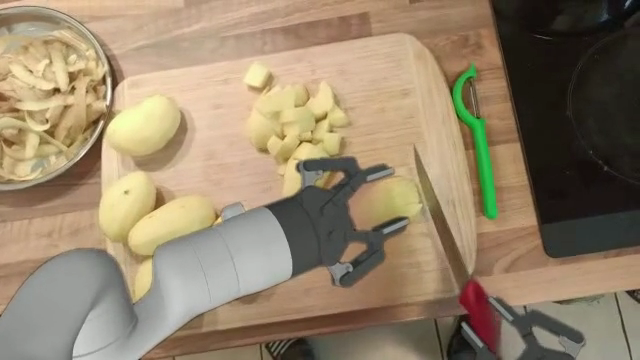}
\caption{Kinova Gen3}
\end{subfigure}\hfill
\begin{subfigure}[t]{0.245\textwidth}\centering
\includegraphics[width=\linewidth]{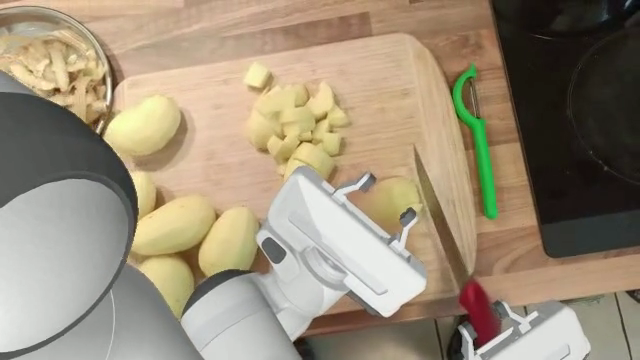}
\caption{Franka Emika Panda}
\end{subfigure}\\[3pt]
\begin{subfigure}[t]{0.245\textwidth}\centering
\includegraphics[width=\linewidth]{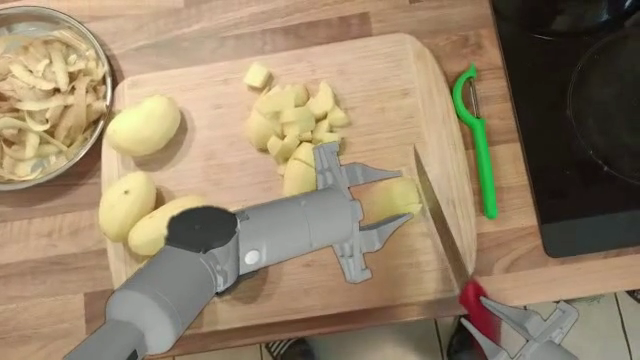}
\caption{AgileX PiPER}
\end{subfigure}\hfill
\begin{subfigure}[t]{0.245\textwidth}\centering
\includegraphics[width=\linewidth]{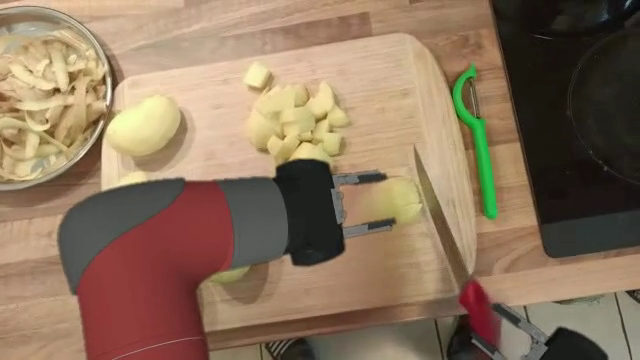}
\caption{Rethink Robotics Sawyer}
\end{subfigure}\hfill
\begin{subfigure}[t]{0.245\textwidth}\centering
\includegraphics[width=\linewidth]{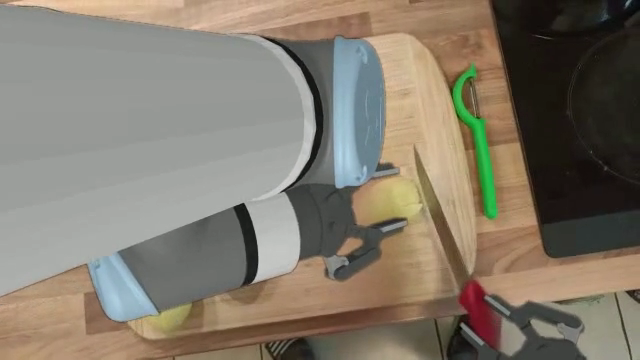}
\caption{UR10e}
\end{subfigure}\hfill
\begin{subfigure}[t]{0.245\textwidth}\centering
\includegraphics[width=\linewidth]{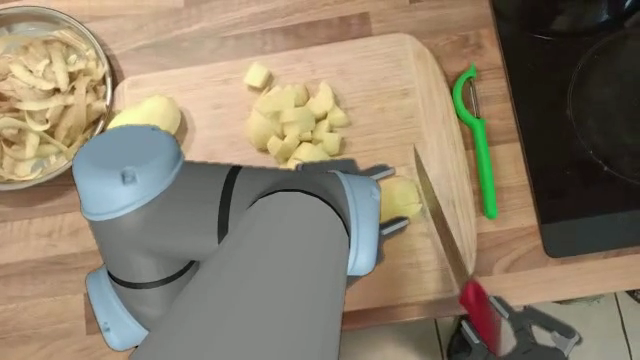}
\caption{UR5e}
\end{subfigure}\\[3pt]
\begin{subfigure}[t]{0.245\textwidth}\centering
\includegraphics[width=\linewidth]{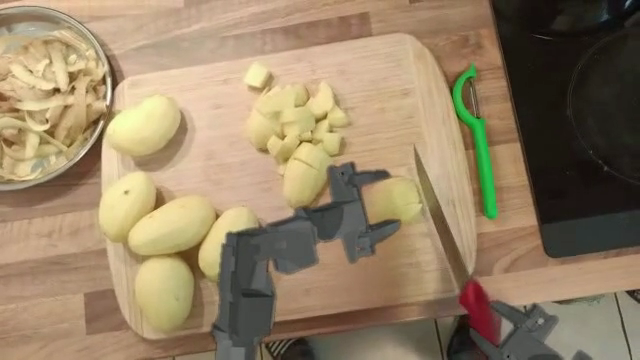}
\caption{WidowX-250 6DOF}
\end{subfigure}\hfill
\begin{subfigure}[t]{0.245\textwidth}\centering
\includegraphics[width=\linewidth]{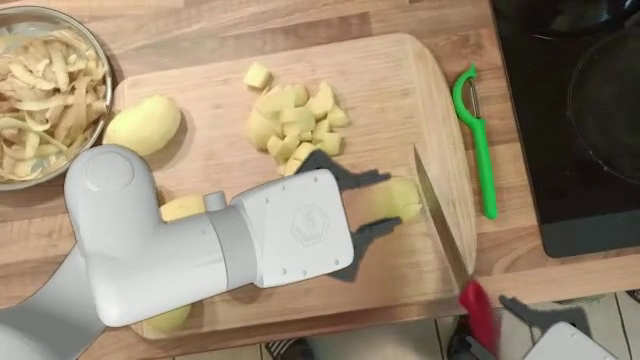}
\caption{UFACTORY xArm 7}
\end{subfigure}\hfill
\begin{subfigure}[t]{0.245\textwidth}\centering
\includegraphics[width=\linewidth]{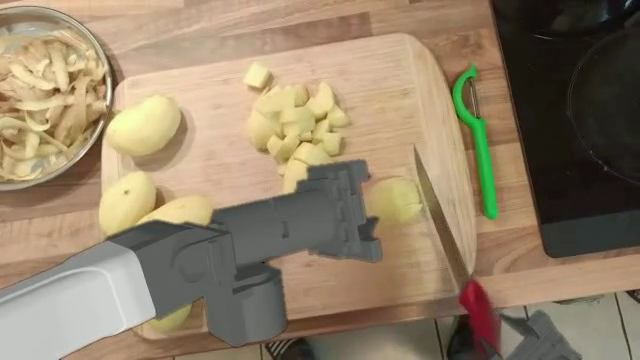}
\caption{I2RT YAM}
\end{subfigure}\hfill
\begin{subfigure}[t]{0.245\textwidth}\centering
\includegraphics[width=\linewidth]{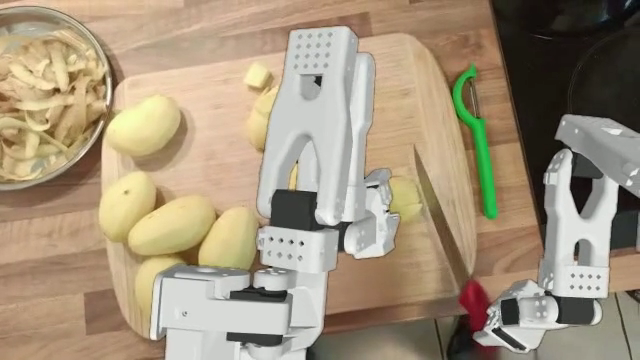}
\caption{SO-ARM101}
\end{subfigure}
\caption{\textbf{Ego2Robot cross-embodiment retargeting.}
(a) Original human egocentric video frame; (b)--(p) the same interaction
retargeted to 15 different robot embodiments.}
\label{fig:ego2robot_grid}
\end{figure}

\textbf{Cross-embodiment retargeting.}
Qualitatively, Figure~\ref{fig:ego2robot_grid} shows a single human egocentric interaction retargeted to 15 robot embodiments, ranging from industrial and collaborative manipulators (e.g., Franka, KUKA, Kinova, and UR arms) to low-cost arms (e.g., PiPER, WidowX, and SO-ARM101) and the bimanual ALOHA~2 platform. Although these embodiments differ substantially in kinematic structure, degrees of freedom, and workspace, the retargeted end effectors reproduce the same interaction, and all embodiments are handled through the shared registry-based interface rather than separate synthesis pipelines.

\subsection{Synthesized-Data Quality}
\label{sec:exp_ego2robot_quality}

We assess our implementation of the Ego2Robot pipeline
against the reference release using the standard of
Section~\ref{sec:ego2robot_eval}
(Table~\ref{tab:ego2robot_quality}). Our implementation is
better on eleven of the fourteen metrics, the reference
implementation on two, IK rotation error and human-residue
severity, and key-event visual alignment is a tie. The
largest margins are kinematic smoothness, where the jerk P95
is 6.4$\times$ lower, and the temporal metrics of motion
naturalness, where pose continuity is 42.3 points higher and
pause-and-recovery 38.5 points higher on the normalized
scale. The largest regression is human-residue severity,
33.3 points lower. Among the eleven, the velocity-reference
violation rates are near-identical, 3.89\% against 3.94\%,
and we do not read that difference as meaningful.

\textbf{Protocol.} Both pipelines consume the same source
corpus of human egocentric episodes, organized into five
top-level categories and 39 second-level operations with
five episodes each. The four kinematic metrics are computed
on all 195 synthesized episodes of each pipeline and
macro-averaged per episode. The ten ordinal metrics are
scored on a stratified sample of 39 episodes, one drawn at
random from each second-level operation so that every
operation is represented exactly once, and both pipelines
are graded by the same model under the same standard. The
level counts in Table~\ref{tab:ego2robot_quality} do not
always sum to 39, because episodes the grader marked unknown
or not applicable are excluded.

\begin{table}[!ht]
\centering
\caption{\textbf{Synthesized-data quality of the reference
Ego2Robot implementation and ours}, under the standard of
Section~\ref{sec:ego2robot_eval}. Kinematic metrics are
per-episode macro averages over all 195 synthesized
episodes and are lower-is-better; the ordinal metrics are
normalized scores in percent over a stratified sample of 39
episodes and are higher-is-better, with the accompanying
triple giving the number of episodes at levels 0, 1, and 2.
Triples need not sum to 39, because episodes marked unknown
or not applicable are excluded. The better value of each
metric is in bold; $\Delta$ is in points and is defined only
where the two sides share a scale.}
\label{tab:ego2robot_quality}
\small
\setlength{\tabcolsep}{5pt}
\begin{tabular}{@{}lrcrcr@{}}
\toprule
 & \multicolumn{2}{c}{\textbf{Reference implementation}}
 & \multicolumn{2}{c}{\textbf{LoongForge-Embodied}} & \\
\cmidrule(lr){2-3}\cmidrule(lr){4-5}
\textbf{Metric} & Value & 0/1/2 & Value & 0/1/2
 & $\Delta$ \\
\midrule
\multicolumn{6}{@{}l}{\textit{Kinematic feasibility}
 (lower is better)} \\
Velocity violation rate (\%) & 3.938 & --
 & \textbf{3.886} & -- & -- \\
Acceleration violation rate (\%) & 8.934 & --
 & \textbf{6.829} & -- & -- \\
Trajectory jerk P95 (unit/s$^3$) & 1370.9 & --
 & \textbf{213.5} & -- & -- \\
IK rotation error ($^\circ$) & \textbf{5.393} & --
 & 7.299 & -- & -- \\
\midrule
\multicolumn{6}{@{}l}{\textit{Visual quality}
 (normalized score, higher is better)} \\
Key-event alignment & 65.4 & 1/25/13 & 67.1 & 4/15/16
 & $+1.8$ \\
Object fidelity & 80.8 & 0/15/24 & \textbf{92.1} & 2/2/34
 & $+11.3$ \\
Human-residue severity & \textbf{96.2} & 1/1/37 & 62.8
 & 6/17/16 & $-33.3$ \\
Visual-spatial plausibility & 79.5 & 0/16/23
 & \textbf{84.2} & 2/8/28 & $+4.7$ \\
\midrule
\multicolumn{6}{@{}l}{\textit{Motion naturalness}
 (normalized score, higher is better)} \\
Overall naturalness & 59.0 & 0/32/7 & \textbf{68.0}
 & 4/17/18 & $+9.0$ \\
Pose continuity & 51.3 & 5/28/6 & \textbf{93.6} & 0/5/34
 & $+42.3$ \\
Velocity rhythm & 57.7 & 0/33/6 & \textbf{79.5} & 2/12/25
 & $+21.8$ \\
Path plausibility & 65.4 & 0/27/12 & \textbf{73.1}
 & 4/13/22 & $+7.7$ \\
Pause and recovery & 61.5 & 0/30/9 & \textbf{100.0}
 & 0/0/39 & $+38.5$ \\
Bimanual coordination & 62.5 & 0/27/9 & \textbf{81.4}
 & 3/7/25 & $+18.9$ \\
\bottomrule
\end{tabular}
\end{table}

\textbf{Kinematic feasibility.} The jerk P95 falls from
1370.9 to 213.5 unit/s$^3$ and the acceleration-reference
violation rate from 8.93\% to 6.83\%, so the trajectories
our implementation retargets are both smoother at their
roughest and less often outside the acceleration envelope of
the target embodiment. Velocity-violation rates are
effectively equal. The one kinematic regression is IK
rotation error, 7.30$^\circ$ against 5.39$^\circ$: the
achieved end-effector orientation tracks the commanded
orientation less closely than in the reference
implementation. We report this without attributing it to a
specific stage of the pipeline.

\begin{figure}[!ht]
    \centering
    \captionsetup[subfigure]{font=small,skip=2pt}
    \begin{subfigure}[b]{0.24\textwidth}
        \centering
        \includegraphics[width=\linewidth]{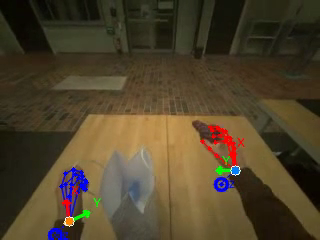}
        \caption{}
    \end{subfigure}\hfill
    \begin{subfigure}[b]{0.24\textwidth}
        \centering
        \includegraphics[width=\linewidth]{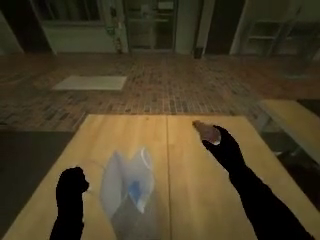}
        \caption{}
    \end{subfigure}\hfill
    \begin{subfigure}[b]{0.24\textwidth}
        \centering
        \includegraphics[width=\linewidth]{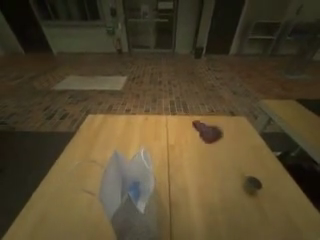}
        \caption{}
    \end{subfigure}\hfill
    \begin{subfigure}[b]{0.24\textwidth}
        \centering
        \includegraphics[width=\linewidth]{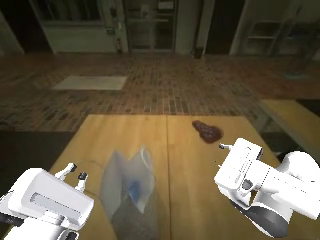}
        \caption{}
    \end{subfigure}

    \vspace{0.4em}

    \begin{subfigure}[b]{0.24\textwidth}
        \centering
        \includegraphics[width=\linewidth]{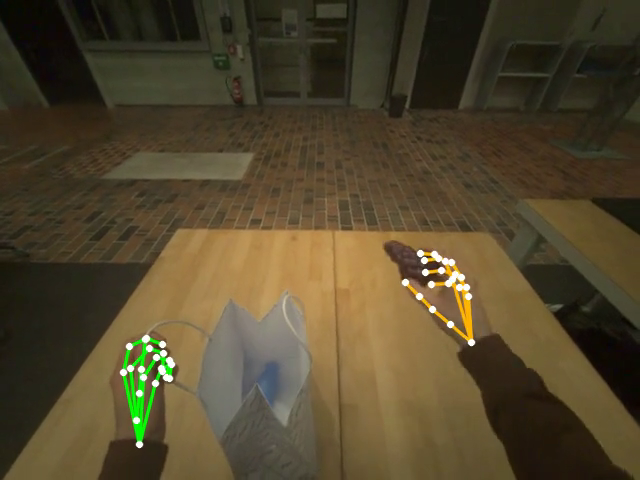}
        \caption{}
    \end{subfigure}\hfill
    \begin{subfigure}[b]{0.24\textwidth}
        \centering
        \includegraphics[width=\linewidth]{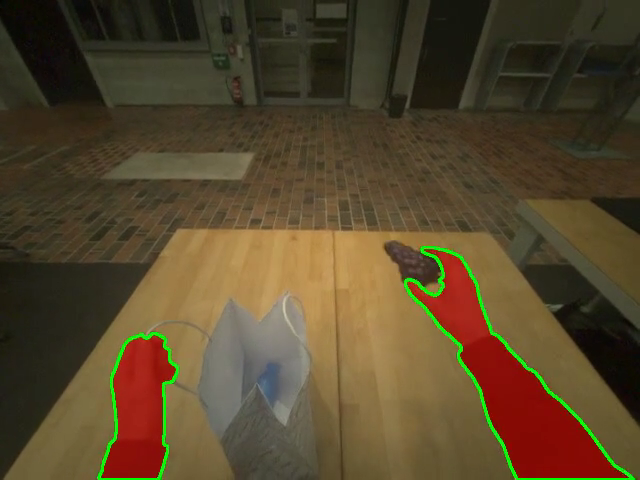}
        \caption{}
    \end{subfigure}\hfill
    \begin{subfigure}[b]{0.24\textwidth}
        \centering
        \includegraphics[width=\linewidth]{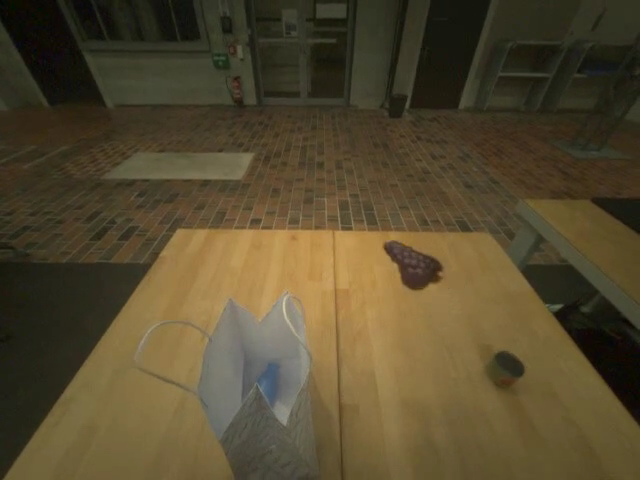}
        \caption{}
    \end{subfigure}\hfill
    \begin{subfigure}[b]{0.24\textwidth}
        \centering
        \includegraphics[width=\linewidth]{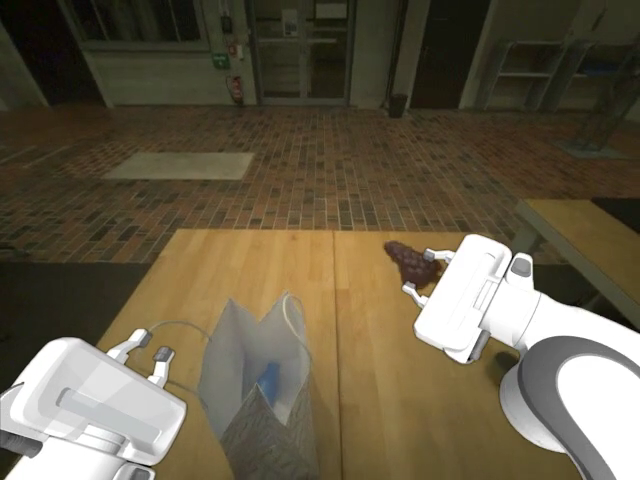}
        \caption{}
    \end{subfigure}
    \caption{\textbf{Cross-embodiment synthesis stages for the reference Ego2Robot
    implementation and our metric-guided variant.} All eight panels show the same
    frame of the same human egocentric episode: the reference implementation in
    the top row, (a)--(d), ours in the bottom row, (e)--(h). Left to right:
    reconstructed hand keypoints, hand and forearm segmentation, the frame after
    human removal and background inpainting, and the retargeted robot rendered
    into the scene. The inpainted frame in (g) preserves table edges, floor
    structure, and object texture that are lost in (c), the visual difference
    behind the object-fidelity margin in
    Table~\ref{tab:ego2robot_quality}. At this
    frame the reference gripper in (d) closes short of the target object, leaving
    a visible gap, while in (h) the fingers enclose it. Our variant was tuned
    against the standard of Section~\ref{sec:ego2robot_eval}, and these panels
    show what that standard selects for at the level of a single frame.}
    \label{fig:ego2robot_frames}
\end{figure}

\textbf{Visual quality.} Object fidelity and visual-spatial
plausibility favor our implementation by 11.3 and 4.7
points, with 34 of 38 and 28 of 38 scored episodes at the
top level.
Figure~\ref{fig:ego2robot_frames} shows this difference
on a single frame. Key-event visual alignment is unchanged: the
1.8-point gap in the table is an artifact of differing
denominators, since four of our episodes could not be
scored, and on the 35 episodes scored on both sides each
pipeline is better on eight episodes and the two agree on
the remaining 19, giving identical means. Human-residue
severity is a clear regression. Our implementation leaves
identifiable human limbs in 23 of 39 episodes at level 0 or
1, against 2 of 39 for the reference implementation, so 41\%
of our episodes reach the top level where 95\% of the
reference episodes do. This is the one axis on which the
reference release is substantially better, and the remaining
metrics do not compensate for it: removing human appearance
from the rendered output is a failure mode distinct from
producing a kinematically sound trajectory.

\textbf{Motion naturalness.} Our implementation scores
higher on all six metrics, and the two largest gaps are
temporal. Every one of the 39 episodes reaches the top level
on pause-and-recovery, against 9 of 39 for the reference
implementation, and 34 of 39 do so on pose continuity,
against 6 of 39. Velocity rhythm and bimanual coordination
follow at 21.8 and 18.9 points. The two smallest gaps,
overall naturalness and path plausibility, come with a tail
that the aggregate score hides: our implementation has four
level-0 episodes on each, where the reference implementation
has none. Its distribution is the more bimodal of the two,
with more episodes at the top level and a small number of
outright failures, whereas the reference implementation
concentrates in the middle, placing 32 of 39 episodes at
level 1 on overall naturalness.

\textbf{What these numbers support.} The ordinal half of the
comparison rests on a single grading pass over 39 episodes,
without human re-annotation or a measure of inter-rater
agreement, and with one episode per second-level operation a
difference of one or two episodes on a metric falls within
the resolution of the sample. Only the large margins should
be read as real: the 42.3-point gap on pose continuity
corresponds to 33 single-level changes across the 39
episodes, whereas the 4.7-point gap on visual-spatial
plausibility corresponds to fewer than four. The kinematic
metrics are computed on the full set and are not subject to
this limitation, but they describe whether a trajectory is
executable rather than whether it is useful for training,
which Section~\ref{sec:exp_ego2robot} measures separately.

\subsection{Optimization Analysis}
\label{sec:opt_analysis}

Section~\ref{sec:model_specific} groups the dominant training bottlenecks
of the evaluated models into four categories (Table~\ref{tab:model_opt_summary}):
fragmented execution, memory--compilation coupling, data delivery, and
distributed communication. This subsection examines the first three using
representative experiments; communication is examined in
Section~\ref{sec:dist_comm}. Complete per-model analyses, including additional
profiles, ablations, and fidelity checks, are provided in
Appendix~\ref{app:model_analysis}.

\begin{wrapfigure}{r}{0.46\textwidth}
    \centering\captionsetup{font=small}\vspace{-1.0em}
    \includegraphics[width=\linewidth]{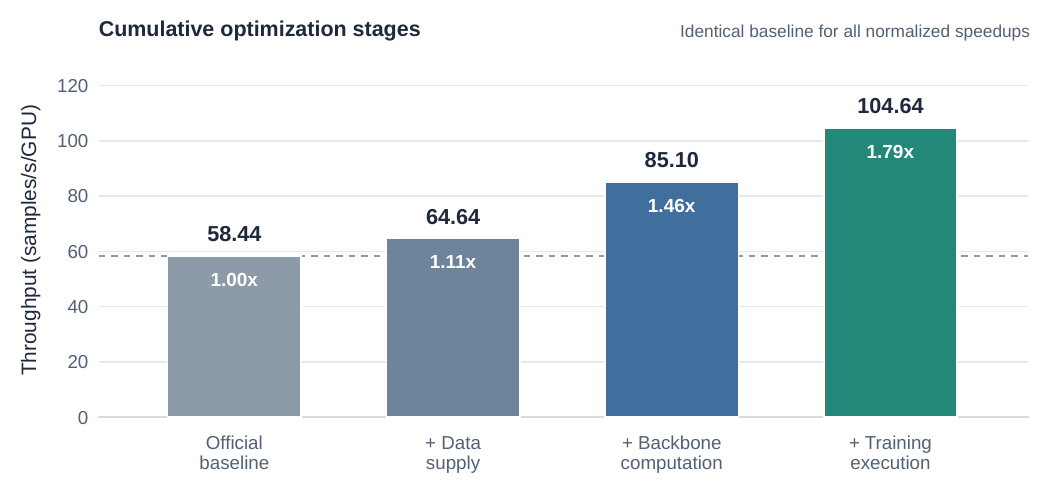}
    \caption{\textbf{Cumulative training throughput improvements on
    GR00T N1.7.} Throughput (samples/s/GPU) after successive groups of
    optimizations. The numbers inside the bars are speedups relative to
    the same official baseline; numerical labels above the bars indicate
    absolute throughput.}
    \label{fig:gr00t_ablation}
\end{wrapfigure}

\paragraph{Execution-bound training: GR00T N1.7.}
Profiling of GR00T N1.7 shows that forward
computation of the frozen Qwen3-VL backbone dominates the step and that
78.2\% of kernel launches last less than 20\,$\mu$s, indicating that the
workload is limited by fragmented execution rather than by individual
matrix-multiplication kernels.
Table~\ref{tab:gr00t_ablation} and Fig.~\ref{fig:gr00t_ablation} report
the throughput improvement obtained by successively enabling data,
computation, and execution optimizations. The official Isaac-GR00T
implementation achieves 58.44 samples/s/GPU in the supplied ablation
setting.

Episode-aware indexing and deterministic asynchronous shard
prefetching increase throughput to 64.64 samples/s/GPU (1.11$\times$).
Caching multimodal metadata, bypassing the unused vocabulary projection,
and fusing backbone operations raise throughput to 85.10
samples/s/GPU (1.46$\times$). With full-iteration CUDA Graph capture,
frozen-backbone cross-iteration pipelining, and streamlined gradient
processing, the final configuration achieves 104.64 samples/s/GPU,
corresponding to a 1.79$\times$ end-to-end speedup. These are
\emph{cumulative} results; they do not individually isolate the benefit
of each technique.

The results show that removing redundant multimodal data processing and
frozen-backbone computation improves the steady-state workload, while
graph-based execution and cross-iteration overlap reduce the overhead of a
fragmented training pipeline. The ablation establishes the \emph{combined}
value of the three optimization groups; a pipeline-only timing comparison is
still needed to quantify overlap efficiency independently.

\paragraph{Exact kernel replacement: Wall-OSS-0.5.}
For Wall-OSS-0.5, the dominant cost is not a single hot operator but the
volume of unfused element-wise work and kernel launches in its Transformer
and MoE pathways, where intermediate-tensor materialization also sets the
binding constraint on the per-device batch size. Replacing SwiGLU and
RMSNorm with bitwise-exact fused kernels lowers peak allocated GPU memory
from 92,362 to 73,244\,MiB (20.70\%), and the two kernels together with
fp32 normalization coverage raise throughput from 25.1 to 29.6 samples/s
per device relative to the communication-optimized configuration
(Appendix~\ref{app:walloss}). Both kernels pass bitwise forward and
backward comparisons against the eager operators, so the memory constraint
is relaxed without introducing any numerical deviation.

\begin{wrapfigure}{r}{0.48\textwidth}
    \centering\captionsetup{font=footnotesize}\vspace{-1.0em}
    \includegraphics[width=\linewidth]{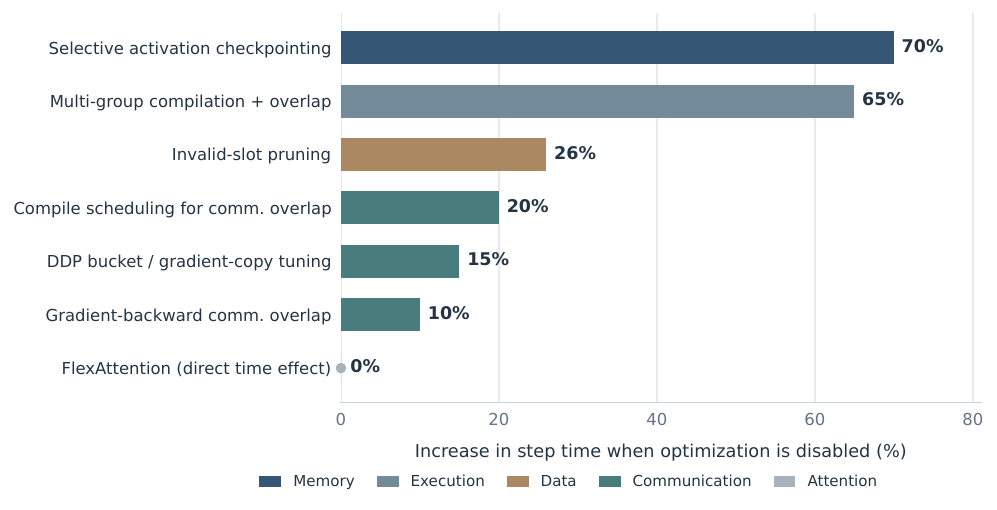}
    \caption{\textbf{Optimization ablation for $\pi_{0.5}$.}
    Each bar reports the increase in training step time when the
    corresponding optimization is disabled while the remaining
    optimizations stay enabled, measured on eight GPUs.
    The measurements are independent, not cumulative; compilation and
    communication-overlap interventions may interact.}
    \label{fig:pi05_ablation}
\end{wrapfigure}

\paragraph{Memory--compilation coupling: $\pi_{0.5}$.}
We analyze the contribution of individual $\pi_{0.5}$ optimizations by
measuring the change in training step time after disabling each component
while retaining the remaining optimizations (Fig.~\ref{fig:pi05_ablation}).
Disabling selective activation checkpointing increases the step time by
70\%, while disabling multi-group compilation and its associated overlap
optimizations increases it by 65\%. Removing invalid-slot pruning incurs a
further 26\% step-time penalty in its independent ablation. Communication-
related optimizations are also consequential: disabling compile-aware
communication scheduling, DDP gradient-bucket tuning, and gradient-backward
overlap increases step time by 20\%, 15\%, and 10\%, respectively. In
contrast, FlexAttention does not exhibit a measurable direct step-time
contribution in this ablation; its reported benefit lies primarily in
reducing attention-related memory overhead, which supports other
memory-sensitive choices. These percentages are independent ablation
measurements and must not be summed to estimate the total speedup,
especially because the compilation and communication-overlap effects are
coupled.

\paragraph{Data-path bottlenecks.}
For several models, a substantial part of the realized speedup comes from
the data path rather than from GPU computation. For DreamZero,
first-frame-only processing (decoding only the first frame per view when
video latents come from the cache) reduces iteration time by 26.83\% over the full
steady-state window (8.28\% in an early window and 36.77\% in a late window
with input stalls in the full-frame control) and lowers peak host
shared-memory usage from 43.48 to 21.82\,GiB; because the control includes
data-loading stalls, this gain should not be interpreted as a pure GPU
compute improvement. For Cosmos 3, migrating preprocessing to the GPU reduces
the \texttt{get\_batch} share of the profiled iteration from 45\% to 5\%, a
relative share rather than an end-to-end speedup. For X-VLA, increasing
data-loading parallelism from two to eight workers reduces step time from
1,081 to 704\,ms, after which the steady-state profile exhibits approximately
98\% GPU busy time and 97.5\% overlap between gradient AllReduce and backward
computation, indicating that data delivery rather than communication was the
appropriate optimization target. These measurements come from different
models and protocols and should not be compared directly.

\FloatBarrier

\subsection{Distributed and Communication Efficiency}
\label{sec:dist_comm}

We investigate distributed-strategy selection, low-precision communication,
hardware sensitivity and communication overlap, and numerical fidelity in
distributed training. The evidence is drawn from experiments related to
communication on LingBot-VA, DreamZero, and FastWAM,
supplemented by collective communication benchmarks on both a higher- and a lower-bandwidth interconnect. These are case studies from distinct models,
configurations, and measurement cohorts; their results are reported side by
side and must not be pooled, multiplied, or compared directly across models.
Measurement protocols and additional per-model results are provided in
Appendix~\ref{app:model_analysis}.

\begin{wrapfigure}{r}{0.42\textwidth}
  \centering\captionsetup{font=footnotesize}\vspace{-1.0em}
  \includegraphics[width=\linewidth]{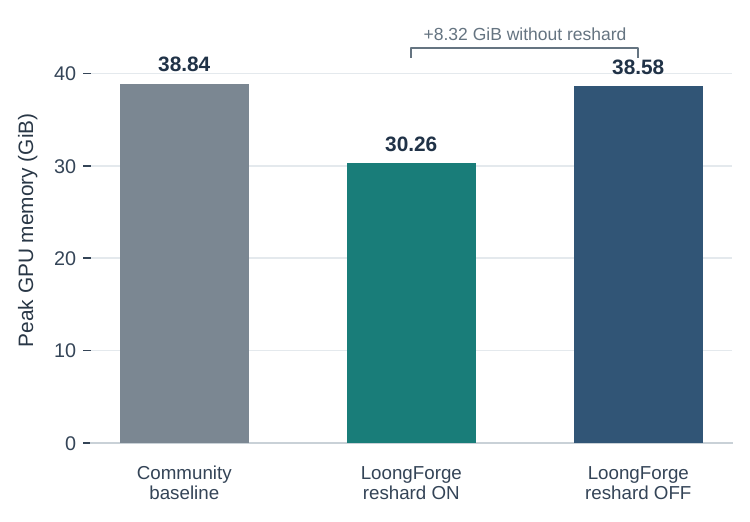}
  \caption{\textbf{Peak per-process GPU memory under different FSDP2 reshard policies.}
  Memory measurements use the maximum recorded per-process value across eight GPUs.
  Disabling reshard between gradient-accumulation microsteps increases peak memory
  from 30,986 MiB to 39,508 MiB (an additional 8,522 MiB, or 8.32 GiB).}
  \label{fig:lingbotva_fsdp_memory}
\end{wrapfigure}

\paragraph{Distributed strategy: trading memory for communication.}
When device memory permits, reducing parameter sharding or resharding can
remove communication from the critical path. For LingBot-VA, the
gradient-accumulation-aware FSDP2 strategy retains full parameters across
consecutive microsteps, avoiding redundant AllGather operations at the cost
of higher parameter memory residency. As shown in
Figure~\ref{fig:lingbotva_fsdp_memory}, disabling intermediate resharding
increases peak per-process GPU memory from 30.26~GiB to 38.58~GiB, while
disabling this scheduling change increases end-to-end step time by 21.16\%. The memory footprint without intermediate resharding
remains close to that of the community baseline (38.84~GiB), indicating that
this optimization primarily trades available device memory for reduced
communication rather than introducing an additional large memory footprint
relative to the original training configuration.

The same trade-off recurs in other models, each measured against its own
baseline. For DreamZero, disabling resharding after the forward pass reduces
iteration time by 25.17\% in a matched A/B comparison.
For FastWAM, replacing ZeRO-1 with
DDP increases throughput from 11.85 to 13.98 samples/s/GPU (18.0\%) in the
incremental analysis. For Cosmos 3,
reducing FSDP sharding for selected weights raises the source-reported
cross-platform throughput ratio from 1.50 to 1.66; unlike the two
preceding comparisons, this ratio is measured against a different
hardware platform rather than a same-platform baseline.

\begin{figure}[!ht]
  \centering
  \includegraphics[width=0.78\textwidth]{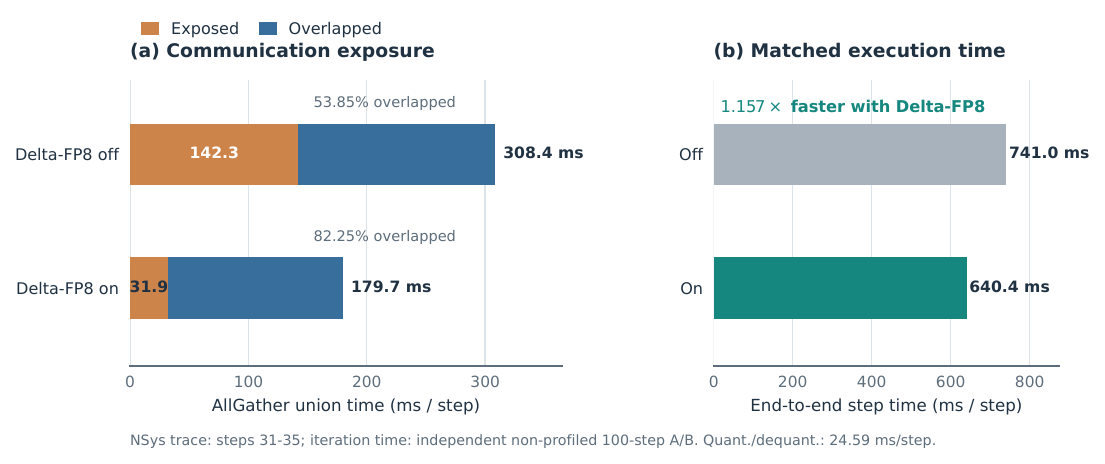}
  \caption{\textbf{Delta-FP8 communication mechanism in DreamZero.}
  (a) AllGather union time partitioned into the exposed component and the
  portion overlapped with other work, measured by NSys on steps 31--35.
  (b) Iteration times measured independently without profiling for the
  matched Delta-FP8 off/on pair (100 steps; steps 30--100 reported).
  Overlap is inferred from the reported union and exposed times, not from
  SM occupancy or hardware utilization.}
  \label{fig:dreamzero_fp8}
\end{figure}

\paragraph{Low-precision communication.}
DreamZero's Wan2.2-TI2V-5B recipe uses Delta-FP8 parameter AllGather under
FSDP2, which transfers quantized parameter updates while preserving BF16 computation.
To understand the communication mechanism, we profile Delta-FP8 on the
current FSDP2 recipe (Figure~\ref{fig:dreamzero_fp8}). FP8 delta transfer
decreases the AllGather time union from 308.39 to 179.72\,ms per step, while its
\emph{exposed} time falls from 142.31 to 31.90\,ms/step. Correspondingly, the
measured overlap ratio rises from 53.85\% to 82.25\%. The quantization and
dequantization kernels account for 24.59\,ms/step in the profile, which does
not negate the reduction in communication exposure. In a separate,
unprofiled 100-step comparison, iteration time falls from 0.7410 to
0.6404\,s, corresponding to a $1.157\times$ speedup. This distinction
between the profiling trace and the steady-state timing is important: the
reduction in exposed AllGather time is a mechanism-level observation, not
itself the end-to-end speedup.

Under DDP, FastWAM quantizes gradient buckets into block-scaled FP8 and
exchanges them through AllToAll--AllGather. In paired eight-GPU
measurements, this lowers normalized communication traffic to 0.508 of BF16
AllReduce, reduces median iteration time from 1090.6 to 962.4~ms, and
increases throughput from 14.671 to 16.626 samples/s/GPU (13.3\%).

\begin{figure}[!ht]
  \centering
  \includegraphics[width=0.78\textwidth]{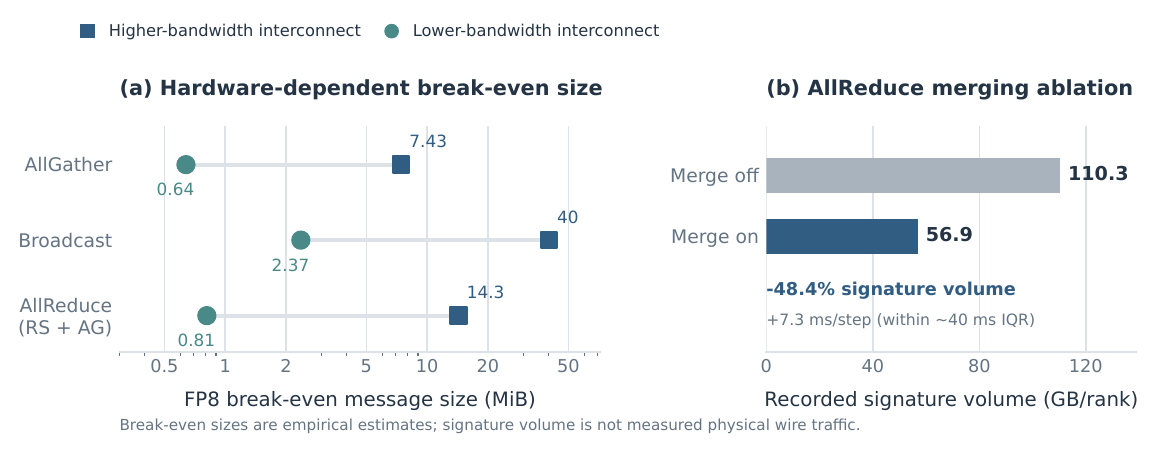}
  \caption{\textbf{Hardware sensitivity and communication-merging ablation.}
  (a) Experimentally estimated FP8 break-even message sizes for selected collective operations on a higher-bandwidth and a lower-bandwidth interconnect.
  (b) Recorded AllReduce signature-set volume with and without merging. Although the reported signature volume decreases by 48.4\%, the median step-time difference is only +7.3 ms, within measured variability (IQR approximately 40 ms).
  Signature-set volume should not be interpreted as measured physical wire traffic.}
  \label{fig:comm_hardware_ablation}
\end{figure}

\paragraph{Hardware sensitivity and communication volume.}
We further examine how hardware interconnects and communication scheduling affect compression efficiency (Fig.~\ref{fig:comm_hardware_ablation}). For AllGather, the measured FP8 compression break-even message size decreases from 7.43 MiB on the higher-bandwidth platform to approximately 655 KiB on the lower-bandwidth platform, indicating greater benefits from compression under constrained interconnect bandwidth. On such platforms, the gradient AllReduce can be replaced by the FP8 AllToAll--AllGather path described in Section~\ref{sec:fp8}; including the block-scale overhead, this reduces the bytes transmitted by each rank to approximately $0.51\times$ those of a BF16 ring AllReduce in theory. However, reducing communication volume alone does not guarantee acceleration. Enabling AllReduce merging reduces the recorded communication signature volume from 110.3 to 56.9 GB per rank (48.4\%), yet changes the median step time by only 7.3 ms, within measurement variability. This observation highlights the importance of exposed communication latency and computation--communication overlap, rather than communication volume alone.

\paragraph{Numerical fidelity.}
The low-precision communication paths above show small deviations within
short training windows. For DreamZero, the 100-step total-loss mean absolute relative error (MARE) between the
Delta-FP8 enabled and disabled variants is 0.1084\%, with no NaNs or skipped
updates. FastWAM's FP8 gradient communication
was reported to align in training loss with BF16, but per-step numerical
differences were not provided. These metrics differ across models and are
therefore not combined, and neither establishes long-horizon convergence or
downstream task equivalence.

\FloatBarrier

\section{Conclusion}
\label{sec:conclusion}
We presented LoongForge-Embodied, a native-PyTorch framework for unifying
and accelerating the training of heterogeneous VLA and WAM models. Its key
design principle is to separate model-specific computation from reusable
training infrastructure: configuration-driven model registration, a
lightweight model interface, and an extensible training engine allow
models with distinct computation graphs and training objectives to share
orchestration, distributed execution, data processing, checkpointing, and
evaluation. On top of this abstraction, shared acceleration mechanisms,
spanning efficient attention, compilation, CUDA Graphs, operator and
optimizer fusion, data delivery, and FP8-based communication for DDP and
FSDP, are integrated once at the framework level, while architecture-specific
and hardware-aware choices remain configurable for each model. The
framework further provides a unified data infrastructure with
cross-embodiment data synthesis and a training-aligned evaluation framework
with composable model and benchmark adaptation, supporting an integrated
workflow from data preparation to evaluation. LoongForge-Embodied achieves
1.47$\times$--4.38$\times$ end-to-end training throughput improvements over
the corresponding open-source implementations while maintaining
comparable benchmark success rates. Applying the same approach to the
evaluation loop, whose serial execution cost dominates the development
loop once training is accelerated, reduces evaluation wall-clock time
by 4.1$\times$--23.2$\times$ on the benchmark runs reported in
Section~\ref{sec:eval_throughput}.

\section*{Contributors}
\label{sec:contributors}
Contributors are listed in alphabetical order by their first names.
An asterisk ($^{*}$) denotes the team leader.

\vspace{0.4em}
\noindent{\small\renewcommand{\arraystretch}{1.15}%
\begin{tabular*}{\textwidth}{@{}@{\extracolsep{\fill}}llllll@{}}
Haicheng Jiang & Henghua Zhang$^{*}$ & Jun Shen & Junjie Xiao & Junping Xiao & Kangye Hu \\
Qiu Zhang & Shuai Han & Wanping Zhang & Weiji Chen & Xiangyu Peng & Xiaochuan Ye \\
Yizhan Zhao & Youwei Chen & Yuehang Sun & & & \\
\end{tabular*}}

% =====================================================================
\bibliographystyle{plainnat}
\bibliography{references}

\clearpage
\appendix
\setcounter{figure}{0}
\setcounter{table}{0}
\renewcommand{\thefigure}{A\arabic{figure}}
\renewcommand{\thetable}{A\arabic{table}}
\renewcommand{\theHfigure}{A\arabic{figure}}
\renewcommand{\theHtable}{A\arabic{table}}

\section{Detailed Model-Specific Optimization Analysis}
\label{app:model_analysis}

For each model, this appendix first describes the architecture, dominant
training bottlenecks, and optimization design summarized in
Section~\ref{sec:model_specific} (Table~\ref{tab:model_opt_summary}), and
then provides the complete analyses summarized in
Sections~\ref{sec:opt_analysis} and~\ref{sec:dist_comm}. Experiments
that appear in the main text are referenced rather than repeated.

\FloatBarrier

\subsection{$\pi_{0.5}$}
\label{app:pi05}

\paragraph{Architecture and bottlenecks.}
The $\pi_{0.5}$ model uses a dual-branch architecture that couples a PaliGemma-based vision-language backbone with a Gemma-based action expert for flow-matching-based action generation. Unlike conventional Transformer architectures, the two branches interact through joint attention at each layer, where their query, key, and value representations are concatenated for shared attention computation and subsequently separated for branch-specific processing. The attention mechanism follows a structured prefix-action pattern: visual-language prefix tokens attend bidirectionally within the prefix, while action tokens attend to both the prefix and the entire action sequence. This architecture introduces several performance bottlenecks. First, conventional attention implementations incur substantial memory overhead from explicit attention masks, grouped-query attention (GQA) key/value replication, and intermediate tensor materialization across branches. Second, activation storage and fine-grained operations, including normalization, positional encoding, and gated residual computation, increase memory pressure and execution overhead, constraining the feasible per-device batch size. Third, the coupled computation across branches complicates distributed parameter sharding and communication scheduling. These challenges become particularly pronounced on multi-GPU platforms without high-bandwidth GPU interconnects, where gradient synchronization and insufficient communication-computation overlap can substantially limit training throughput.

\paragraph{Optimization design.}
To address these bottlenecks, LoongForge-Embodied jointly optimizes attention computation, activation memory, graph execution, and distributed communication. We first replace conventional attention with FlexAttention, exploiting the structured prefix-action mask and native GQA support to avoid explicit key/value replication and reduce attention-related memory overhead. The attention mask is constructed once per training step and reused across layers, while cross-branch RoPE and adaptive normalization operations are consolidated to reduce redundant computation and intermediate tensors. Building on these memory savings, we employ selective activation checkpointing on the SigLIP vision encoder, whose relatively low recomputation cost enables a more favorable memory-computation trade-off than uniform checkpointing. Together with pruning invalid input slots, this reduces unnecessary computation and memory consumption, allowing larger effective per-device batches with less reliance on gradient accumulation. To improve execution efficiency, we introduce a multi-group compilation strategy that partitions Transformer layers into separately compiled groups, enabling operator fusion while preserving scheduling opportunities for overlapping gradient synchronization with backward computation. We further optimize DDP gradient buckets, gradient-buffer reuse, and communication scheduling to reduce synchronization overhead, particularly on bandwidth-constrained interconnects. For FSDP execution, we additionally constrain module wrapping according to the cross-branch dependencies of joint attention, avoiding unnecessarily fine-grained parameter gathering. These optimizations are designed to work together: reducing attention memory overhead enables more efficient recomputation and batch configurations, while compilation granularity is selected to balance kernel fusion against communication overlap.

The leave-one-out ablation of the $\pi_{0.5}$ optimizations is reported in Section~\ref{sec:opt_analysis} (Figure~\ref{fig:pi05_ablation}).

\paragraph{End-to-end training throughput.}
\begin{wrapfigure}{r}{0.4\textwidth}
    \centering\captionsetup{font=small}\vspace{-1.0em}
    \includegraphics[width=\linewidth]{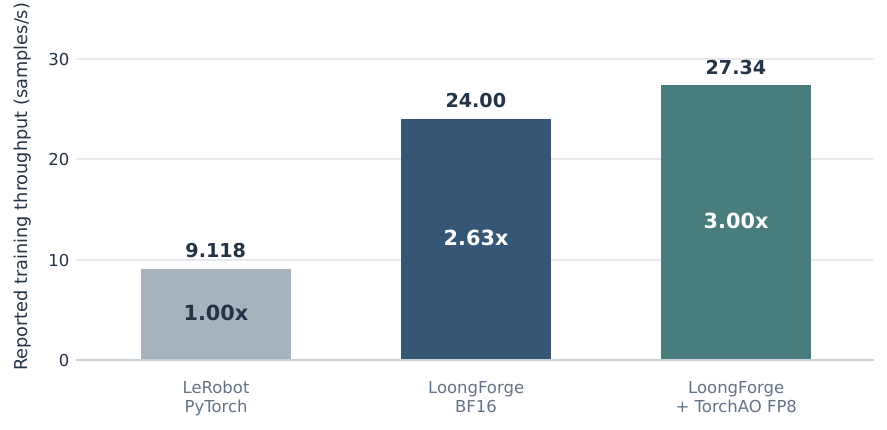}
    \caption{\textbf{Reported $\pi_{0.5}$ training throughput.}
    Training throughput on eight GPUs with a global batch size of
    256, comparing the LeRobot PyTorch baseline, LoongForge-Embodied
    with BF16 optimizations, and LoongForge-Embodied with TorchAO FP8.
    The values are reported as measured by the source in a single unit;
    no per-GPU relabeling is applied here.}
    \label{fig:pi05_throughput}
\end{wrapfigure}
Under the reported eight-GPU configuration with a global batch size of
256, $\pi_{0.5}$ training throughput increases from 9.118 samples/s in the
LeRobot PyTorch implementation to 24.0 samples/s with the BF16
LoongForge-Embodied pipeline, and to 27.34 samples/s with TorchAO FP8
(Fig.~\ref{fig:pi05_throughput}). Relative to the LeRobot reference, these
correspond to approximately $2.63\times$ and $3.00\times$ throughput,
respectively. The reference implementation is the LeRobot PyTorch
codebase, not the original JAX training implementation. The source
reports these three values in a single unit without specifying a
per-GPU normalization; to avoid overstating the comparison we retain
the reported unit rather than relabeling it.

\paragraph{FP8 computation efficiency and numerical behavior.}
We further evaluate FP8 linear computation using separate 200-iteration
training comparisons on the same eight-GPU configuration. Converting
all 218 eligible linear layers yields a per-GPU throughput of
27.4301 samples/s, compared with 24.1948 samples/s for BF16, an
improvement of 13.37\%. A trace-guided configuration retaining 216
active linear layers achieves 27.2991 samples/s (12.83\% above BF16).

\begin{figure}[!ht]
    \centering
    \includegraphics[width=\textwidth]{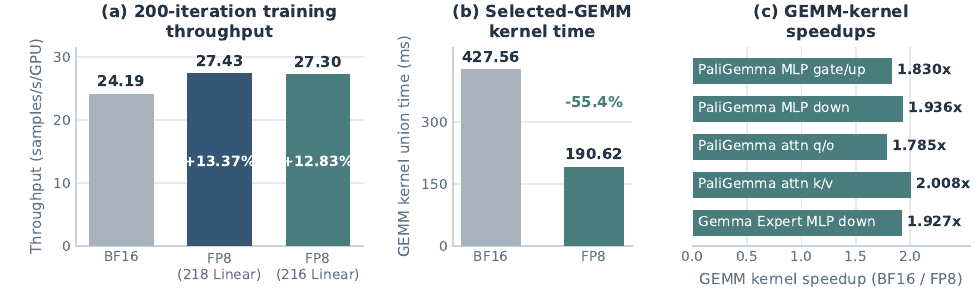}
    \caption{\textbf{FP8 efficiency analysis of $\pi_{0.5}$.}
    (a) Per-GPU throughput measured in separate 200-iteration BF16 and
    FP8 training runs (eight GPUs; global batch size 256).
    Converting 218 and 216 linear layers to FP8 improves throughput by
    13.37\% and 12.83\%, respectively. (b) Measured union time of the GPU
    kernels associated with the selected GEMM operations, profiled with
    Kineto (steps 20--21), decreases from 427.56 to 190.62\,ms. Kernel
    time should not be interpreted as end-to-end training time.
    (c) Representative GEMM-kernel speedups from FP8 computation: each
    ratio uses measured BF16 and FP8 median kernel durations for the
    indicated linear-layer family. These are kernel-level rather than
    end-to-end speedups.}
    \label{fig:pi05_fp8_analysis}
\end{figure}

\begin{wraptable}{r}{0.46\textwidth}
    \centering\captionsetup{font=small}\vspace{-1.0em}
    \caption{Measured 200-step $\pi_{0.5}$ FP8 training efficiency and
    numerical consistency on eight GPUs (global batch size 256).
    Loss metrics compare each FP8 configuration against the BF16
    reference.}
    \label{tab:pi05_fp8_numerics}
    \setlength{\tabcolsep}{2.6pt}
    \scriptsize
    \begin{tabular}{lrrr}
        \toprule
        Configuration & Thrpt. & Pearson $r$ & Mean err. \\
         & (samples/s/GPU) & & (\%) \\
        \midrule
        BF16 & 24.1948 & -- & -- \\
        FP8 (218 Linear) & 27.4301 & 0.99974017 & 0.5882 \\
        FP8 (216 Linear) & 27.2991 & 0.99973545 & 0.5623 \\
        \bottomrule
    \end{tabular}
\end{wraptable}
Kineto profiling of the selected GEMM kernels shows a reduction in their
union execution time from 427.56 to 190.62\,ms (55.4\%), while the
smaller end-to-end gain reflects non-GEMM computation, FP8 conversion,
communication, and other execution costs
(Fig.~\ref{fig:pi05_fp8_analysis}(a),(b)).
Over 200 iterations, the action-loss Pearson correlation relative to
BF16 is 0.99974017 for the 218-layer configuration and 0.99973545 for
the 216-layer configuration, with mean relative errors of 0.5882\% and
0.5623\%, respectively (Table~\ref{tab:pi05_fp8_numerics}). Neither FP8
run reports NaNs or skipped iterations. These observations support
short-horizon numerical consistency, but do not independently establish
full-convergence or downstream task-performance equivalence.

\paragraph{GEMM-kernel-level evidence.}
Figure~\ref{fig:pi05_fp8_analysis}(c) further decomposes the FP8 gains
at the kernel level for five representative linear-layer families.
Measured median kernel durations show BF16/FP8 speedups ranging from
$1.785\times$ (PaliGemma attention q/o projections) to $2.008\times$
(k/v projections), with the two largest MLP projections at $1.830\times$
and $1.936\times$. These are kernel-level measurements taken from
profiler steps 20--21 and are not end-to-end speedups; they indicate
where FP8's arithmetic gains originate once per-tensor scaling overhead
is amortized on large GEMMs.

\FloatBarrier

\subsection{GR00T N1.7}
\label{app:gr00t}

\paragraph{Architecture and bottlenecks.}
GR00T N1.7 is a VLA model comprising a Qwen3-VL backbone and a 32-layer diffusion transformer (DiT) action head trained with flow matching. The backbone processes multimodal observations and language instructions to extract visual-language representations, which are subsequently combined with robot states to condition continuous action generation. In the standard training configuration, the backbone remains frozen while only the action head is updated, resulting in an asymmetric computational workload: the backbone performs expensive multimodal forward computation, whereas the action head requires both forward and backward propagation. Despite eliminating backbone gradient computation, several performance bottlenecks remain. First, the Qwen3-VL forward pass involves substantial redundant computation and fragmented GPU execution, including repeated construction of position and attention metadata, unnecessary vocabulary projection, and numerous fine-grained RoPE and pointwise operations. Dynamic metadata construction further introduces CPU--GPU synchronization, leaving GPU execution pipelines intermittently idle. Second, multimodal data loading incurs redundant video decoding across dataset shards, causing CPU resource contention and unstable GPU utilization. Finally, frequent kernel launches, gradient processing, and distributed communication introduce additional execution gaps, limiting the effectiveness of conventional operator-level optimizations. These characteristics indicate that improving GR00T N1.7 training efficiency requires coordinated optimization across data preparation, backbone computation, and training execution.

\paragraph{Optimization design.}
To address these bottlenecks, LoongForge-Embodied introduces a series of optimizations tailored to the frozen-backbone architecture of GR00T N1.7. At the data level, we establish episode-aware sample indexing to eliminate redundant video decoding and employ asynchronous shard prefetching with deterministic augmentation replay, thereby overlapping data preparation with computation without altering the original sampling semantics. At the computation level, we exploit the relatively stable multimodal input layouts to cache position and attention metadata, avoiding repeated host-side construction and synchronization. We further bypass the unused language-model vocabulary projection, since action training consumes backbone hidden states rather than language logits, and fuse RoPE, RMSNorm, and MLP pointwise operations to reduce intermediate memory traffic and kernel launch overhead. Beyond these optimizations, we employ full-iteration CUDA Graph capture to eliminate CPU scheduling overhead across forward propagation, backward propagation, gradient synchronization, and optimizer updates. More importantly, we exploit the frozen backbone to construct a cross-iteration execution pipeline that separates backbone inference from action-head training into two CUDA Graphs. Since the backbone output for batch $n+1$ is independent of the action-head parameter updates for batch $n$, the backbone forward pass of the next batch can overlap with the current batch's action-head training and gradient communication. This scheduling strategy hides otherwise exposed execution and communication latency without introducing parameter-version inconsistencies. Finally, we integrate gradient scaling into a capturable AdamW optimizer and eliminate redundant gradient-buffer operations to reduce memory traffic during parameter updates. Together, these optimizations transform the fragmented training execution of GR00T N1.7 into a more continuous GPU pipeline while preserving the original training semantics.

\begin{figure}[!ht]
    \centering
    \captionsetup[subfigure]{font=small,skip=2pt}
    \begin{subfigure}[b]{0.525\textwidth}
        \centering
        \includegraphics[width=\linewidth]{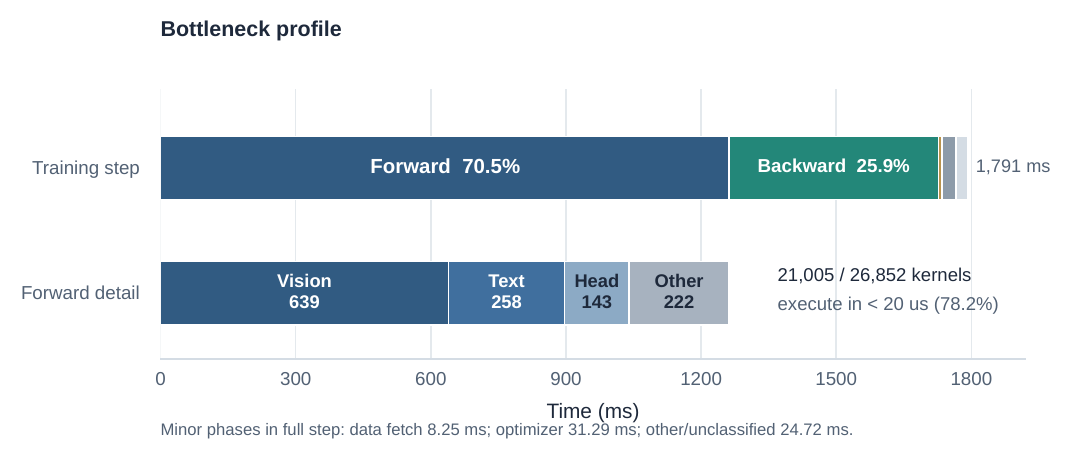}
        \caption{Training bottleneck profile.}
    \end{subfigure}\hfill
    \begin{subfigure}[b]{0.455\textwidth}
        \centering
        \includegraphics[width=\linewidth]{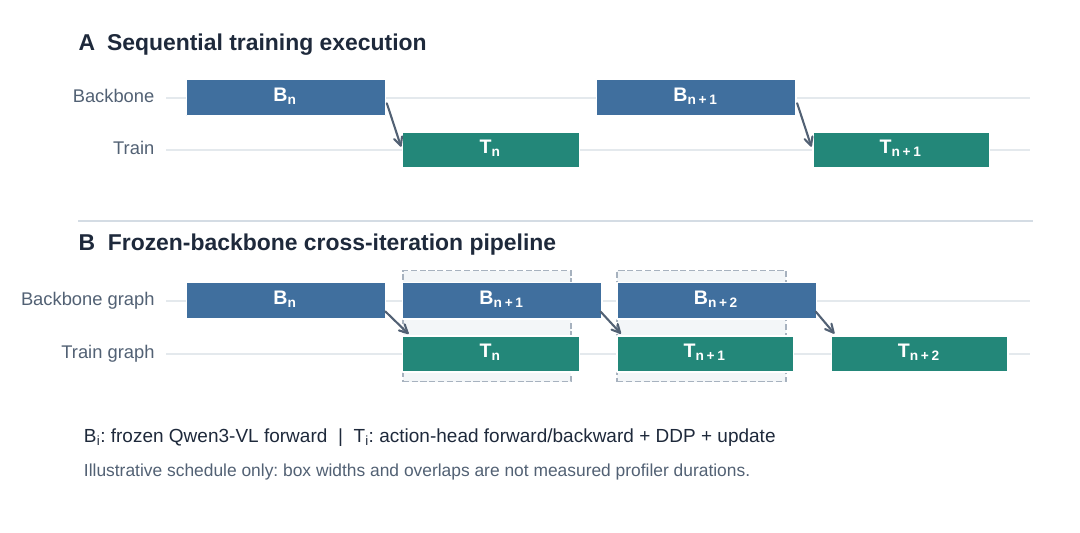}
        \caption{Cross-iteration scheduling.}
    \end{subfigure}
    \caption{\textbf{GR00T N1.7 bottleneck profile and the scheduling
    enabled by the frozen backbone.} (a) Breakdown of a profiled training
    step and its forward computation, using the measurements reported in
    the implementation notes. The small residual in the full-step timing is
    retained as unclassified overhead rather than assigned to an
    unsupported component. The kernel count highlights the frequency of
    short GPU operations. (b) $B_i$ denotes frozen Qwen3-VL forward
    computation for batch $i$, and $T_i$ denotes action-head training
    including gradient communication and optimizer update. In the optimized
    schedule, $B_{i+1}$ overlaps with $T_i$. \emph{Panel (b) is an
    explanatory schematic, not a measured GPU timeline; widths are not
    proportional to elapsed time.}}
    \label{fig:gr00t_profile}
\end{figure}

\paragraph{Bottleneck characterization.}
We first profile the training iteration of GR00T N1.7 to identify the
dominant sources of overhead. As shown in Fig.~\ref{fig:gr00t_profile}(a),
forward computation accounts for 1,262.23\,ms of the 1,791\,ms profiled
step (70.5\%), with the vision and text components of the frozen Qwen3-VL
backbone accounting for 638.90\,ms and 258.23\,ms, respectively. The
backward pass contributes another 464.52\,ms. The runtime is also highly
fragmented: 21,005 out of 26,852 kernel launches (78.2\%) have individual
durations below 20\,$\mu$s. Together with the redundant episode decoding
and repeated multimodal metadata construction identified in our profiling,
these measurements motivate optimizations beyond isolated
matrix-multiplication kernels, including reduction of host-side
synchronization, launch overhead, and uncovered communication.

\paragraph{Cumulative optimization ablation.}
The cumulative ablation is discussed in Section~\ref{sec:opt_analysis}
(Figure~\ref{fig:gr00t_ablation}); Table~\ref{tab:gr00t_ablation} lists
the corresponding configurations and per-stage throughput.

\paragraph{Execution overlap and fidelity.}
Figure~\ref{fig:gr00t_profile}(b) illustrates the rationale behind the
model-specific execution strategy. A frozen Qwen3-VL forward pass for
batch $n+1$ does not depend on the action-head parameter update for batch
$n$, allowing its execution to overlap with action-head training and DDP
communication on the preceding batch. This avoids a parameter-version
dependency that would arise if the backbone were trainable. The diagram
is schematic rather than a measured timing trace: quantifying
pipeline-only speedup and the actual reduction in exposed NCCL time
requires a separate matched Nsight trace or isolated ablation. 
\begin{wraptable}{r}{0.46\textwidth}
    \centering\captionsetup{font=small}\vspace{-1.0em}
    \caption{\textbf{Cumulative GR00T N1.7 acceleration ablation.}
    Throughput is reported per GPU; each row includes the optimizations of
    all preceding rows.}
    \label{tab:gr00t_ablation}
    \setlength{\tabcolsep}{3pt}
    \small
    \begin{tabular}{lrr}
        \toprule
        Configuration & Samples/s/GPU & Speedup \\
        \midrule
        Official Isaac-GR00T & 58.44 & 1.00$\times$ \\
        + Data supply & 64.64 & 1.11$\times$ \\
        + Backbone computation & 85.10 & 1.46$\times$ \\
        + Training execution & \textbf{104.64} & \textbf{1.79$\times$} \\
        \bottomrule
    \end{tabular}
\end{wraptable}
Under the matched model initialization, data sequence, seed, learning-rate
schedule, and distributed setup documented in the implementation notes,
the reported action-loss mean absolute error and maximum absolute error
are both zero between the baseline and optimized variants. This supports
fidelity for the recorded comparison but should not be extrapolated to
untested devices or configurations.

\paragraph{Takeaway.}
The results show that GR00T N1.7 benefits from a coordinated optimization
strategy: removing redundant multimodal data and frozen-backbone
computation improves the steady-state workload, while graph-based
execution and cross-iteration overlap reduce the overhead of a fragmented
training pipeline. The resulting 1.79$\times$ improvement is obtained
without a reported deviation in the paired action-loss checks. The
ablation establishes the \emph{combined} value of the three optimization
groups; a pipeline-only timing comparison is still needed to quantify
overlap efficiency independently.

\FloatBarrier

\subsection{X-VLA}
\label{app:xvla}

\paragraph{Architecture and bottlenecks.}
X-VLA adopts a VLA architecture that combines a Florence-2-based vision-language backbone with a Transformer-based action prediction head. Its visual encoder, DaViT, processes multi-view observations through a combination of convolutional layers, window attention, and channel attention, while the language encoder integrates visual and linguistic representations for downstream action prediction. This heterogeneous architecture introduces substantial computational overhead from attention operations, matrix multiplications, and numerous fine-grained memory-bound operators, including normalization, element-wise operations, and tensor transformations. Beyond these computational costs, the original implementation incurs additional CPU–GPU synchronization overhead during visual feature extraction and attention-mask processing. In particular, scalar extraction and data-dependent indexing introduce host synchronization and dynamic tensor shapes, which interrupt graph execution and hinder effective compilation. Meanwhile, gradient communication is largely overlapped with backward computation under DDP, making communication optimization less beneficial than eliminating synchronization and execution overhead. These characteristics motivate optimization strategies that improve computational graph continuity, reduce fine-grained execution overhead, and enhance data-loading efficiency.

\paragraph{Optimization design.}
To address these bottlenecks, LoongForge-Embodied applies model-specific computational graph restructuring together with reusable compilation and data pipeline optimizations. First, we eliminate unnecessary host synchronization in multi-view image encoding by replacing scalar-dependent validity checks and dynamic boolean indexing with fixed-shape feature extraction and tensor-based masking. We further simplify the Florence-2 encoder's attention-mask handling by bypassing redundant validity checks when all attention positions are valid, preserving the original attention semantics while removing synchronization barriers. These modifications establish a more static and compilation-friendly computational graph, enabling effective application of `torch.compile`. Building on this restructuring, TorchInductor reduces Python dispatch and kernel-launch overhead through graph-level optimization and operator fusion, particularly benefiting the numerous fine-grained operations in the visual backbone. In addition, we optimize the multi-view data pipeline through parallel data loading, pinned memory, and asynchronous prefetching to reduce GPU stalls caused by data preparation. Together, these optimizations illustrate how model-specific graph restructuring can unlock reusable framework-level acceleration mechanisms while addressing the distinctive execution characteristics of X-VLA.

\paragraph{X-VLA profiling.}
\begin{wrapfigure}{r}{0.46\textwidth}
    \centering\captionsetup{font=small}\vspace{-1.0em}
    \includegraphics[width=\linewidth]{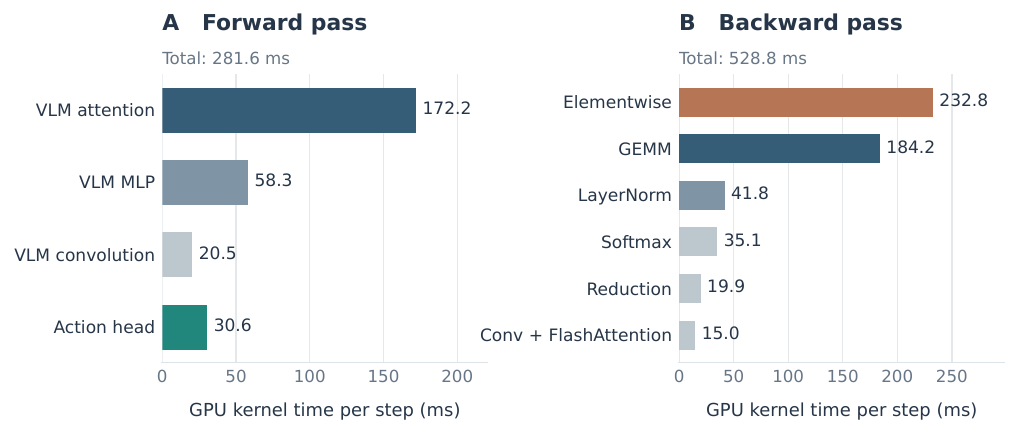}
    \caption{\textbf{Operator-level profiling of X-VLA.}
    (A) Forward GPU kernel time attributed to the VLM and action head;
    (B) backward GPU kernel time grouped by operator type.
    Values are per-step kernel durations measured on one process with a
    per-device batch size of 45 and averaged over 30 steady-state iterations.
    DDP gradient communication is excluded from the backward decomposition.
    The dominant costs are VLM attention in the forward pass and element-wise
    and GEMM kernels in the backward pass.}
    \label{fig:xvla_operator_profile}
\end{wrapfigure}
We first analyze the operator-level execution characteristics of X-VLA
(Fig.~\ref{fig:xvla_operator_profile}). The visual-language backbone
accounts for 250.9\,ms of the 281.6\,ms forward GPU kernel time, with
attention operations alone accounting for 172.2\,ms. The backward pass
spends 232.8\,ms on element-wise kernels and 184.2\,ms on GEMM kernels,
out of 528.8\,ms of total backward GPU kernel time. These measurements
show that X-VLA combines expensive attention computations with substantial
fine-grained operator overhead. In addition, profiling of the original
encoder reveals a redundant attention-mask check associated with roughly
200\,ms of host synchronization per step. This figure quantifies the
identified synchronization hotspot, rather than a standalone reduction
in end-to-end step time, and motivates graph restructuring to remove
synchronization points and enable effective compilation.

\begin{figure}[!ht]
    \centering
    \captionsetup[subfigure]{font=small,skip=2pt}
    \begin{subfigure}[b]{0.465\textwidth}
        \centering
        \includegraphics[width=\linewidth]{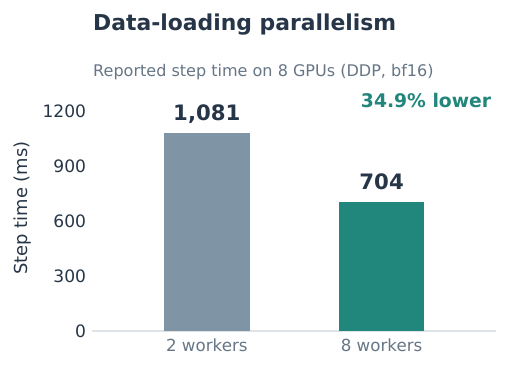}
        \caption{Data-pipeline tuning.}
    \end{subfigure}\hspace{0.017\textwidth}%
    \begin{subfigure}[b]{0.438\textwidth}
        \centering
        \includegraphics[width=\linewidth]{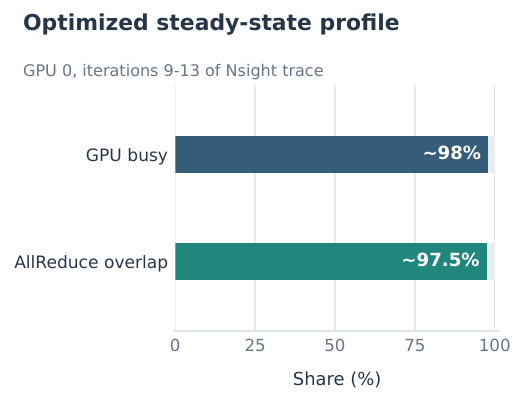}
        \caption{Steady-state execution.}
    \end{subfigure}
    \caption{\textbf{Data-pipeline tuning and steady-state execution of
    optimized X-VLA.} (a) Reported training step times with two versus
    eight data-loader workers on the 8-GPU DDP workload (bf16, three-view
    HDF5 input). The reduction from 1,081 to 704\,ms corresponds to a
    34.9\% lower step time (1.54$\times$ higher step throughput); this
    comparison is distinct from the full-framework speedup. (b) In the
    reported 8-GPU DDP profiling run, GPU~0 maintains approximately 98\%
    busy time during steady iterations 9--13, and approximately 97.5\% of
    DDP AllReduce is overlapped with the backward pass. These values
    characterize a separate Nsight trace and should not be interpreted as
    a paired before/after measurement of step time.}
    \label{fig:xvla_dataloader}
\end{figure}

\paragraph{Data-pipeline tuning and steady-state efficiency.}
As shown in Fig.~\ref{fig:xvla_dataloader}(a), increasing the reported
data-loading parallelism from two to eight workers reduces step time
from 1,081\,ms to 704\,ms, a 34.9\% reduction (1.54$\times$ higher step
throughput). The optimized steady-state profiling run further exhibits
approximately 98\% GPU busy time and 97.5\% overlap between gradient
AllReduce and backward computation
(Fig.~\ref{fig:xvla_dataloader}(b)).
These findings indicate that improving data delivery and removing
host-side synchronization are better justified optimization targets
than additional DDP communication tuning for this workload. Since the
currently available records do not provide matched per-stage step times
for synchronization removal and compilation, we do not attribute a
numerical fraction of the end-to-end acceleration to either technique. The 1.79$\times$ end-to-end speedup reported for X-VLA in
Section~\ref{sec:e2e_speedup} comes from a separate end-to-end throughput
measurement against the official implementation.

\FloatBarrier

\subsection{Wall-OSS-0.5}
\label{app:walloss}

\paragraph{Architecture and bottlenecks.}
Wall-OSS-0.5 is a VLA model built upon Qwen2.5-VL, comprising a 32-block vision Transformer and a 36-layer language model, together with a mixture-of-experts (MoE) action prediction pathway trained with flow matching. Its training pipeline involves visual feature extraction, multimodal representation processing, expert routing, and continuous action prediction, resulting in heterogeneous computation and communication patterns. Profiling reveals two major performance bottlenecks. First, frequently invoked operators such as SwiGLU and RMSNorm introduce substantial memory traffic and kernel-launch overhead due to intermediate tensor materialization and fragmented pointwise computation. These costs are particularly pronounced in the Transformer blocks and MoE components, where repeated element-wise operations contribute significantly to iteration latency and activation memory consumption. The latter is the binding constraint on the per-device batch size for this model, so reducing activation memory is as consequential as reducing step time. Second, the distributed training pipeline suffers from exposed parameter communication and inefficient scheduling. In particular, the module registration order differs from the actual forward execution order, causing parameter prefetching to miss critical dependencies and introducing head-of-line blocking during FSDP parameter materialization. 

\paragraph{Optimization design.}
To address these bottlenecks, LoongForge-Embodied introduces model-specific optimizations targeting both operator efficiency and distributed execution. At the operator level, we implement bitwise-exact CUDA kernels for SwiGLU and RMSNorm that preserve the rounding points and arithmetic ordering of the original PyTorch eager implementations. The fused SwiGLU kernel eliminates intermediate activation materialization and recomputes activations during backward propagation, while the optimized RMSNorm retains ATen reduction operations to preserve numerical behavior and fuses the surrounding pointwise computations. We further extend both kernels to mixed bf16/fp32 execution and eliminate redundant activation casting, reducing memory allocation and data movement without altering the original computation semantics. We also implement visual indexing operations in CUDA and pack the QKV projections into a single layout, further reducing redundant memory accesses. At the distributed execution level, we optimize FSDP resharding and distributed Muon parameter prefetching to overlap communication with computation. In particular, we reconstruct the forward prefetch chain according to the model's actual execution order rather than its module registration order, preventing demand-driven parameter gathering and communication-induced stalls. Topology-aware NCCL configuration and bucketed optimizer updates further reduce per-step communication and launch overhead. Together, these optimizations improve training efficiency while preserving the numerical fidelity of the original implementation.

\begin{figure}[!ht]
    \centering
    \captionsetup[subfigure]{font=small,skip=2pt}
    \begin{subfigure}[b]{0.46\textwidth}
        \centering
        \includegraphics[width=\linewidth]{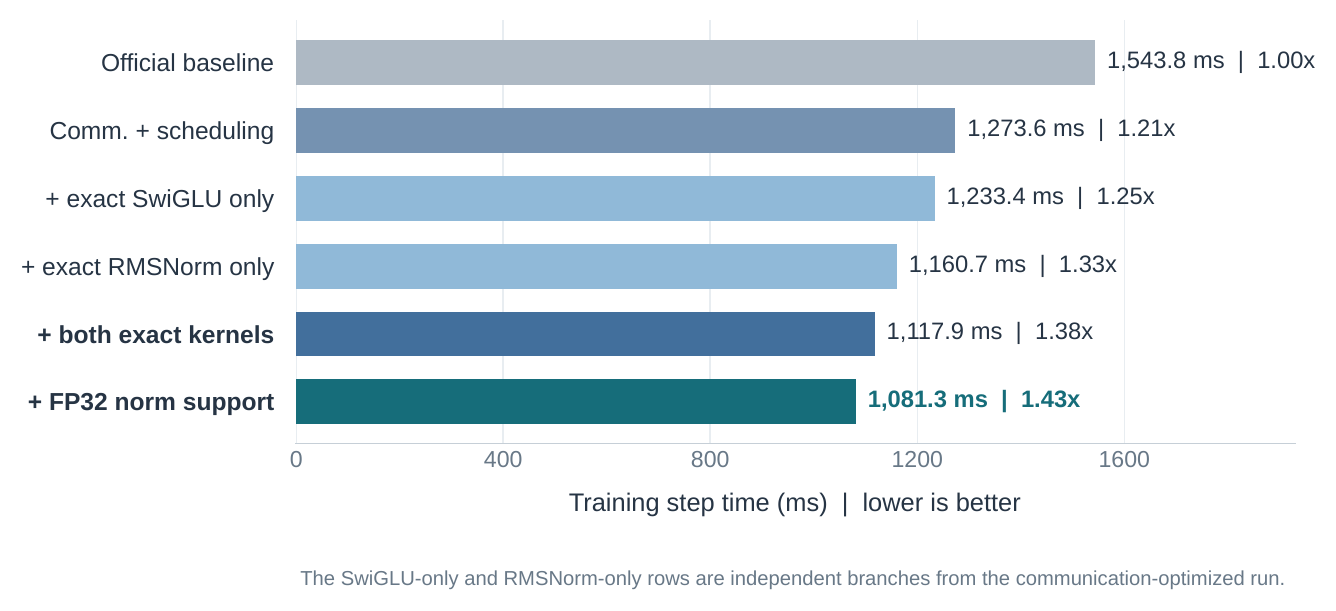}
        \caption{Training step-time ablation.}
    \end{subfigure}\hspace{0.02\textwidth}%
    \begin{subfigure}[b]{0.46\textwidth}
        \centering
        \includegraphics[width=\linewidth]{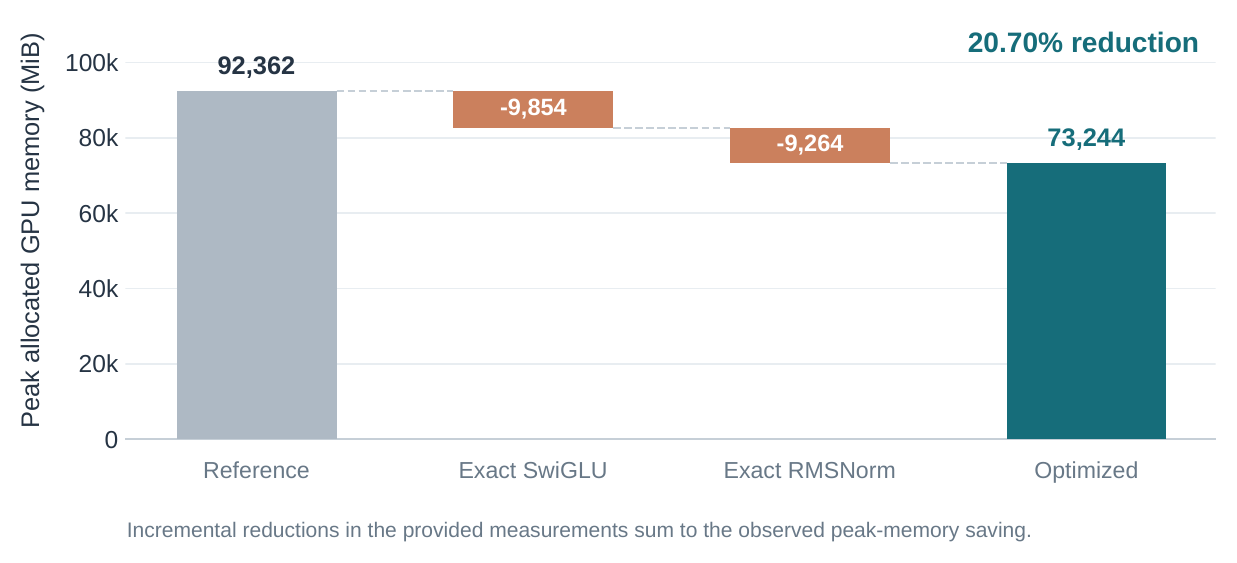}
        \caption{Peak allocated GPU memory.}
    \end{subfigure}
    \caption{\textbf{Step-time ablation and peak-memory savings for
    Wall-OSS-0.5.} (a) All rows use the same eight-GPU LIBERO bf16
    workload. The SwiGLU-only and RMSNorm-only rows are independent
    additions to the communication-optimized configuration; the last two
    rows enable both exact kernels and then fp32 RMSNorm coverage.
    Speedups are relative to the original implementation. (b) Waterfall of
    the measured peak-memory contributions of exact SwiGLU and exact
    RMSNorm, whose reductions are additive in this configuration. Values
    are in MiB and refer to allocated rather than reserved memory.}
    \label{fig:walloss_step_time}
\end{figure}

\paragraph{End-to-end step-time ablation.}
We evaluate the Wall-OSS-0.5 optimizations on LIBERO using eight GPUs, bf16 training, and a per-device batch size of 32, with the original implementation as the reference. Optional FP8 computation and communication paths are excluded. As shown in Fig.~\ref{fig:walloss_step_time}(a), the communication and scheduling optimizations reduce the step time from 1,543.8 to 1,273.6\,ms (1.21$\times$). Starting from this configuration, the exact SwiGLU and exact RMSNorm kernels individually reduce the step time to 1,233.4 and 1,160.7\,ms, respectively; these two rows are independent ablation branches rather than consecutive steps. Enabling both kernels yields 1,117.9\,ms, and extending exact RMSNorm to the fp32 normalization path further reduces the step time to 1,081.3\,ms. Overall, the optimized implementation reduces the step time by 29.96\% relative to the original implementation in this breakdown; the end-to-end speedup reported for Wall-OSS-0.5 in Section~\ref{sec:e2e_speedup} is measured separately in the end-to-end training configuration. Relative to the communication-optimized configuration, the two exact kernels and fp32 normalization support raise the throughput from 25.1 to 29.6 samples/s per device.

\paragraph{Peak-memory savings.}
The two exact kernels also reduce peak allocated GPU memory (Fig.~\ref{fig:walloss_step_time}(b)). Exact SwiGLU removes 9,854\,MiB, and exact RMSNorm removes a further 9,264\,MiB in the measured configuration. Their combined reduction of 19,118\,MiB lowers peak allocated memory from 92,362 to 73,244\,MiB (20.70\%), reflecting reduced intermediate tensor materialization and pointwise memory traffic.

\begin{wrapfigure}{r}{0.46\textwidth}
    \centering\captionsetup{font=small}\vspace{-1.0em}
    \includegraphics[width=\linewidth]{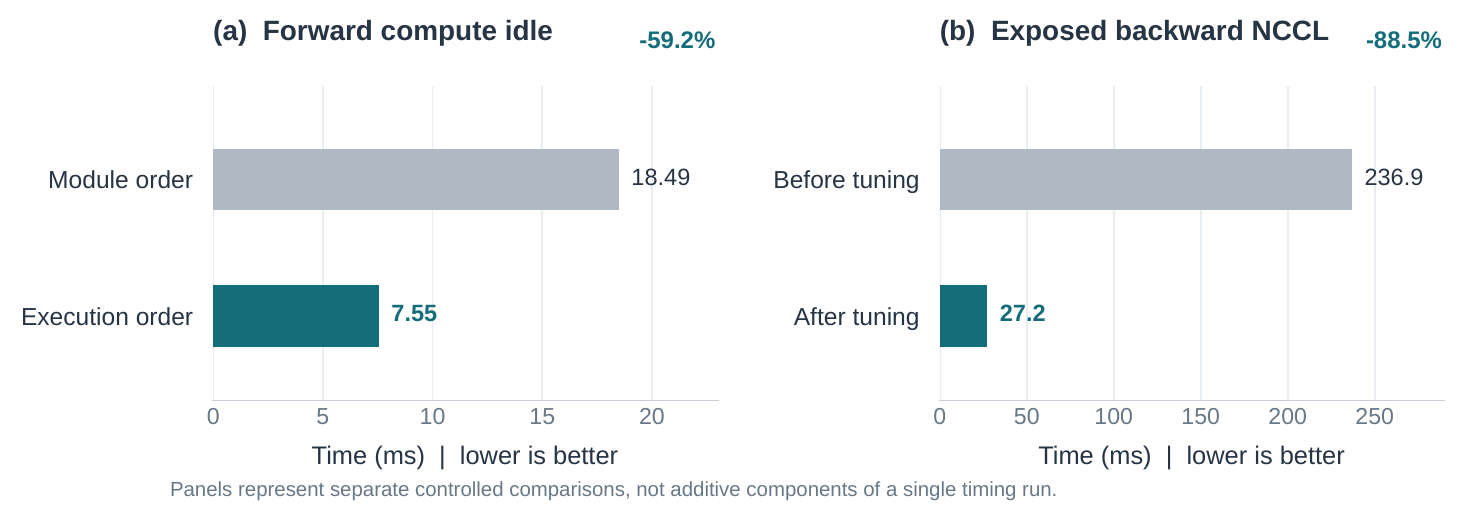}
    \caption{\textbf{Communication and scheduling profiling of Wall-OSS-0.5.}
    (a) Forward compute idle time before and after relinking the
    forward-prefetch chain according to the actual execution order.
    (b) Exposed backward NCCL time before and after FSDP resharding and
    prefetch tuning. The panels refer to separate comparisons with different
    baseline conditions.}
    \label{fig:walloss_profiling}
\end{wrapfigure}

\paragraph{Communication and scheduling profiling.}
Fig.~\ref{fig:walloss_profiling} examines the scheduling-related optimizations. Rebuilding the distributed Muon forward-prefetch chain in the actual execution order reduces the forward compute idle time from 18.49 to 7.55\,ms (59.2\%), while the end-to-end iteration time decreases by approximately 1.15\% in the same controlled comparison. Separately, FSDP resharding and prefetch tuning reduce the exposed backward NCCL time from 236.9 to 27.2\,ms (88.5\%). The two panels correspond to distinct controlled comparisons and are not additive components of a single timing measurement. Together, they indicate that both execution-order-aware prefetching and effective communication overlap matter for this heterogeneous VLA training pipeline.

\paragraph{Numerical fidelity.}
The exact SwiGLU kernel preserves the intermediate bf16 rounding points and backward arithmetic ordering of the eager implementation, and exact RMSNorm keeps the ATen reductions while fusing only pointwise operations; both pass bitwise forward and backward comparisons against the eager operators, including bf16 and fp32 coverage for RMSNorm. The prefetch-chain reordering preserves the training loss bitwise over a paired 40-step comparison. 

\FloatBarrier

\subsection{FastWAM}
\label{app:fastwam}

\paragraph{Architecture and bottlenecks.}
FastWAM is a WAM built upon the Wan2.2-TI2V-5B video DiT, employing a Mixture-of-Transformers (MoT) architecture to jointly model video and action generation through flow matching. Its MoT backbone consists of two structurally symmetric but parameter-independent Transformer experts for video latents and action tokens. Each expert maintains separate attention projections, normalization layers, and feed-forward networks, while cross-modal interaction is achieved by concatenating their query, key, and value representations into a shared self-attention operation at each layer. During training, a frozen video VAE encodes input frames into latent representations, while text embeddings are precomputed offline, leaving the two DiT experts and proprioceptive state encoder to be optimized. This architecture introduces two major performance bottlenecks. First, video encoding involves computationally expensive 3D convolutions, while the repeated normalization, modulation, and RoPE operations across MoT layers incur substantial kernel-launch overhead and intermediate memory traffic. Second, the independent expert parameters generate a large volume of gradients requiring synchronization across devices. This communication overhead becomes particularly pronounced on multi-GPU platforms without high-bandwidth GPU interconnects, such as the bandwidth-constrained platform used in our experiments (Section~\ref{sec:dist_comm}), where limited interconnect bandwidth and additional parameter synchronization under ZeRO-1 can make distributed training communication-bound.

\paragraph{Optimization design.}
To address these bottlenecks, LoongForge-Embodied applies a combination of model-specific computation optimizations and hardware-aware distributed training strategies. On the computation side, we optimize the frozen VAE encoder using channels-last memory layouts to reduce unnecessary tensor layout conversions during 3D convolution, and employ Triton kernels for efficient RoPE computation and fused LayerNorm–modulation operations in the MoT blocks, reducing intermediate memory accesses and kernel-launch overhead. We further adopt selective graph compilation for MoT blocks and VAE encoding, eliminate redundant all-true attention masks to enable more efficient attention kernels, and employ fused AdamW with cuDNN autotuning to improve optimizer and convolution efficiency. On the distributed training side, we replace the original ZeRO-1 strategy with DDP when device memory permits, avoiding additional parameter synchronization while enabling customized gradient communication. To mitigate the bandwidth bottleneck, we introduce an FP8 AllToAll–AllGather communication scheme that quantizes gradient buckets into block-scaled FP8 representations, exchanges gradient shards through AllToAll, performs local reductions in FP32, and redistributes the reduced gradients through FP8 AllGather. This design reduces communication volume without lowering the precision of forward and backward computation. Crucially, we preserve asynchronous communication–computation overlap by configuring GPU execution concurrency and leaving the precision exemptions disabled, since the exemption path is currently synchronous and would serialize gradient synchronization. Optional error-feedback compensation can further mitigate accumulated quantization errors during extended training. Together, these strategies adapt both computation and communication to FastWAM's dual-expert architecture and the underlying hardware characteristics, improving end-to-end training efficiency while maintaining training fidelity.

\paragraph{Incremental acceleration.}
\begin{wrapfigure}{r}{0.46\textwidth}
    \centering\captionsetup{font=small}\vspace{-1.0em}
    \includegraphics[width=\linewidth]{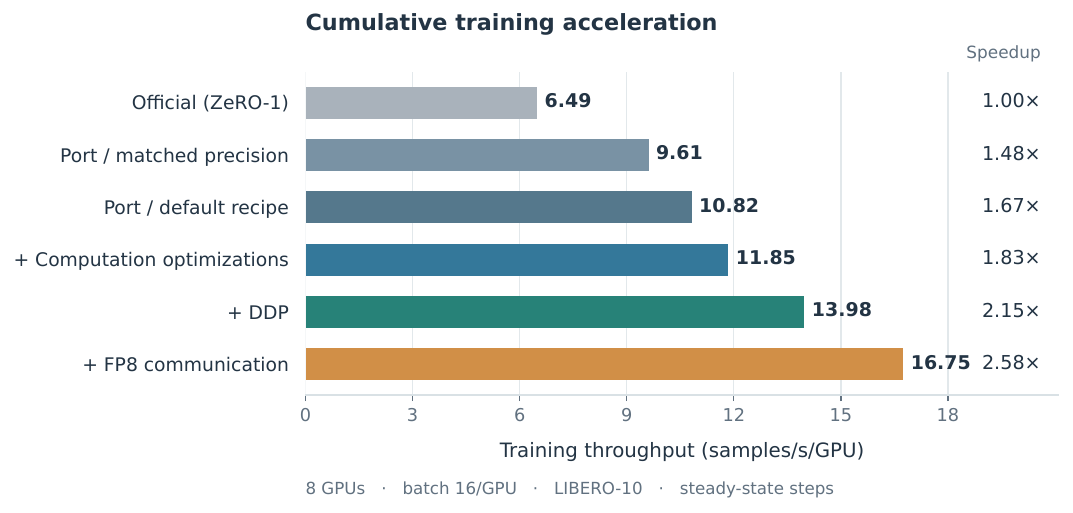}
    \caption{\textbf{Incremental training acceleration for FastWAM.}
    Measured on eight GPUs (batch size 16 per GPU, LIBERO-10);
    throughput is measured over steady-state iterations. The
    precision-matched port follows the official FP32-master-weight and
    autocast configuration, whereas the default LoongForge recipe uses BF16
    parameters without training autocast. All speedups are relative to the
    official steady-state result (6.49 samples/s/GPU). The final stage
    additionally enables asynchronous FP8 gradient communication with the
    precision exemptions disabled and
    \texttt{CUDA\_DEVICE\_MAX\_CONNECTIONS=8}.}
    \label{fig:fastwam_speedup}
\end{wrapfigure}
We investigate the effects of the model- and hardware-specific optimizations
for FastWAM using eight GPUs, with a per-GPU batch size of
16. Figure~\ref{fig:fastwam_speedup} shows an incremental analysis starting
from the community ZeRO-1 implementation. The official steady-state baseline
achieves 6.49 samples/s/GPU, whereas porting the implementation to LoongForge
with the official precision recipe preserved yields 9.61 samples/s/GPU
(1.48$\times$). With the default LoongForge BF16 training recipe and all
additional optimizations disabled, throughput reaches 10.82 samples/s/GPU
(1.67$\times$). Computation-side optimizations further increase throughput
to 11.85 samples/s/GPU, corresponding to a 9.5\% improvement over the
unoptimized port. Replacing ZeRO-1 with DDP increases throughput to
13.98 samples/s/GPU, a further 18.0\% improvement. Finally, with the
asynchronous FP8 communication path enabled and precision exemptions disabled,
throughput reaches 16.75 samples/s/GPU (2.58$\times$ relative to the official
steady-state baseline). The communication configuration also sets
\texttt{CUDA\_DEVICE\_MAX\_CONNECTIONS=8}; therefore, the final incremental
change should not be interpreted as an isolated FP8 quantization effect.
In a separate overall-throughput comparison, the official implementation
and optimized FastWAM obtain 6.35 and 16.63 samples/s/GPU, respectively
(2.62$\times$); these two measurement cohorts are kept separate throughout
the analysis.

\begin{figure}[!ht]
    \centering
    \includegraphics[width=0.6\textwidth]{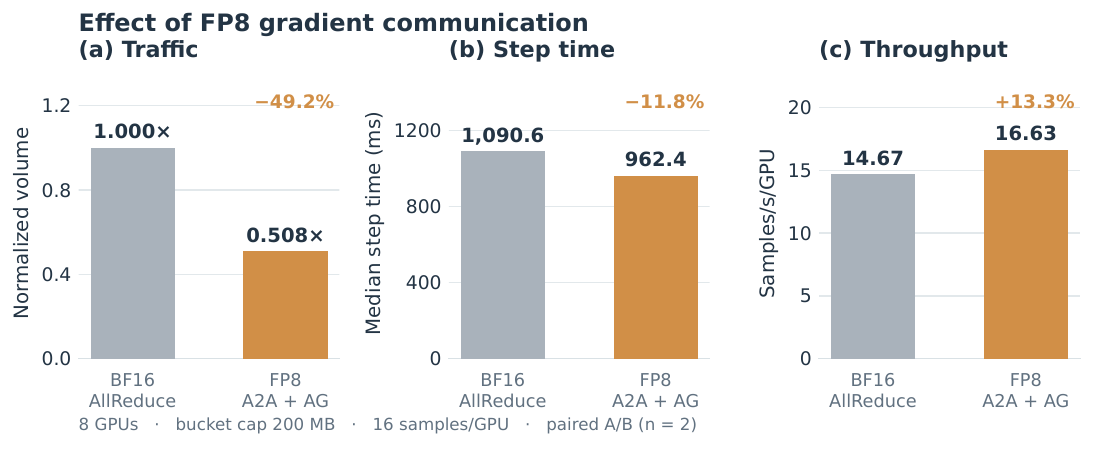}
    \caption{\textbf{Impact of FP8 AllToAll--AllGather gradient
    communication on FastWAM:} (a) normalized communication traffic,
    (b) median iteration time, and (c) per-GPU training throughput.
    Results are paired measurements on eight GPUs (batch size 16 per
    GPU, gradient bucket cap 200 MB; two alternating runs). FP8
    communication reduces transferred data to 0.508$\times$ that of BF16
    AllReduce, reducing iteration time from 1090.6 to 962.4 ms and
    increasing throughput from 14.671 to 16.626 samples/s/GPU. These
    numbers belong to one matched experiment and should not be combined
    with the separate ablation in Fig.~\ref{fig:fastwam_overlap}.}
    \label{fig:fastwam_fp8}
\end{figure}

\begin{figure}[!ht]
    \centering
    \includegraphics[width=0.6\textwidth]{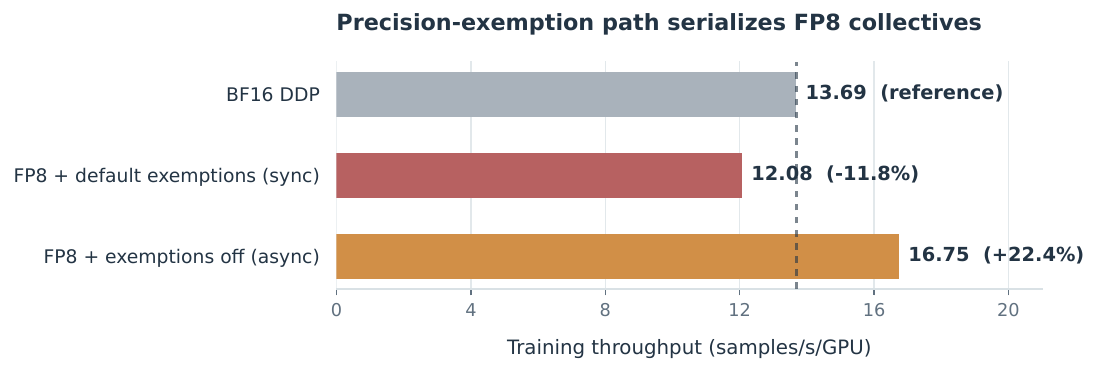}
    \caption{\textbf{Effect of the precision-exemption path on FastWAM's
    FP8 gradient communication} in an independent eight-GPU experiment on a
    later code revision. With the default precision exemptions enabled, the
    current implementation serializes the affected collectives and throughput
    falls below the matched BF16 DDP baseline; with exemptions disabled,
    asynchronous overlap is retained and throughput reaches
    16.75 samples/s/GPU, versus 13.69 for the baseline. Exemptions are
    optional and are disabled in the reported FastWAM configuration.
    Percent changes in the figure are calculated from the displayed rounded
    throughput values.}
    \label{fig:fastwam_overlap}
\end{figure}

\paragraph{Communication efficiency and execution overlap.}
Figure~\ref{fig:fastwam_fp8} isolates the contribution of FP8 gradient
communication in paired eight-GPU measurements. Replacing BF16 AllReduce
with block-scaled FP8 AllToAll--AllGather lowers normalized communication
traffic from 1.000 to 0.508, while reducing median iteration time from
1090.6 to 962.4 ms. Per-GPU throughput improves from 14.671 to
16.626 samples/s, an increase of 13.3\%. Realizing this gain requires that the compressed collectives still
overlap with backward computation. Precision exemptions, which keep selected
gradient buckets in higher precision, are optional in LoongForge-Embodied;
FastWAM meets its fidelity target without them, so they are disabled in the
reported configuration. Figure~\ref{fig:fastwam_overlap} records what the
alternative costs in the current code path: with exemptions enabled, the FP8
hook obtains 12.08 samples/s/GPU, below the matched BF16 DDP baseline of
13.69 samples/s/GPU, because the present exemption implementation serializes
the affected collectives; with exemptions disabled, the asynchronous path is
retained and throughput reaches 16.75 samples/s/GPU. This serialization
reflects how the exemption path is currently implemented rather than a
fundamental constraint, since an asynchronous exemption path is possible; we
therefore report the comparison as a configuration 
\begin{wrapfigure}{r}{0.46\textwidth}
    \centering\captionsetup{font=small}\vspace{-1.0em}
    \includegraphics[width=\linewidth]{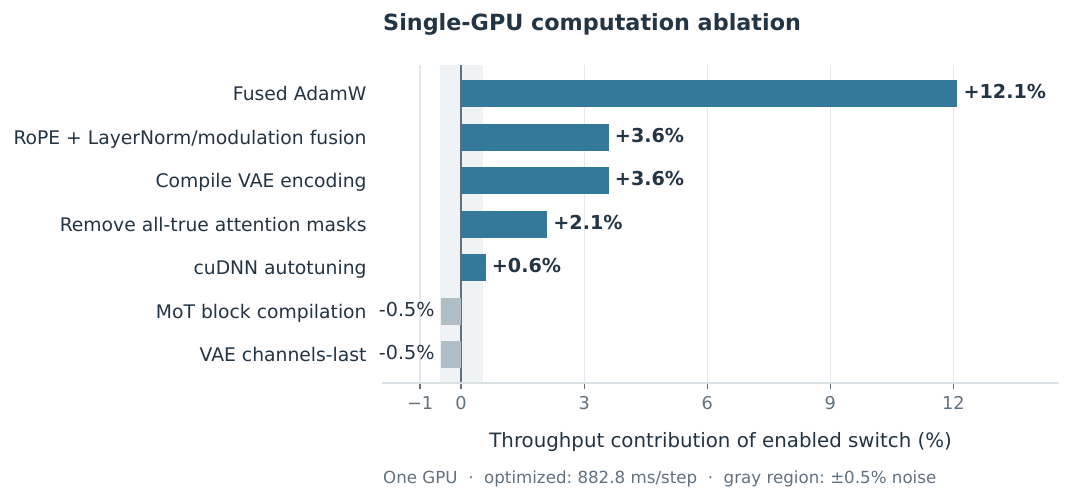}
    \caption{\textbf{Single-GPU computation ablation.} Each bar shows the
    reported contribution of one enabled optimization, evaluated by
    disabling that switch from the otherwise optimized configuration
    (882.8 ms/iteration). The gray band denotes approximately $\pm0.5\%$
    measurement noise; leave-one-out effects are contextual and are not
    additive. Fused AdamW exhibits the largest individual benefit, while
    MoT compilation and VAE channels-last are within measurement noise
    under this configuration.}
    \label{fig:fastwam_compute}
\end{wrapfigure}
note for the evaluated
setting rather than as a general claim about gradient compression.

\paragraph{Computation ablation and training fidelity.}
To examine model computation without inter-GPU communication, we further
perform a single-GPU leave-one-out ablation, shown in
Fig.~\ref{fig:fastwam_compute}. Fused AdamW offers the largest individual
throughput contribution (12.1\%), followed by RoPE and LayerNorm--modulation
fusion (3.6\%), compiled VAE encoding (3.6\%), and removal of redundant
all-true attention masks (2.1\%). By contrast, the effects of MoT block
compilation and VAE channels-last fall within approximately $\pm0.5\%$
measurement noise on this GPU. Across the full computation configuration,
iteration time decreases from 1090.9 to 882.8 ms, corresponding to a
23.6\% throughput improvement on one GPU, substantially larger than the
9.5\% computational gain observed in the eight-GPU setting. This contrast
indicates that exposed gradient communication limits the realized benefit
of local computation optimizations.

Finally, Table~\ref{tab:fastwam_fidelity} summarizes the available fidelity
evidence. Under a deterministic eight-GPU configuration on the higher-bandwidth platform, the LoongForge
and official implementations exhibit bitwise-identical video and action losses
across 300 steps and eight ranks, and all 1,649 trainable parameter tensors
match at the checked iterations. The FP8 communication implementation was
separately reported to maintain loss alignment with its BF16 counterpart.
Because raw long-horizon FP8 loss trajectories and downstream success-rate
measurements were not provided, we do not infer additional convergence
statistics or task-performance equivalence from these checks.

\begin{table}[!ht]
    \centering
    \caption{Training fidelity checks reported for FastWAM. Strict equality
    refers only to the deterministic comparison with the official ZeRO-1
    implementation; FP8 communication was separately reported to align in
    training loss with BF16, but per-step numerical differences were not
    provided.}
    \label{tab:fastwam_fidelity}
    \small
    \setlength{\tabcolsep}{3.5pt}
    \begin{tabular}{p{0.27\textwidth}p{0.22\textwidth}p{0.21\textwidth}p{0.22\textwidth}}
        \toprule
        \textbf{Comparison} & \textbf{Measured scope} & \textbf{Reported check} & \textbf{Result} \\
        \midrule
        Official ZeRO-1 vs.\ LoongForge (deterministic settings) & Video loss; 300 steps $\times$ 8 ranks & Maximum absolute difference & $0$ (bitwise equal) \\
        Official ZeRO-1 vs.\ LoongForge (deterministic settings) & Action loss; 300 steps $\times$ 8 ranks & Maximum absolute difference & $0$ (bitwise equal) \\
        Official ZeRO-1 vs.\ LoongForge (deterministic settings) & 1,649 trainable parameter tensors at steps 1, 2, and 20 & \texttt{torch.equal} for all tensors & All equal \\
        BF16 DDP vs.\ FP8 communication & Training loss comparison & Reported convergence alignment & Aligned; no raw trajectory provided \\
        \bottomrule
    \end{tabular}
\end{table}

\FloatBarrier

\subsection{LingBot-VA}
\label{app:lingbotva}

\paragraph{Architecture and bottlenecks.}
LingBot-VA employs a Transformer-based architecture for joint video and action modeling, with a backbone consisting of 30 WanTransformerBlocks. Its training computation combines self-attention over video-action representations, cross-attention to textual conditions, and feed-forward transformations. Unlike conventional fixed-length Transformer workloads, LingBot-VA processes variable-length video sequences with structured attention masks that capture temporal and cross-modal dependencies. This architecture introduces several performance challenges. First, varying sequence lengths require dynamic attention-mask construction, introducing substantial host-side dispatch and kernel-launch overhead. Second, full-block activation checkpointing reduces memory consumption but requires recomputation during backpropagation, amplifying the cost of frequently executed operations such as positional encoding, normalization, and attention. Third, distributed training with gradient accumulation incurs redundant parameter synchronization when FSDP repeatedly reshards and gathers parameters across consecutive microsteps. Fourth, variable-length samples cause per-rank computation imbalance within each gradient-accumulation step, so faster ranks wait for the slowest rank at gradient synchronization. Consequently, efficient training requires not only accelerating the computationally intensive Transformer operations but also eliminating redundant communication and fine-grained execution overhead specific to LingBot-VA's training workflow.

\paragraph{Optimization design.}
To address these bottlenecks, LoongForge-Embodied introduces a combination of model-specific computation, 
execution, and distributed-training optimizations. First, we adopt a gradient-accumulation-aware FSDP2 strategy 
that retains unsharded parameters across consecutive microsteps and performs resharding after optimizer updates, 
avoiding repeated AllGather operations at the cost of increased parameter memory residency. Second, rather than 
compiling the entire Transformer block, we selectively apply \texttt{torch.compile} to stable computational boundaries, 
including modulation, masking, and loss reduction, while dynamically compiling \texttt{create\_block\_mask} 
to accommodate varying sequence lengths. This approach reduces Python dispatch and kernel-launch overhead without 
imposing restrictive static-shape assumptions. Third, we replace Q/K RMSNorm with fused Transformer Engine kernels 
and implement model-specific Triton kernels for both forward and backward RoPE computations, eliminating expensive 
complex-number operations, redundant type conversions, and intermediate tensor materialization. These optimizations 
are particularly effective under activation recomputation, where the same operations are repeatedly executed. Fourth,
a cost-balanced sampler re-assigns the samples of each global batch across ranks by estimated computation cost and 
aligns the cost of corresponding microsteps across ranks, reducing synchronization stalls without changing the set 
of samples consumed per optimizer step. Together, they reduce redundant computation and communication while preserving 
the original training semantics, contributing to 
\begin{wrapfigure}{r}{0.46\textwidth}
    \centering\captionsetup{font=small}\vspace{-1.0em}
    \includegraphics[width=\linewidth]{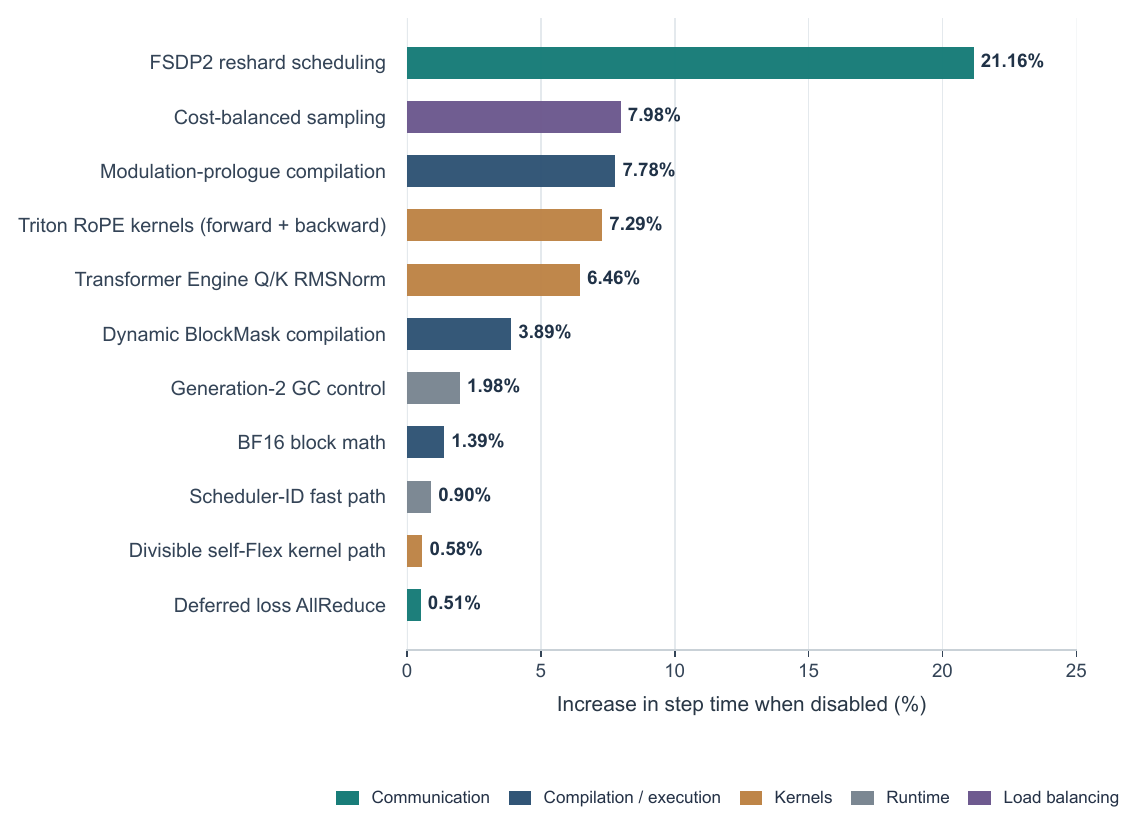}
  \caption{\textbf{Leave-one-out ablation for LingBot-VA.}
  Each bar reports the increase in end-to-end step time when the indicated
  optimization is disabled while all others remain enabled (eight GPUs,
  LIBERO, mean over steps 11--30 of 50-step runs; run-to-run coefficient of
  variation 0.11\%). Effects below the 3$\sigma$ noise band
  ($\approx$0.4\%) are omitted. Leave-one-out effects are contextual and
  \emph{not additive}, and must not be interpreted as a cumulative ablation.}
  \label{fig:lingbotva_ablation}
\end{wrapfigure}
a 2.20$\times$ end-to-end training speedup over the corresponding 
community baseline.

\paragraph{Optimization contributions.}
Figure~\ref{fig:lingbotva_ablation} reports a leave-one-out ablation of the
optimizations targeting LingBot-VA's execution, communication, and
load-balancing bottlenecks. Disabling FSDP2 reshard scheduling increases step
time by 21.16\%, the largest effect, reflecting the overhead of repeated
parameter gathers across gradient-accumulation microsteps. Cost-balanced
sampling, compiling the modulation prologue, the Triton RoPE kernels, and
Transformer Engine Q/K RMSNorm follow at 7.98\%, 7.78\%, 7.29\%, and 6.46\%,
while dynamic BlockMask compilation and generation-2 GC control contribute
3.89\% and 1.98\%. These are leave-one-out effects measured against the same
fully optimized configuration, not successive increments of a single
cumulative configuration. On the LIBERO benchmark, with identical model
architecture, training data, and hyperparameters, the fully optimized
framework reduces the unprofiled mean step time from 17.59~s to approximately
8.0~s, a 2.20$\times$ end-to-end speedup over the community baseline.

\paragraph{FSDP2 reshard tradeoff.}
This trade-off of the reshard scheduling change
is analyzed in Section~\ref{sec:dist_comm}
(Figure~\ref{fig:lingbotva_fsdp_memory}).

\paragraph{Residual bottlenecks.}
\begin{wrapfigure}{r}{0.46\textwidth}
    \centering\captionsetup{font=small}\vspace{-1.0em}
    \includegraphics[width=\linewidth]{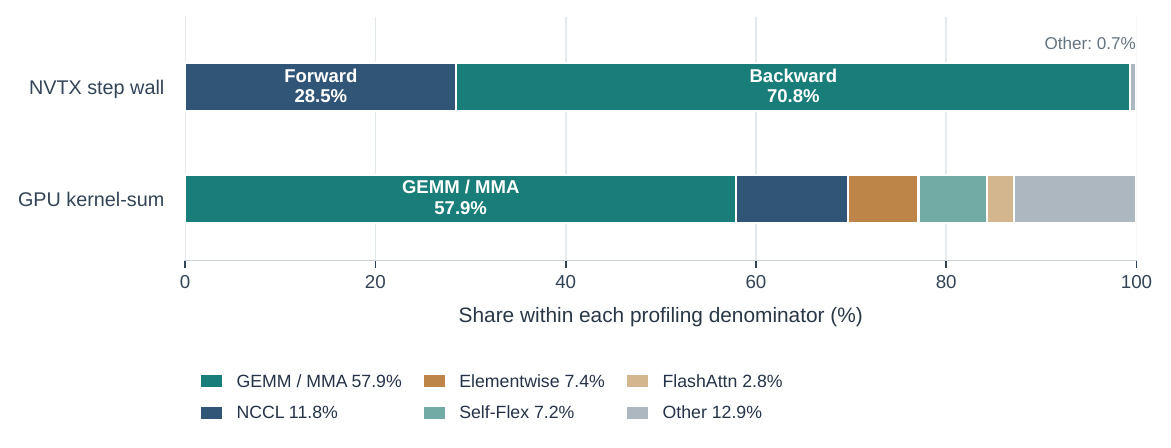}
  \caption{\textbf{Remaining training-time costs after optimization.}
  The upper bar shows the forward/backward phase shares of NVTX step wall;
  the lower bar shows the fraction of cumulative GPU kernel duration by category.
  These bars intentionally use \emph{different denominators}: kernel sums may
  exceed step wall because of overlapping streams. GPU kernel busy is 99.22\%,
  NCCL overlap is 64.81\%, and exposed NCCL time accounts for only 4.48\%
  of the profiled step wall.}
  \label{fig:lingbotva_profile}
\end{wrapfigure}
Figure~\ref{fig:lingbotva_profile} summarizes the optimized training profile.
Backward accounts for 70.81\% of the profiled step wall, including activation
recomputation, while GEMM/MMA operators represent 57.91\% of cumulative GPU
kernel time. GPU kernel busy reaches 99.22\%, and 64.81\% of NCCL execution
overlaps non-communication GPU work, leaving 4.48\% of the profiled step
wall as exposed NCCL time. The optimized execution is therefore dominated
by sustained Transformer computation rather than avoidable host-launch gaps
or fully exposed collective communication. Together, the ablation,
memory comparison, and profiling results support the effectiveness of
LingBot-VA-specific execution and communication optimizations while
highlighting the tradeoff between parameter residency and communication cost.

\FloatBarrier

\subsection{DreamZero}
\label{app:dreamzero}

\paragraph{Architecture and bottlenecks.}
DreamZero is a WAM that jointly models future visual dynamics and robot actions through a video diffusion backbone. We consider its Wan2.2-TI2V-5B variant, which extends a 30-layer DiT with action and state registers, allowing video and action tokens to interact through shared transformer blocks with blockwise causal attention. The model jointly optimizes visual dynamics prediction and action denoising, resulting in substantial computational demands from attention, feed-forward operations, and their backward passes. Beyond computation, its large transformer backbone introduces considerable distributed training overhead. Under FSDP, repeated parameter AllGather operations and parameter resharding between forward and backward execution can expose significant communication latency, particularly when communication cannot be effectively overlapped with computation. Furthermore, DreamZero incorporates pretrained text, video, and visual encoders, whose execution requirements differ across training stages. In particular, frozen components may incur unnecessary parameter communication when included in FSDP-managed modules, while redundant video decoding and feature processing introduce additional CPU, memory, and data-loading overhead. These characteristics make efficient DreamZero training dependent on coordinated optimization of transformer execution, parameter communication, and model-specific data processing.

\paragraph{Reference implementation.}
DreamZero's baseline is built from the official code release and its
Wan2.2-TI2V-5B training script, tuned where necessary for the evaluated
setting; we refer to it as the official-code reference.

\paragraph{Optimization design.}
To address these bottlenecks, LoongForge-Embodied applies a combination of communication-aware, computation-aware, and data-aware optimizations tailored to DreamZero's execution characteristics. First, we optimize FSDP parameter management by disabling resharding after the forward pass to avoid redundant parameter AllGather operations, and employ Delta-FP8 communication to reduce parameter transfer volume through quantized parameter updates while preserving BF16 computation. We further adapt FSDP wrapping and parameter residency to the execution requirements of individual components, excluding appropriate frozen modules from unnecessary sharding and communication while retaining their required computation. Second, we optimize the computationally intensive causal attention blocks through whole-block graph compilation, enabling compiler optimizations across a larger execution scope and reducing overhead associated with fragmented execution. Compared with separately compiling fine-grained attention sub-functions, this approach provides greater opportunities for operation fusion and memory optimization. Finally, we streamline DreamZero's data processing through precomputed video latents and text embeddings, together with first-frame-only processing: when video latents are read from the precomputed cache, the only remaining consumer of raw pixels is the first frame fed to the CLIP image encoder, so the loader decodes one frame per view instead of the full clip, avoiding redundant video decoding and unnecessary data movement. These strategies are complemented by a fused AdamW optimizer to reduce parameter update overhead. Together, these model-aware optimizations address the interaction between computation, communication, and data processing while preserving the underlying training objective.

\paragraph{Optimization attribution.}
\begin{wrapfigure}{r}{0.46\textwidth}
    \centering\captionsetup{font=small}\vspace{-1.0em}
    \includegraphics[width=\linewidth]{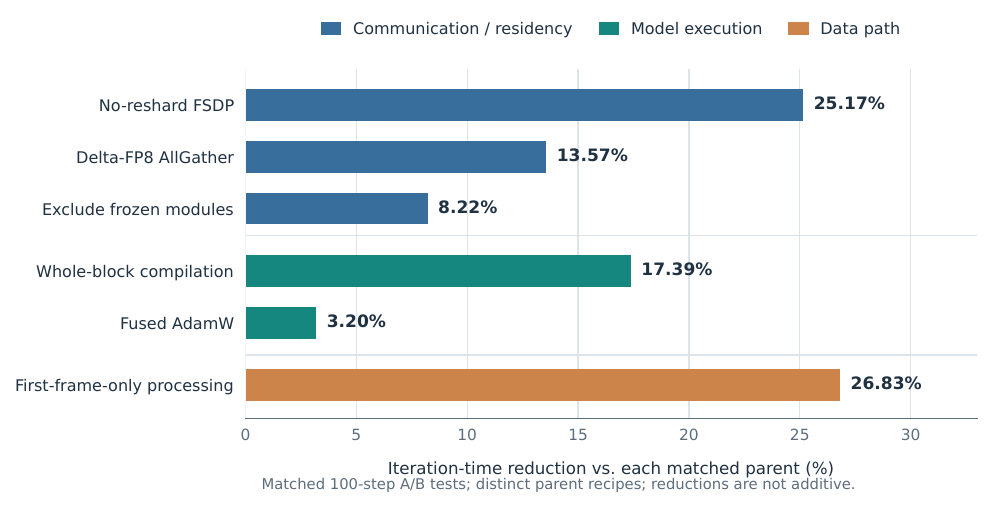}
  \caption{\textbf{DreamZero model-specific optimization attribution.}
  Iteration-time reductions are measured by toggling the indicated optimization
  in a matched baseline on DreamZero Wan2.2-5B. Each reported value uses a
  100-step comparison (steady-state steps 30--100); Delta-FP8 uses a separate
  matched-parent remeasurement. Results come from distinct A/B parents and
  are \emph{not additive}. The first-frame-only comparison includes
  data-loading stalls in its full-frame control.}
  \label{fig:dreamzero_attribution}
\end{wrapfigure}
We first examine the effect of individual optimization decisions on DreamZero
Wan2.2-TI2V-5B. Figure~\ref{fig:dreamzero_attribution} summarizes six
matched A/B comparisons performed with eight GPUs and a fixed strict-cache
workload. Eliminating redundant FSDP resharding reduces iteration time by
25.17\%, while Delta-FP8 AllGather contributes a further 13.57\% reduction
relative to its own matched, no-Delta parent. Excluding appropriately frozen
modules from FSDP-managed parameter communication reduces iteration time by
8.22\%. For model execution, compiling the entire causal attention block
rather than relying solely on more fragmented execution reduces iteration
time by 17.39\%; the fused AdamW optimizer provides an additional 3.20\%
reduction in its separately measured configuration. First-frame-only
processing yields a 26.83\% reduction in its full measurement window.
Separately, on the otherwise optimized online recipe, reading precomputed
video latents and text embeddings instead of running the frozen encoders
every step, with full video decoding unchanged, reduces iteration time by
33.05\% (1.093~s to 0.732~s, averaged over updates 11--20 of 20-update
runs).
Importantly, these improvements are measured against \emph{different} parent
recipes, and therefore must not be summed to estimate the performance of the
complete system.

\paragraph{Delta-FP8 communication mechanism.}
The profiling-based analysis of Delta-FP8 parameter AllGather is
reported in Section~\ref{sec:dist_comm} (Figure~\ref{fig:dreamzero_fp8}).

\begin{figure}[!ht]
    \centering
    \includegraphics[width=0.6\textwidth]{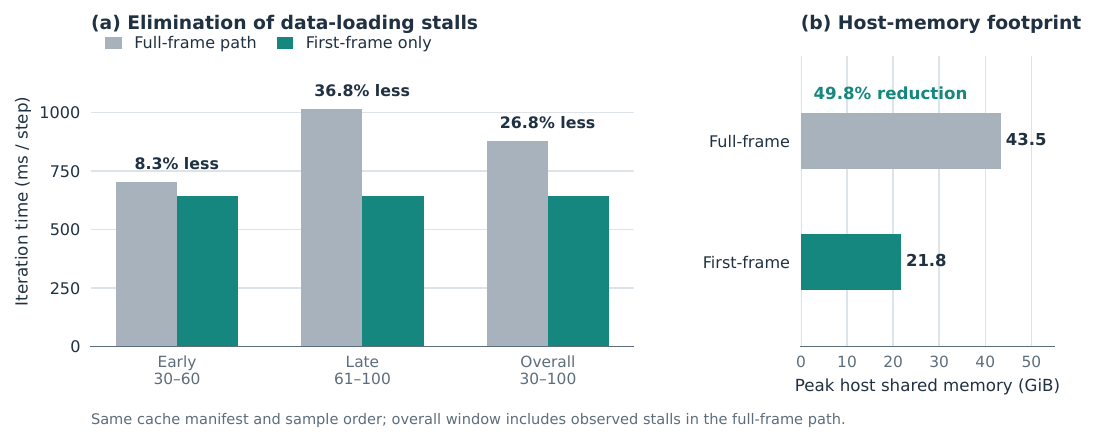}
  \caption{\textbf{DreamZero first-frame-only data processing.}
  (a) Iteration time for the early, late, and complete steady-state windows.
  These overlapping windows are diagnostic views of the same 100-step
  comparison, not independent experiments. (b) Observed peak host shared
  memory under matched strict-cache input and sample order. The full-frame
  control exhibits later-stage data-loading stalls, which are retained in
  the reported full-window result.}
  \label{fig:dreamzero_firstframe}
\end{figure}

\paragraph{First-frame-only data processing.}
DreamZero's precomputed-feature training path also benefits from avoiding
redundant video decoding. Figure~\ref{fig:dreamzero_firstframe} distinguishes
an early window with relatively stable data loading (steps 30--60) from a
late window in which the full-frame control experiences periodic input stalls
(steps 61--100). First-frame-only processing reduces iteration time by 8.28\%
in the early window and 36.77\% in the late window; the full steady-state
window shows a 26.83\% reduction. Peak host shared-memory usage decreases
from 43.48 to 21.82\,GiB. Thus, the optimization improves both the normal
data-processing path and its resilience against host-side input stalls; the
full-window speedup should not be interpreted as a pure GPU compute gain.

\paragraph{Numerical fidelity.}
For numerical fidelity, we compare the optimized variants against a fixed
\begin{wraptable}{r}{0.46\textwidth}
  \centering\captionsetup{font=small}\vspace{-1.0em}
  \caption{\textbf{DreamZero short-window numerical fidelity.}
  The loss reference A0 and the matched Delta-FP8 off/on candidates were
  measured in separate runs with the same initial weights and strict-cache
  data contract. All values are for steps 30--100 of 100-step runs.}
  \label{tab:dreamzero_fidelity}
  \setlength{\tabcolsep}{3pt}
  \small
  \begin{tabular}{lrr}
    \toprule
    Configuration & Step (s) & MARE vs A0 (\%) \\
    \midrule
    Reference A0 & 0.980355 & 0.0000 \\
    Delta-FP8 disabled    & 0.740985 & 0.1686 \\
    Delta-FP8 enabled     & \textbf{0.640407} & \textbf{0.0616} \\
    \bottomrule
  \end{tabular}
\end{wraptable}
reference initialized from the same model weights. As shown in
Table~\ref{tab:dreamzero_fidelity}, the 100-step total-loss MARE of Delta-FP8 enabled relative to the immutable A0
reference is 0.0616\%, while the Delta-FP8 disabled parent differs from A0
by 0.1686\%. The enabled and disabled variants differ by 0.1084\% in
pointwise total-loss MARE. All reported runs complete 100 training steps
without NaNs or skipped updates. These observations establish short-window
numerical consistency but do not constitute evidence of equivalent
long-horizon convergence or downstream task success.

\FloatBarrier

\subsection{Cosmos 3}
\label{app:cosmos3}

\paragraph{Architecture and bottlenecks.}
Cosmos 3 is an omnimodal world model that is built on an MoT architecture that combines autoregressive reasoning with diffusion-based generation of continuous modalities, including video and actions. Its training workflow involves multimodal data preprocessing, VAE-related computation, transformer execution, and FSDP-based distributed training. This heterogeneous workflow introduces several performance bottlenecks. First, CPU-intensive data preprocessing incurs substantial overhead in batch preparation, causing GPU stalls and limiting effective utilization. Second, the large memory footprint of model parameters and intermediate activations constrains the per-device batch size and consequently limits training throughput. Third, although FSDP reduces memory consumption through parameter sharding, repeated parameter gathering and inter-device communication introduce additional overhead that cannot always be hidden behind computation. These bottlenecks exhibit different characteristics across GPU platforms, motivating both model-specific and hardware-aware optimizations.

\paragraph{Optimization design.}
To address these bottlenecks, we optimize Cosmos 3 across data processing, computation, and distributed execution. We first migrate computationally intensive preprocessing operations from the CPU to the GPU, reducing batch preparation overhead and improving GPU utilization. For model execution, we employ torch.compile to optimize computational graphs and reduce memory consumption, enabling larger per-device batch sizes. We further leverage FP8 computation to exploit hardware-supported low-precision arithmetic and improve computational throughput. Beyond these optimizations, we introduce a hardware-aware FSDP strategy that balances memory consumption against communication overhead. Specifically, only frozen weights may skip sharding, since leaving a trainable weight unsharded would also replicate its gradients and optimizer state. In Cosmos 3 SFT the reasoning expert of the MoT backbone is frozen while the generation expert is trained. When device memory permits, we therefore exclude the largest frozen tensors of the reasoning expert (token embeddings, the output head, the feed-forward projections, and the query, key, and value projections) from sharding, keeping them resident on every device instead of re-gathering them at every step. This strategy effectively trades available memory capacity for reduced communication overhead along the critical execution path. Because excluding a frozen tensor from sharding changes only where its parameters reside and not the computation performed, training accuracy is unaffected. Together, these optimizations address the distinct performance bottlenecks of Cosmos 3 while preserving its original training objective.

\paragraph{Optimization analysis.}
\begin{wrapfigure}{r}{0.46\textwidth}
    \centering\captionsetup{font=small}\vspace{-1.0em}
    \includegraphics[width=\linewidth]{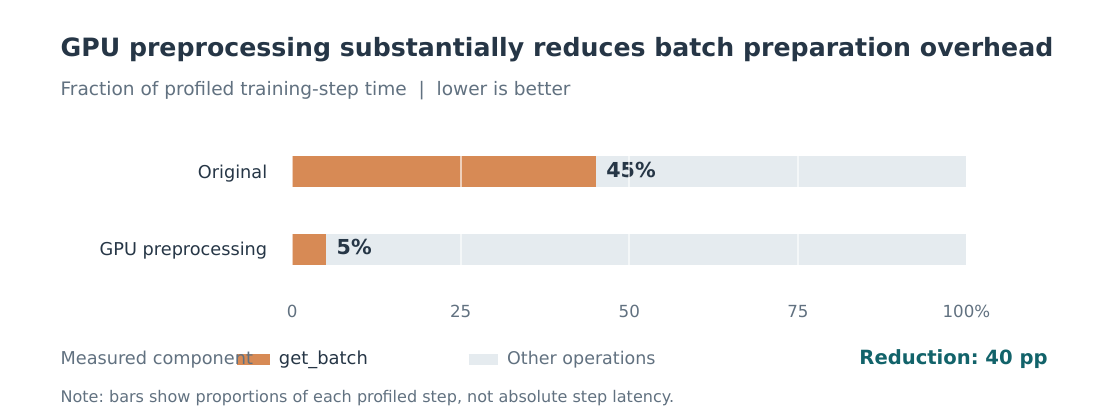}
    \caption{Impact of moving Cosmos 3 preprocessing from the CPU to the GPU.
    The \texttt{get\_batch} share of profiled training-step time drops from
    45\% to 5\%. Both bars show within-iteration time fractions rather
    than measured absolute latencies.}
    \label{fig:cosmos3_preprocessing}
\end{wrapfigure}
We analyze the individual optimizations for Cosmos 3 from three perspectives:
data preparation, execution and memory efficiency, and hardware-aware
computation and communication. Figure~\ref{fig:cosmos3_preprocessing}
shows that migrating CPU-side preprocessing to the GPU decreases the
\texttt{get\_batch} share of the profiled iteration from 45\% to 5\%, a
reduction of 40 percentage points. This result indicates that the data
pipeline is substantially less prominent in the iteration profile.
Because these are relative time shares rather than absolute latencies,
they should not be interpreted as an end-to-end training speedup.

\begin{figure}[!ht]
    \centering
    \includegraphics[width=\textwidth]{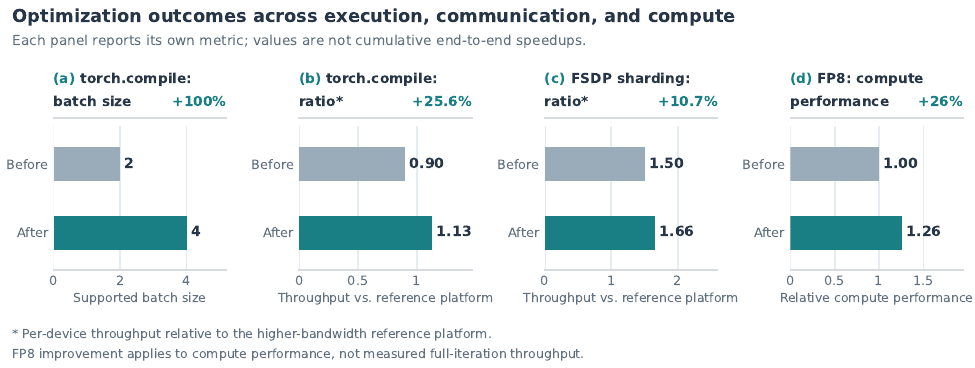}
    \caption{Reported effects of optimizations specific to Cosmos 3,
    all measured on the larger-memory, lower-interconnect-bandwidth
    platform: (a) supported per-device batch size with
    \texttt{torch.compile}; (b) cross-platform throughput ratio with
    compilation enabled; (c) the same ratio under selective FSDP
    sharding; and (d) compute performance with FP8, normalized to a
    baseline of 1.0. The ratio in (c) is per-device training
    throughput on this platform divided by that on the higher-bandwidth
    reference platform at matched model and global batch size; a value
    above 1.0 means one device here replaces more than one reference
    device. The two panels come from separate measurements and do not
    share a baseline. These are metric-specific comparisons, not an
    aggregate end-to-end throughput improvement.}
    \label{fig:cosmos3_optimization}
\end{figure}

Figure~\ref{fig:cosmos3_optimization} summarizes the complementary
execution, communication, and computation improvements. Enabling \texttt{torch.compile} increases the supported per-device batch size from 2 to 4
and raises the cross-platform throughput ratio from 0.90 to 1.13,
moving this platform from below to above per-device parity with the
higher-bandwidth reference. In a separately reported measurement,
reducing FSDP sharding for selected weights raises the same ratio
from 1.50 to 1.66, consistent with the benefit of
trading available device memory for reduced exposed communication.
Enabling FP8 yields an approximately 26\% improvement in compute
performance. These measurements target distinct bottlenecks and were
not reported as a single controlled end-to-end ablation; their gains
therefore cannot be multiplied or added to estimate total speedup.

\begin{table}[!ht]
\centering
\caption{Summary of measured Cosmos 3 optimization outcomes. Values
are from distinct reported measurements and should not be interpreted
as a cumulative speedup.}
\label{tab:cosmos3_optimization}
\small
\setlength{\tabcolsep}{7pt}
\begin{tabularx}{\textwidth}{@{}l X l l@{}}
\toprule
\textbf{Optimization} & \textbf{Measured metric} & \textbf{Before} & \textbf{After} \\
\midrule
GPU preprocessing & \texttt{get\_batch} share of iteration & 45\% & 5\% \\
\texttt{torch.compile} & Supported batch size & 2 & 4 \\
\texttt{torch.compile} & Cross-platform throughput ratio$^{\dagger}$ & 0.90 & 1.13 \\
Selective FSDP sharding & Cross-platform throughput ratio$^{\dagger}$ & 1.50 & 1.66 \\
FP8 computation & Normalized compute performance & 1.00 & 1.26 \\
\bottomrule
\end{tabularx}
\vspace{2pt}
\parbox{\textwidth}{\footnotesize $^{\dagger}$Per-device training throughput on
the larger-memory, lower-interconnect-bandwidth platform divided by per-device
training throughput on the higher-bandwidth reference platform.}
\end{table}

Overall, the available evidence shows that Cosmos 3 benefits from
optimization at multiple levels: GPU-side batch preparation reduces
data-path overhead, graph compilation relaxes the batch-size memory
constraint, selective FSDP sharding exploits available memory to
reduce communication, and FP8 improves computation efficiency.
A uniform samples-per-second measurement under matched training
settings, together with loss and downstream-task comparisons, is
still required to quantify combined end-to-end acceleration and
establish training fidelity across the full optimization stack.

\FloatBarrier

\section{Evaluation Acceleration Details}
\label{app:eval_accel}

\begin{wraptable}{r}{0.46\textwidth}
\centering\captionsetup{font=small}\vspace{-1.0em}
\caption{Throughput sweeps over the number of simulator instances per
inference server. Rates are reported under the steady-state convention
on the sweep workload and are not comparable with
Table~\ref{tab:eval_speedup}.}
\label{tab:eval_n_sweep}
\small
\setlength{\tabcolsep}{5pt}
\begin{tabular}{@{}llr@{}}
\toprule
\textbf{Pool / model} & \textbf{$N$} & \textbf{Episodes/min} \\
\midrule
\multirow{4}{*}{Host CPU / FastWAM}
 & 16 & 70.7 \\
 & 24 & 77.8 \\
 & 48 & 81.8 \\
 & 80 & 81.9 \\
\midrule
\multirow{3}{*}{GPU / $\pi_{0.5}$}
 & 8  & 13.47 \\
 & 10 & 14.05 \\
 & 12 & 12.61 \\
\bottomrule
\end{tabular}
\end{wraptable}

This appendix records the configuration sweeps and timing
arithmetic in
Section~\ref{sec:eval_throughput}.

\paragraph{Parallelism sweeps.}
Table~\ref{tab:eval_n_sweep} lists the throughput sweeps used to select
the number of simulator instances per inference server. For the
host-CPU pool, throughput increases steeply up to $N{=}24$ and is
within half a percent of its peak from $N{=}48$ onward, so additional
instances consume host cores without improving throughput. For the GPU
pool, throughput peaks at $N{=}10$ and degrades at $N{=}12$, where the
per-instance memory footprint and the inference server contend for the
same device. These sweeps serve configuration
selection; their absolute rates are not comparable with the run
throughput of Table~\ref{tab:eval_speedup}.

\paragraph{Thread-pool capping.}
When several simulator instances share one device, each worker process
is capped to a single thread for the numerical libraries and the tensor
framework, rather than allowing each process to size its pools from the
host core count. On a 20-episode run of a single task, this reduces
wall-clock time from 239.7\,s to 215.4\,s, a 10.1\% reduction, with no
change in success rate.

\paragraph{Timing arithmetic.}
The accelerated LIBERO run with FastWAM under software
rendering totals 979\,s, composed of 791\,s of worker
execution and 187\,s of model loading and health checks across eight
inference servers. Fixed startup cost is therefore about 19\% of that
run, whereas the same absolute cost is under 3\% of the 1.9\,h RoboTwin
run; counting worker startup as well, the fixed cost before the first
episode is 0.09 to 0.10\,h, or 5 to 6\% of the RoboTwin runs. The two
reporting conventions therefore diverge far more on short runs than on
full ones.
On the reference path, model
loading was not timed separately and is deducted at the same magnitude
when computing steady-state times, which amounts to approximately
190\,s against a 173.7\,min LIBERO reference time (1.8\%) and a
negligible fraction of the 41.03\,h, 28.6\,h, and 32.0\,h RoboTwin
reference
times (below 0.2\%).

\paragraph{Derived reference time.}
The two RoboTwin reference times for FastWAM are not direct
5,000-episode measurements. They are obtained from a single
10,000-episode reference run over 16 slots at a measured 94\% slot
occupancy. The half of its episodes that corresponds to the clean task
configuration accounts for 457.7 slot-hours, that is 28.6\,h on 16
slots, and the half that corresponds to the randomized configuration
accounts for 512.1 slot-hours, that is 32.0\,h; these are the two
entries in Table~\ref{tab:eval_speedup}. For these two rows no separate
rollout-only measurement exists on the reference path, so the
steady-state column repeats the end-to-end value. The $\pi_{0.5}$
reference time is a direct measurement with the same checkpoint and
protocol over eight slots.

\paragraph{Scheduling variants.}
The $\pi_{0.5}$ run was executed with two dispatch policies. Plain
longest-processing-time-first dispatch with a uniform chunk size of 25
episodes completes in 1.88\,h. Placing all chunks of long-horizon tasks
ahead of short ones and reducing their chunk size to 10 episodes
completes in 1.77\,h, which is the value reported in
Table~\ref{tab:eval_speedup}. Success rates under the two policies
differ within run-to-run variation.

\paragraph{Portability of the configuration.}
The value of $N$ at which throughput saturates depends on host core
count, device memory, model size, and rendering settings. The sweeps
above were obtained in the eight-GPU configuration used throughout this
report and indicate trends rather than a transferable setting. Reusing
a value of $N$ calibrated elsewhere does not raise an error; it
manifests as throughput below the calibrated table, so the sweep is
repeated whenever the hardware, the model, the chunk size, or the
request-merging batch size changes.

\FloatBarrier

\end{document}